\documentclass[11pt]{article}

\usepackage[preprint]{acl}

\usepackage{times}
\usepackage{latexsym}
\usepackage[T1]{fontenc}
\usepackage[utf8]{inputenc}
\usepackage[stretch=40,shrink=30]{microtype}
\usepackage{graphicx}

\providecommand{\DeclareUnicodeCharacter}[2]{}
\DeclareUnicodeCharacter{2264}{\ensuremath{\leq}}
\DeclareUnicodeCharacter{2265}{\ensuremath{\geq}}
\DeclareUnicodeCharacter{2208}{\ensuremath{\in}}
\DeclareUnicodeCharacter{2209}{\ensuremath{\not\in}}
\usepackage{textcomp}
\usepackage{booktabs}
\usepackage{multirow}
\usepackage{array}
\usepackage{amsmath}
\usepackage{amssymb}
\usepackage{pifont}
\usepackage{longtable}
\usepackage{placeins}
\usepackage{multicol}
\AtBeginEnvironment{multicols}{\nolinenumbers}

\usepackage{xcolor}
\usepackage{fvextra}
\DefineVerbatimEnvironment{wrappedverbatim}{Verbatim}{breaklines,breakanywhere,breaksymbol={}}
\usepackage{etoolbox}
\AtBeginEnvironment{wrappedverbatim}{\scriptsize}
\usepackage{tcolorbox}
\tcbuselibrary{skins,breakable}
\tcbset{mypromptstyle/.style={enhanced,breakable,boxrule=0.5pt,colframe=black,arc=4pt,
  colback=gray!3,left=6pt,right=6pt,top=6pt,bottom=6pt,boxsep=2pt}}
\makeatletter
\newcommand{\PromptFile}[2]{%
  \paragraph{#1}\leavevmode\par\@nobreakfalse\penalty0\vspace{-\parskip}%
  \begin{tcolorbox}[mypromptstyle]
  \VerbatimInput[fontsize=\scriptsize,breaklines=true,breakanywhere=true,breaksymbolleft={},breaksymbolright={}]{#2}
  \end{tcolorbox}}
\makeatother
\newcommand{\tabstyle}{\footnotesize\renewcommand{\arraystretch}{1.08}}

\newcommand{\dualcaption}[5]{%
  \refstepcounter{#1}\label{#2}%
  \refstepcounter{#1}\label{#3}%
  \par\begingroup\fontsize{10pt}{12pt}\selectfont
  \leftskip=0pt\rightskip=0pt\parfillskip=0pt plus 1fil
  \noindent\csname #1name\endcsname~\ref{#2}: #4; \csname #1name\endcsname~\ref{#3}: #5\par
  \endgroup}

\title{From Evidence to Judgment: Trace-Augmented Reasoning under Conflicting Evidence}
\author{
  Shubham Mishra$^{\spadesuit}$\quad
  Atharv Sharma$^{\spadesuit}$\quad
  Dhruv Kumar$^{\spadesuit}$\quad
  \textbf{Harsh Sharma}$^{\diamondsuit}$\\[0.5em]
  \normalfont $^{\spadesuit}$Birla Institute of Technology and Science, Pilani\\
  \normalfont $^{\diamondsuit}$Carnegie Mellon University, Pittsburgh\\[0.3em]
  \normalfont \textbf{Correspondence:} \href{mailto:f20220763@pilani.bits-pilani.ac.in}{\texttt{f20220763@pilani.bits-pilani.ac.in}}
}

\begin{document}
\maketitle
\renewcommand{\thefootnote}{}\footnotetext{Code, benchmark, and results are available at an anonymous repository: \url{https://anonymous.4open.science/r/rag_reason-2F8F/}.}
\renewcommand{\thefootnote}{\arabic{footnote}}

\begin{abstract}
Retrieved documents often disagree, contain outdated information, or fail to
support any answer, and retrieval-augmented generation (RAG) systems have no
reliable way to decide which documents to trust. We treat this as a reasoning
problem and test it separately from answer generation. We release $943$
training and validation examples and a disjoint $736$-example benchmark, and fine-tune four
language models to write a structured think trace before answering: the model
judges each document, infers the conflict pattern, and writes a grounded
answer or a justified refusal. Our evaluation framework, CATS (Conflict-Aware
Trust Score), scores grounded refusal, behavioural adherence, factual grounding,
and single-truth recall; a human study validates factual grounding and
single-truth recall ($\kappa=0.78$ and $0.79$). The think trace improves the
answer/refuse decision: under all three prompts, untuned models score at most
$0.429$ Refusal F1 (F1 with refusal as the positive class) and trace-tuned
models at least $0.826$. On the answerable examples a model answers, factual
grounding under the 40-word prompt is at most $0.11$ for untuned and
$0.86$--$0.88$ for trace-tuned models. Cutting the prompt from $1{,}965$ words to
$40$ leaves trace-tuned scores within the benchmark's measurement noise
($\pm0.034$). Answer-only fine-tuning trails trace supervision on every metric
under the minimal prompt.
\end{abstract}

\section{Introduction}
\label{sec:intro}

\begin{figure*}[t]
\centering
\begin{minipage}[b]{3.30in}
\centering
\includegraphics[width=3.30in,height=1.97in]{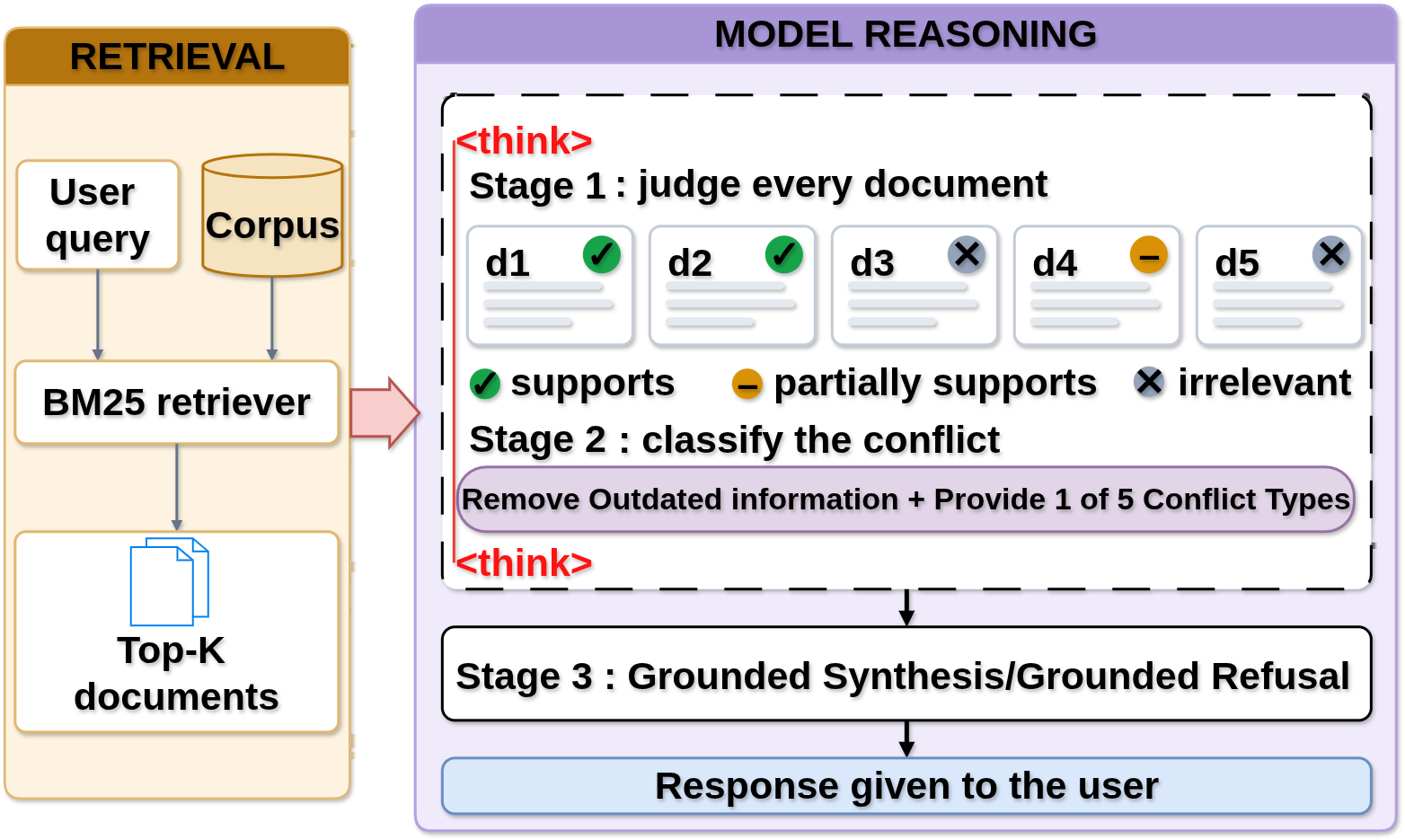}
\end{minipage}
\hfill
\begin{minipage}[b]{2.76in}
\centering
\includegraphics[width=2.76in,height=1.97in]{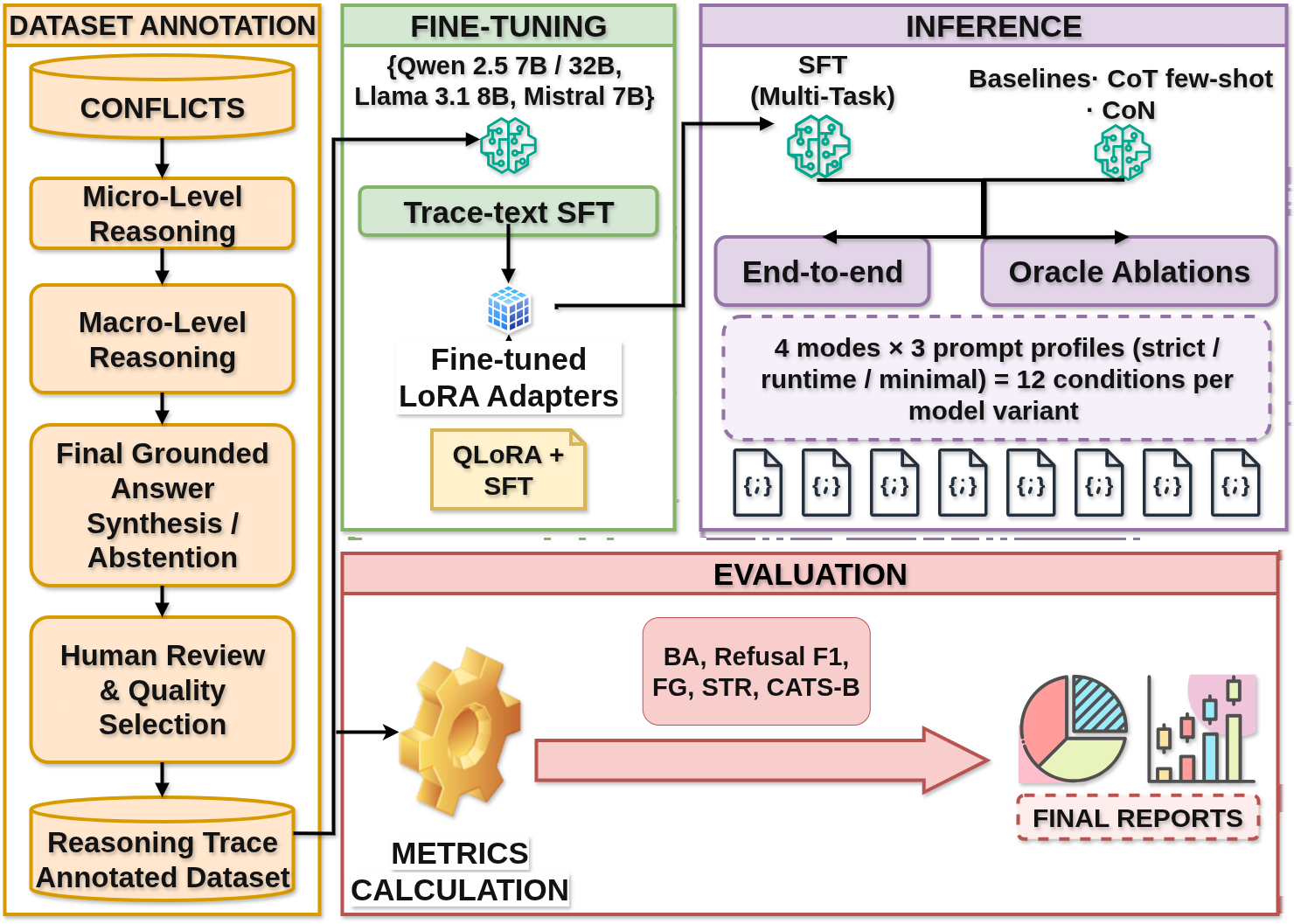}
\end{minipage}
\dualcaption{figure}{fig:rag}{fig:system}%
{The think trace. Left: retrieval returns the top-$K$ documents for a query.
Right: inside the \texttt{<think>} block the model gives each document a verdict
(check: supports; dash: partially supports; cross: irrelevant) and then classifies
the conflict among the documents. Outside the block, Stage~3 writes a grounded
answer or a justified refusal}%
{System overview. A weighted committee of language-model annotators produces the
supervision corpus. Models are fine-tuned to emit the three-stage trace. CATS
scores grounded refusal, behavioural adherence, factual grounding, and
single-truth recall.}
\end{figure*}

Retrieval-augmented generation conditions a generator on a small set of retrieved
snippets
\citep{lewis2020rag,borgeaud2022,gao2024retrievalaugmentedgenerationlargelanguage,guu2020realm},
ideally with faithful attribution to them.
In practice, retrieval rarely returns a coherent evidence set: the documents
vary in relevance, quality, credibility, and recency. They may contain
conflicting claims, provide only partial support, or collectively be
insufficient to answer the user's query.

Such a set requires the model to reason in sequence before it writes a single
word of its answer. It must decide which document says what, how the documents
relate to one another, and whether what remains is sufficient to answer. A
system that treats the snippets as interchangeable context produces a fluent,
confidently wrong answer. A system that refuses whenever sources disagree is
no better, since disagreement is often compatible with a properly scoped
answer.

Existing work addresses parts of this problem. \textit{DRAGged into Conflicts}
\citep{cattan2025dragged} provides expert conflict-type labels, and \textit{Chain-of-Note}
\citep{yu2024chainofnote} produces per-document notes.
\textit{Trust-Score} and \textit{Trust-Align} \citep{song2025trustalign} measure and
train refusal and citation behaviour. Closest in framing, \citet{wu2026conflictaware}
derive a conflict signal from generation probabilities and use it to select
training data and build preference pairs; our think trace records the reasoning
as explicit text.
\citet{wang2025conflictingevidence} resolve conflicting evidence through
inference-time multi-agent debate, and \citet{wang2026conflictrag} pair a learned
conflict detector with credibility scoring. \citet{muhamed2025refusalbench}
evaluate selective refusal under flawed context and find that it is a trainable
capability, which is consistent with our results. The gap that remains is a
single framework that supervises and evaluates per-document reasoning,
cross-document conflict inference, and justified abstention together.

We close that gap with a \emph{structured think trace}. Before answering, the
model writes out its own reasoning in three stages (Figure~\ref{fig:rag}).
\textbf{Stage~1} addresses which document says what: it assigns each snippet a
verdict (\textit{supports}, \textit{partially supports} or \textit{irrelevant}
for the query) with an extracted key fact and a source-quality label.
\textbf{Stage~2} addresses how the documents relate to one another: it
aggregates those verdicts into one of five conflict patterns, following the
taxonomy of \citet{cattan2025dragged}. \textbf{Stage~3} addresses whether what
remains is sufficient to answer: it produces a citation-linked answer or a
justified abstention, conditioned on that pattern (\S\ref{sec:method}). The
model emits the trace in a fixed output format. This makes the trace
machine-checkable and lets us attribute a failure to a specific stage.
\citet{liu2026craft} also train auditable reasoning traces, by reinforcement
learning on multi-hop QA. Prompting can produce conflict-aware answers, but in our experiments it does not
produce a reliable decision of whether to answer at all. Separating the two is
this paper's central test.

We run this test on a new 736-example benchmark whose examples and queries do
not appear in the training data (\S\ref{sec:data}). Trace supervision raises
Refusal F1, the F1 score for correctly refusing when the retrieved documents
cannot support an answer, from at most $0.429$ for any untuned model to at least
$0.826$ for every fine-tuned model, under each of the three prompts we test
(1,965, 195 and 40 words long; \S\ref{sec:sft}), while untuned behaviour
varies widely across them. Under the 40-word prompt, factual grounding on the
answerable examples a model answers separates the models in the same way
(\S\ref{sec:whatdoesnot}).

\textbf{Contributions.} \textbf{(1) A conflict-aware RAG resource}
(\S\ref{sec:data}): 943 training and validation examples and a disjoint
736-example benchmark. Each of the $10{,}343$ documents carries a verdict and a
source-quality label; non-irrelevant documents also carry a key fact, a
snippet-anchored quote and a rationale.
\textbf{(2) Evidence that trace supervision improves refusal and grounding}
(\S\ref{sec:results}): Refusal F1 rises from at most $0.429$ to at least
$0.826$, grounding on the answerable examples a model answers rises under sparse prompts, and
fine-tuned behaviour stays stable when we reduce the prompt to forty words. \textbf{(3) CATS} (\S\ref{sec:cats}): an evaluation framework that
reports four components separately, each over the examples to which it applies.
We validate factual grounding and single-truth recall against a human study and
compute grounded refusal deterministically.
\textbf{(4) Oracle ablations} (\S\ref{sec:oracle}): supplying gold information
to one stage at a time shows which stage limits the system and supports the
three-stage decomposition.

\section{Related Work}
\label{sec:related}

\subsection{Knowledge and Evidence Conflicts}

Retrieval-augmented generation conditions a model on corpus-retrieved documents at
inference time
\citep{guu2020realm,lewis2020rag,izacard2023atlas,borgeaud2022,gao2024retrievalaugmentedgenerationlargelanguage,fan2024surveyragmeetingllms,ram-etal-2023-context}.
Retrieved documents routinely carry conflicting information, which complicates
generation and has become a research area of its own
\citep{xu2024knowledgeconflictsllmssurvey}. We scope this paper to conflicts
\emph{among retrieved documents}. A model's parametric knowledge is the
information it memorized during training. Conflict between that knowledge and
retrieved evidence, the context--memory setting targeted by much recent work
\citep{huo2025microact,zhang2025faithfulrag,ye2026seeingthrough}, is out of scope
here. We evaluate whether our systems reason correctly over the supplied document
set. \textit{ConflictBank} \citep{su2024conflictbankbenchmarkevaluatinginfluence} evaluates how models handle conflicts within retrieved knowledge, within parametric knowledge, and in their interplay. \textit{WikiContradict} \citep{hou2024wikicontradict} evaluates how models answer given explicit and implicit contradictions between Wikipedia passages, using human-annotated instances. \textit{ECON} \citep{jiayang2024econ} studies detection and resolution of evidence conflicts directly.
\textit{DRAGged into Conflicts} \citep{cattan2025dragged} is closest to our
setting. It proposes a conflict taxonomy spanning no-conflict,
complementary, conflicting-opinion, outdated, and misinformation cases. It demonstrates that models often fail
to adapt their responses without explicit conflict-aware guidance. Most of this
literature focuses on \emph{detecting} or \emph{classifying} conflict.
Our framework supervises the reasoning that produces the
classification. It also evaluates whether a model's document-level verdicts,
conflict inference, and final response behaviour are mutually consistent.

\subsection{Reasoning Supervision and Robustness}

Several lines of work supervise or elicit intermediate reasoning.
\textit{Chain-of-Thought} prompting established that generating intermediate
steps improves complex reasoning \citep{wei2022cot}. \textit{Chain-of-Note}
\citep{yu2024chainofnote} adapts this to retrieval by generating sequential
reading notes for each retrieved document. This improves robustness to noisy
and irrelevant passages and lets a model answer ``unknown'' when neither
retrieved nor intrinsic knowledge suffices. Adaptive adversarial training
\citep{fang2024enhancingnoiserobustnessretrievalaugmented} and robustness
training \citep{yoran2024makingretrievalaugmentedlanguagemodels} mitigate
retrieval noise, and \textit{GSM-IC} \citep{shi2023largelanguagemodelseasily}
shows that irrelevant context substantially degrades reasoning. Our work differs in two respects. First, each stage has a closed label set, so
Stage-1 and Stage-2 decisions can be scored separately against gold
annotations. Second, and more importantly, \textit{we treat the reasoning trace
as an object of evaluation.} This requires the three stages to be separable. We
test this by supplying gold information for one stage at a time and asking
whether each intervention moves only the metrics that stage should govern
(\S\ref{sec:oracle}).

\subsection{Measuring Trust and Conflict Alignment}

Trustworthiness metrics for RAG largely score an answer against the retrieved
set as a whole. Conflict-aware evaluation requires that the
\emph{expected} response depend on the kind of disagreement present. A balanced
summary and a decisive correction are different correct behaviours, so the
scoring rule has to depend on the conflict type. Evaluations of RAG commonly report end-to-end answer quality
\citep{gao2024retrievalaugmentedgenerationlargelanguage}. \textit{Trust-Score} \citep{song2025trustalign}
advances this by evaluating trustworthiness through answer
correctness, citation reliability, and refusal behaviour. Its companion
\textit{Trust-Align} aligns models to improve that score. Trust-Align's
held-out refusal material forms Track~B (refusal; \S\ref{sec:data}) of our
resource.
\textit{FaithEval}
\citep{ming2025faitheval} probes whether models stay faithful to context when it is
unanswerable, inconsistent or counterfactual. CATS extends this line in three ways. It conditions the expected response on
the inferred conflict type. It evaluates grounding at the claim level, as in fine-grained citation evaluation
\citep{xu2024aliiceevaluatingpositionalfinegrained}, counting a
claim only when a document whose annotated verdict is \textit{supports} or
\textit{partially supports} backs it and the response cites that document. It also
reports every component over the subset of examples to which it applies, because
the four CATS components apply to different subsets and pooling them would mix
different example sets in one comparison (\S\ref{sec:cats}). Following
\citet{zheng2023judging}, we validate the LLM judges against a human study and
report agreement for each judged component (\S\ref{sec:human}).

\section{Task Formulation and the Think Trace}
\label{sec:method}

An example consists of a query $q$ and a retrieved set $D=\{d_1,\dots,d_N\}$. The
system must produce either a citation-linked answer or the controlled refusal
string \texttt{CANNOT ANSWER, INSUFFICIENT EVIDENCE}. It reasons only over the
retrieved evidence and refuses when that evidence is insufficient. We supervise
an intermediate representation, the think trace, that the model emits before its
response.

\textbf{Stage 1: micro reasoning (document-level judgment).} Every $d_i$ receives a verdict in
\{\textit{supports}, \textit{partially supports}, \textit{irrelevant}\}, a
one-sentence key fact entailed by the snippet, and a coarse source-quality label
derived from the document's URL. The \textit{partially supports} class is broad,
covering on-topic but incomplete, hedged, scoped, or single-facet evidence,
because binary useful/useless labels cause refusal whenever no single snippet is
individually decisive. \textbf{Stage 2: macro reasoning (conflict analysis).}
The Stage-1 verdicts are aggregated into
exactly one of five conflict types: no conflict, complementary information,
conflicting opinions or research outcomes, outdated information, or
misinformation.
Table~\ref{tab:taxonomy} (Appendix~\ref{app:cats}) gives each type's definition
and the response behaviour it requires. The taxonomy separates
\emph{complementarity} from \emph{conflict}: two snippets can contribute
different necessary pieces without disagreeing. \textbf{Stage 3: grounded synthesis.}
Stages 1 and 2 are emitted \emph{inside}
the \verb|<think>| block. Stage 3 is the response itself and sits outside it. The
final response is conditioned on the conflict label and cites documents at
sentence level with markers such as \texttt{[d3]}, or it is the exact refusal
string. Either form ends with the marker \verb|[[END-OF-ANSWER]]|.
Figure~\ref{fig:system} places this in the overall
pipeline.

\section{Dataset Generation and Validation}
\label{sec:data}

Table~\ref{tab:data} gives the size and composition of the training and benchmark
splits, and Table~\ref{tab:cats} previews the four CATS components defined in
\S\ref{sec:cats}. Appendix~\ref{app:data} gives the construction record and retrieval protocol;
Appendix~\ref{app:human-review} gives the human-review statistics.

\begin{table*}[t]
\centering
\begin{minipage}[t]{0.48\linewidth}\vspace{0pt}
\centering
{\tabstyle\setlength{\tabcolsep}{3pt}
\begin{tabular*}{\linewidth}[t]{@{\extracolsep{\fill}}l rr rr@{}}\toprule
& \multicolumn{2}{c}{\textbf{Train+val}} & \multicolumn{2}{c}{\textbf{Benchmark}}\\
\cmidrule(lr){2-3}\cmidrule(lr){4-5}
\textbf{Conflict type} & $n$ & \% & $n$ & \%\\\midrule
No conflict & 343 & 36.4 & 211 & 28.7\\
Complementary & 286 & 30.3 & 221 & 30.0\\
Conflicting opinions & 158 & 16.8 & 109 & 14.8\\
Outdated & 136 & 14.4 & 158 & 21.5\\
Misinformation & 20 & 2.1 & 37 & 5.0\\
\midrule
Answerable & 649 & 68.8 & 608 & 82.6\\
Refusal & 294 & 31.2 & 128 & 17.4\\
\midrule
\textbf{Records} & \textbf{943} & & \textbf{736} & \\
Retrieved documents & 6,642 & & 3,701 & \\
Docs / record (mean) & 7.04 & & 5.03 & \\
Unique source domains & 2,633 & & 1,431 & \\
\bottomrule\end{tabular*}}

\end{minipage}%
\hspace{0.04\linewidth}%
\begin{minipage}[t]{0.48\linewidth}\vspace{0pt}
\centering
{\tabstyle\renewcommand{\arraystretch}{1.826}\setlength{\tabcolsep}{0pt}
\begin{tabular}[t]{@{}>{\raggedright\arraybackslash}p{\linewidth}@{}}\toprule
\textbf{Grounded Refusal (GR)}\newline whether the system answered exactly when the evidence permits answering; scored deterministically from the answerability label\\[3pt]
\textbf{Behaviour Adherence (BA)}\newline whether the answer followed the response policy its conflict type requires\\[3pt]
\textbf{Factual Grounding (FG)}\newline whether each extracted claim is both supported by an eligible document \emph{and} cited\\[3pt]
\textbf{Single-Truth Recall (STR)}\newline whether the response adopts the target answer as its own conclusion (quoting it or attributing it to a source does not count)\\
\bottomrule\end{tabular}}

\end{minipage}
\dualcaption{table}{tab:data}{tab:cats}%
{\textbf{Dataset composition and scale.} Conflict-type and answerability distribution of the training split (943 records, including validation) and the held-out benchmark (736). $n$ is the number of records and \% their share of the split; refusal examples are those whose gold response is the refusal string (\S\ref{sec:method}). The benchmark has fewer no-conflict cases ($28.7\%$ against $36.4\%$), a $1.5\times$ larger outdated-information share, and a $2.4\times$ larger misinformation share. Full per-source composition in Appendix~\ref{app:data}}%
{\textbf{The four CATS components}, each reported over the subset of examples to which it applies. Appendix~\ref{app:cats} gives the full definitions, the rule for which documents can support a claim (FG), the response policy for each conflict type (BA) and the voting rule of the judge committee.}
\end{table*}

\subsection{Training and validation dataset}

The 943-example training release (862 train / 81 validation) holds 649 answerable
and 294 refusal examples. It combines
three source families: \textsc{Conflicts} \citep{cattan2025dragged}, held-out
refusal examples derived from Trust-Align \citep{song2025trustalign}, and
candidate examples from the benchmark construction pool that were not admitted to the
final benchmark (Appendix~\ref{app:data}).

\subsection{Held-out benchmark dataset}

The benchmark has three tracks. Track~A (508 examples) holds queries
with live web-retrieved documents from four open-domain QA collections. Track~B
(128) holds refusal examples derived from Trust-Align
\citep{song2025trustalign}. Track~C (100) is a curated set for conflict types
that open-web retrieval cannot reliably surface, such as a fact whose truth
value has since changed (Appendix~\ref{app:data}). The tracks cover both answer and refuse decisions and all five conflict types,
with no claim about natural prevalence. The refusal examples come from an
independent, published resource, and two humans reviewed all 128. Refusal F1 uses
all $736$ examples; Behaviour Adherence and Factual Grounding use every example
except correct refusals, and Single-Truth Recall uses its $487$ single-truth
examples (\S\ref{sec:cats}).

\subsection{Annotation and human review}

Each of the $10{,}343$ retrieved documents carries a \textit{verdict} and a
\textit{source-quality label}; non-irrelevant documents also carry an extracted
\textit{key fact}, a snippet-anchored \textit{supporting quote} and a
\textit{rationale}.
Labels come from an annotation committee, a weighted panel of language-model
annotators. At each stage the committee votes on the categorical decision. The key fact, quote, and
rationale are all taken from the highest-weight member that voted for the
winning decision, so the verdict, quote, and rationale agree with one another.
The annotation rules require the key fact to be entailed by its quote.
Appendix~\ref{app:annotation} gives the aggregation rule and the full rules, and
Appendix~\ref{app:annprompts} the released prompts.

Two separate human reviews cover the benchmark and the training data. For the
benchmark, seven reviewers screened the candidate examples: they decided whether
to retain each one, assigned a preliminary conflict type, and rated eight
quality criteria (Appendix~\ref{app:human-review} lists each). This screening
reached $94.77\%$ five-way conflict-type agreement (Cohen's $\kappa=0.92$) and
$98.21\%$ retention agreement ($\kappa=0.92$). Training-side review (two reviewers per example)
reached $83.46\%$ conflict-type agreement ($\kappa=0.77$). A separate study of
100 query--document pairs, balanced across the three verdict classes, checks the
Stage-1 per-document verdict. Three reviewers agree at Fleiss'
$\kappa=0.83$ and agree with the committee at $92\%$ ($\kappa=0.88$ against
the human majority). Appendix~\ref{app:human-review} gives the full protocols.

\section{Trace-Supervised Fine-Tuning}
\label{sec:sft}

\subsection{Objective and prompt variants}

The supervision target is the serialised trace followed by the final response.
Three prompt variants are used: the \textbf{strict} prompt states the full
taxonomy, verdict heuristics, and output format ($1{,}965$ words). The
\textbf{runtime} prompt gives a compact output schema ($195$ words). The
\textbf{minimal} prompt supplies only the query, the documents, and a
four-sentence evidence policy ($40$ words). The training mixture also includes
separate examples that predict only the document verdicts, only the conflict
type, or only the final response, so each component is also trained in
isolation. We train and evaluate across all three variants to test whether
conflict-aware behaviour persists when the prompt gives less instruction
(\S\ref{sec:internalization}).
Appendix~\ref{app:e2eprompts} gives the three prompts verbatim and their
design rationale.

\subsection{Fine-tuning}
All models are fine-tuned with QLoRA
\citep{dettmers2023qloraefficientfinetuningquantized,hu2021loralowrankadaptationlarge},
a parameter-efficient method that trains low-rank adapters (LoRA) atop a
frozen, quantized base model. Training uses the standard
token-level cross-entropy of the trace and response, with no auxiliary loss.
Appendix~\ref{app:sft} gives the hyperparameters, the validation-based checkpoint
rule, and the two training mixtures: recipe~A for Qwen2.5-7B and Qwen2.5-32B, and
recipe~B for Llama-3.1-8B and Mistral-7B. A recipe fixes which task types and
prompt variants a run trains on, and in what proportion.

\section{CATS: Conflict-Aware Trust Score}
\label{sec:cats}

Standard RAG metrics score distinct failures as one bad answer: answering when
evidence is absent, mishandling the conflict type, making unsupported claims,
and omitting the target answer. CATS scores them as four components, each over
the examples to which it applies.

Table~\ref{tab:cats} states what each component measures. Grounded Refusal (GR)
is reported as Refusal F1 and covers all examples. Behaviour Adherence (BA) and
Factual Grounding (FG) cover every example except correct refusals. Single-Truth
Recall (STR) covers the 487 benchmark examples that are answerable, have a gold
answer, and have a single correct conclusion, so it excludes
conflicting-opinion examples. FG counts a claim only when the document
supporting it has a gold (annotated) verdict of \textit{supports} or
\textit{partially supports}. Appendix~\ref{app:cats} sets out each rule. Because
each component is averaged over a different set of examples, scores of different
components, or of other frameworks, are not directly comparable.

\textbf{Judges.} BA, FG, and STR are judged by a judge committee of three
open-weight language models run locally, combined by weighted majority vote with
fixed priority weights ($6$, $3$ and $2$). FG support additionally requires
agreement from at least two judges. Grounded Refusal (GR) is computed from the
answerability label and needs no judge. \S\ref{sec:human} compares the judges
with human review. The judge prompts (one system message, three task prompts,
and the five-type behaviour rubric) appear verbatim in
Appendix~\ref{app:catsprompts}. Metric definitions and the committee's voting rule are in
Appendix~\ref{app:cats}.

\section{Experimental Setup}
\label{sec:setup}

\subsection{Models and evaluation protocol}

{\emergencystretch=3em
We fine-tune four instruction-tuned open-weight models, Qwen2.5-7B and
Qwen2.5-32B \citep{yang2024qwen25}, Llama-3.1-8B \citep{dubey2024llama3} and
Mistral-7B \citep{jiang2023mistral7b}, and compare each against its untuned base
model, the baseline (column \textit{base} in tables), on the four CATS
components. We call the fine-tuned version trace-tuned (\textsc{sft} in tables).
For each model and prompt, the base and fine-tuned versions use the same benchmark examples, prompts, decoding policy,
retry policy, post-processing, and evaluator, and differ only in the LoRA
adapter (Table~\ref{tab:main}). We call one model under one prompt a
configuration. We evaluate every model under the strict and minimal prompts,
with the intermediate runtime prompt reported in Appendix~\ref{app:full}.
Generation is deterministic (temperature $0$), with a token budget set per
prompt type, and stops at the end-of-answer marker. When the output does not
meet the required format, we retry the identical prompt with a larger token
budget and never resample or edit content (Appendix~\ref{app:inference}).
Per-example scores lie in $[0,1]$, so a distribution-free 95\% interval
half-width at $n=736$ is $\pm0.034$. We call this the noise floor and treat
differences below it as uninformative. We report exact sign tests for paired
comparisons across the twelve configurations (four models, three prompts), and
use exact Clopper--Pearson (binomial) intervals on the 128-example refusal
subgroup.\par}

\subsection{Ablation studies}

\textbf{Oracle ablations.} An oracle condition adds the correct (gold)
annotation for one stage to the model's input. To test whether the three stages
are separable, we re-run the fine-tuned and untuned models with one stage's gold
annotation added at a time (\S\ref{sec:oracle}). The \textbf{+conflict label} condition
supplies the gold Stage-2 conflict type, \textbf{+per-doc notes} supplies the
gold Stage-1 verdicts and key facts for each document, and \textbf{+both}
supplies both.
\textbf{Comparison with prompting methods.} \emph{Chain-of-Note}
\citep{yu2024chainofnote} writes per-document reading notes before answering,
the closest analogue to our Stage~1. \emph{Few-shot Chain-of-Thought}
\citep{wei2022cot} supplies worked exemplars of the reasoning format. Both
receive the identical benchmark, decoding policy, and evaluator
(\S\ref{sec:comparators}).

\section{Results}
\label{sec:results}

Table~\ref{tab:main} reports results for each model under the strict and minimal
prompts. For each metric it shows the untuned base model next to its supervised
fine-tuned (SFT) version. Appendix~\ref{app:full} gives all twelve
model$\times$prompt configurations.

\setcounter{dbltopnumber}{1}
\begin{table*}[t]
\centering
{\tabstyle\setlength{\tabcolsep}{4pt}
\begin{tabular*}{\linewidth}[t]{@{\extracolsep{\fill}}l l c@{\hspace{2pt}}c c@{\hspace{2pt}}c c@{\hspace{2pt}}c c@{\hspace{2pt}}c@{}}\toprule
& & \multicolumn{2}{c}{\shortstack{\textbf{Refusal}\\\textbf{F1}}} & \multicolumn{2}{c}{\shortstack{\textbf{Behaviour}\\\textbf{Adherence}}} & \multicolumn{2}{c}{\shortstack{\textbf{Factual}\\\textbf{Grounding}}} & \multicolumn{2}{c}{\shortstack{\textbf{Single-Truth}\\\textbf{Recall}}}\\
\cmidrule(lr){3-4}\cmidrule(lr){5-6}\cmidrule(lr){7-8}\cmidrule(lr){9-10}
\textbf{Model} & \textbf{Prompt} & \textbf{base} & \textbf{\textsc{sft}} & \textbf{base} & \textbf{\textsc{sft}} & \textbf{base} & \textbf{\textsc{sft}} & \textbf{base} & \textbf{\textsc{sft}}\\\midrule
\multirow{2}{*}{Qwen2.5-7B} & strict & 0.429 & \textbf{0.911} & 0.811 & 0.780 & 0.617 & \textbf{0.838} & 0.561 & \textbf{0.687}\\
 & minimal & 0.031 & \textbf{0.881} & 0.737 & \textbf{0.768} & 0.031 & \textbf{0.837} & 0.757 & 0.686\\
\addlinespace[2pt]
\multirow{2}{*}{Qwen2.5-32B} & strict & 0.367 & \textbf{0.905} & 0.789 & \textbf{0.821} & 0.851 & 0.812 & 0.733 & \textbf{0.749}\\
 & minimal & 0.015 & \textbf{0.911} & 0.747 & \textbf{0.809} & 0.108 & \textbf{0.822} & 0.773 & 0.738\\
\addlinespace[2pt]
\multirow{2}{*}{Llama-3.1-8B} & strict & 0.255 & \textbf{0.926} & 0.721 & \textbf{0.805} & 0.705 & \textbf{0.852} & 0.634 & \textbf{0.727}\\
 & minimal & 0.015 & \textbf{0.934} & 0.743 & \textbf{0.795} & 0.074 & \textbf{0.857} & 0.710 & \textbf{0.714}\\
\addlinespace[2pt]
\multirow{2}{*}{Mistral-7B} & strict & 0.352 & \textbf{0.826} & 0.660 & \textbf{0.793} & 0.624 & \textbf{0.796} & 0.563 & \textbf{0.644}\\
 & minimal & 0.000 & \textbf{0.839} & 0.727 & \textbf{0.786} & 0.079 & \textbf{0.792} & 0.700 & 0.627\\
\bottomrule\end{tabular*}}

\caption{\textbf{End-to-end results on the 736-example held-out benchmark.} \textit{base} is the untuned model and \textsc{sft} the same model after trace supervision. Refusal F1 rises from at most $0.429$ to at least $0.826$ in all eight pairs, and under the minimal prompt base Factual Grounding falls to $0.031$--$0.108$ while \textsc{sft} stays at $0.792$--$0.857$. Within each pair the benchmark, prompts, decoding policy and evaluator are identical, and only the LoRA adapter differs. Bold marks a fine-tuned value that exceeds its baseline. Differences below the $\pm0.034$ noise floor (\S\ref{sec:setup}) are uninformative. We show the strict and minimal prompts, the longest and shortest instructions. The intermediate runtime prompt is in Appendix~\ref{app:full}.}
\label{tab:main}
\end{table*}

\begin{table*}[t]
\centering
{\tabstyle\setlength{\tabcolsep}{6pt}
\begin{tabular*}{\linewidth}[t]{@{\extracolsep{\fill}}l l rrrr@{}}\toprule
\textbf{Model} & \textbf{Method} & \textbf{Ref.\,F1} & \shortstack{\textbf{Behaviour}\\\textbf{Adherence}} & \shortstack{\textbf{Factual}\\\textbf{Grounding}} & \shortstack{\textbf{Single-Truth}\\\textbf{Recall}}\\\midrule
\multirow{3}{*}{Llama-3.1-8B} & CoN (prompt-only) & 0.000 & 0.632 & 0.609 & 0.658\\
 & CoT (few-shot) & 0.302 & 0.716 & 0.635 & 0.678\\
 & \textsc{trace-sft} (ours) & \textbf{0.926} & 0.805 & 0.852 & 0.727\\
\addlinespace[2pt]
\multirow{3}{*}{Mistral-7B} & CoN (prompt-only) & 0.000 & 0.685 & 0.522 & 0.676\\
 & CoT (few-shot) & 0.525 & 0.657 & 0.417 & 0.528\\
 & \textsc{trace-sft} (ours) & \textbf{0.826} & 0.793 & 0.796 & 0.644\\
\addlinespace[2pt]
\multirow{3}{*}{Qwen2.5-7B} & CoN (prompt-only) & 0.000 & 0.727 & 0.666 & 0.719\\
 & CoT (few-shot) & 0.402 & 0.781 & 0.734 & 0.724\\
 & \textsc{trace-sft} (ours) & \textbf{0.911} & 0.780 & 0.838 & 0.687\\
\bottomrule\end{tabular*}}

\caption{\textbf{External prompting comparators.} CoN and CoT run the untuned model with Chain-of-Note and few-shot Chain-of-Thought prompts. Rows cover the three 7B and 8B models. \textsc{trace-sft} is our trace-tuned model under the strict prompt, so every method receives a detailed instruction. Ref.\,F1 is Refusal F1.}
\label{tab:comparators}
\end{table*}

\subsection{Answer/refuse decisions improve sharply}
\label{sec:refusal}

The largest and cleanest effect is on the answer/\allowbreak refuse decision. Refusal F1 is
the F1 score for refusing: precision is the share of the model's refusals that
were required, and recall is the share of the 128 refusal examples that
it refused. Every baseline configuration scores $\leq0.429$ on Refusal F1 and
every trace-tuned configuration $\geq0.826$, a mean improvement of $+0.709$
across the eight configurations in Table~\ref{tab:main}.
Decision accuracy, the share of examples that the model correctly answered or
refused, also rises in every configuration. The benchmark is $82.6\%$ answerable,
so a system that answers every query already reaches a decision accuracy of
$0.826$. No baseline configuration exceeds that value by more than the
$\pm0.034$ noise interval, while every trace-tuned configuration reaches
$0.927$--$0.976$. Behaviour Adherence (whether the response follows the policy
its conflict type requires) also improves in $7$ of the $8$ configurations in
Table~\ref{tab:main}, with larger gains for base models that follow instructions
less well. For example, Llama-3.1-8B under the minimal prompt has a decision
accuracy of $0.825$ but near-zero refusal recall.

\subsection{Learnt behaviour is prompt-invariant}
\label{sec:internalization}

Shortening the prompt barely changes the fine-tuned models and sharply lowers the
baseline models' scores. The minimal prompt is forty words long and never names
the trace, the taxonomy, or the citation format. A model that still produces
them has learned this behaviour in training. On each component, no trace-tuned
model varies by more than the $\pm0.034$ noise floor across the strict, runtime,
and minimal prompts. For baseline models the spread (highest minus lowest score
across the three prompts) is $0.240$--$0.398$ on Refusal F1 and
$0.620$--$0.823$ on Factual Grounding. Under the minimal prompt, baseline
grounding collapses toward zero while trace-tuned models hold steady
(Table~\ref{tab:internalization}, Appendix~\ref{app:deconfound}). The same
pattern appears in separate citation metrics (Appendix~\ref{app:citation}). Figure~\ref{fig:internal} plots all three prompts.

\subsection{Single-Truth Recall, Factual Grounding}
\label{sec:whatdoesnot}
\setcounter{dbltopnumber}{3}

Two components show no consistent gain. Single-Truth Recall (STR), the share of
responses that adopt the gold answer as their own conclusion, is higher after
fine-tuning in $7$ of $12$ configurations (sign test $p=0.77$), so
fine-tuning does not improve it consistently. Factual Grounding (FG) is averaged
over a different set of examples for each model: the $608$ answerable examples
plus every refusal example the model answered. A baseline model
that rarely refuses is therefore scored on up to $128$ examples whose
evidence is insufficient by design. When we bound FG with a like-for-like interval on only the examples where the
model answered and the gold label is answerable, the base and
fine-tuned intervals overlap in $6$ of $12$ configurations, all under the strict
or runtime prompt
(Table~\ref{tab:fgdenom}; Appendix~\ref{app:deconfound}). Trace supervision
therefore raises grounding when the instruction is sparse. Under a well-specified
prompt, a baseline model already produces competitive grounded answers but lacks
a reliable decision on whether to answer and the ability to keep its behaviour
when the prompt no longer describes the procedure. Trace supervision supplies
both, so we expect no consistent gain on these two components under
well-specified prompts. Appendix~\ref{app:deconfound} gives the per-configuration
bounds and the effect of prompt length.

\subsection{Oracle ablations: one stage at a time}
\label{sec:oracle}

Each oracle condition (\S\ref{sec:setup}) supplies the gold annotation for one
stage of the trace and leaves the rest of the pipeline unchanged, which isolates
what that stage contributes. We examine each metric separately and use the
metrics an oracle should not affect as controls.
Table~\ref{tab:oracle} and Figure~\ref{fig:oraclefig} give the per-metric effects
(Appendix~\ref{app:oracle}). \textbf{The conflict label affects only
Behaviour Adherence.} Supplying the gold Stage-2 label improves Behaviour
Adherence (mean change $+0.013$; improved in $8$ of $9$ configurations, exact
sign test $p=0.039$; Table~\ref{tab:oracle}). \textbf{The document notes affect grounding and
citation metrics.} Supplying gold Stage-1 notes improves every grounding and
citation metric, each positive in \emph{every} configuration
(Table~\ref{tab:oracle}). Behaviour Adherence also rises slightly, as expected
because correct document verdicts also constrain the conflict inference. Neither
intervention moves mean decision accuracy by more than $0.004$
(Table~\ref{tab:oracle}), and trace-tuned decision accuracy is already
$0.927$--$0.976$.
\textbf{Interpretation.} The two oracle effects add up: supplying both differs
from the sum of the separate effects by at most $0.010$ on every metric
($|\Delta_{\text{both}}-(\Delta_{\text{conflict}}+\Delta_{\text{notes}})|\leq0.010$,
Appendix~\ref{app:oracle}). This supports treating the three stages as
separable. For \emph{baseline} models the oracles lower most metrics, most of all
grounding under the conflict label ($-0.238$). This is a second result, after
prompt invariance (\S\ref{sec:internalization}), that separates fine-tuned from
baseline models.

\subsection{External prompting comparators}
\label{sec:comparators}

We compare against two prompting-only strategies, few-shot Chain-of-Thought
\citep{wei2022cot} and Chain-of-Note-style retrieval notes
\citep{yu2024chainofnote}. Both run on the untuned baseline models with the same
benchmark and evaluator. We run them on the three 7B and 8B models
(Llama-3.1-8B, Mistral-7B and Qwen2.5-7B). Both produce
usable answers (Table~\ref{tab:comparators}); Refusal F1 is $0$ for Chain-of-Note
and $0.302$--$0.525$ for few-shot Chain-of-Thought. Chain-of-Note refuses none
of the $128$ refusal examples. Few-shot Chain-of-Thought refuses $32/128$ to
$85/128$ of them, depending on the model. Structured prompting elicits
conflict-aware \emph{answers}, and its answer/refuse \emph{decision} varies by
method and model.
\textbf{Citation behaviour.} Trace supervision raises citation
\emph{sufficiency}, the share of supported claims that carry at least one
citation, by $+0.363$ on average over the twelve configurations. Citation
\emph{precision}, the share of cited documents that support the claim, changes
little on average.
Appendix~\ref{app:citation} gives the seven-metric analysis, per-model abstention
counts and Chain-of-Note's citation quality, the lowest of any method examined.

\subsection{Answer-only supervision comparison}
\label{sec:answeronly}

The comparators above change only the prompt. Here we fine-tune Llama-3.1-8B and
Qwen2.5-7B on the same 862 base training examples and with the same QLoRA
configuration as the trace-supervised runs, with the trace removed from the
supervision target (Appendix~\ref{app:answeronly}). The trace-supervised runs
add 27--48 further examples ($3$--$6\%$ more data; Appendix~\ref{app:sft}), and
data, model, and optimiser are otherwise identical. Under the minimal prompt, trace
supervision exceeds this answer-only baseline on every metric for both models,
with the largest gap on Refusal Precision, the share of the model's refusals
that were required (Table~\ref{tab:aodelta}, Appendix~\ref{app:answeronly}).

\section{Validating the CATS Judges}
\label{sec:human}

We validated the CATS judge committee against a human study. Four reviewers
produced $650$ reviews of $350$ responses: $300$ responses were reviewed by two
reviewers and $50$ by one. The responses come from two of the four models
(Llama-3.1-8B and Qwen2.5-7B), from untuned and fine-tuned versions, and cover
all five conflict types. Grounded Refusal is computed deterministically, so
the study covers the three judged components. \textbf{Factual Grounding and
Single-Truth Recall agree well with humans} ($\kappa=0.78$ and $0.79$).
\textbf{Behaviour Adherence agrees at} $\kappa=0.41$, lowest on
Misinformation and Complementary information examples; human reviewers agree
least on Complementary and Outdated examples. Appendices
\ref{app:human} and~\ref{app:allagreement} give the study design and coverage.

\section{Conclusion}
\label{sec:conclusion}

We introduced a resource, an evaluation and a training method for deciding when
to answer under conflicting evidence: 943 training/validation and 736 benchmark examples with $10{,}343$
annotated documents, the four CATS components with judges validated against human
review ($\kappa=0.78$ and $0.79$ for Factual Grounding and Single-Truth Recall),
and supervision on a three-stage think trace. Trace supervision sharply
improves the answer/\allowbreak refuse decision (under the three prompts, Refusal F1 rises from at most
$0.429$ to at least $0.826$), and the conflict-aware behaviour persists when the
prompt no longer describes it: under the 40-word prompt, untuned Refusal F1 falls
to at most $0.031$ and Factual Grounding to at most $0.108$, while trace-tuned
scores stay within the noise floor. On the answerable examples a model answers,
Factual Grounding under that prompt is at most $0.11$ for untuned and
$0.86$--$0.88$ for trace-tuned models. Base models already answer competitively under well-specified prompts, and the trace
adds the decision of when to answer. Answer-only fine-tuning on the same
examples trails trace supervision on every metric under the minimal prompt, with the largest gap on Refusal Precision
(\S\ref{sec:answeronly}). The three stages are separable, so we can
measure each on its own: gold document notes improve every grounding and citation
metric, gold conflict labels improve Behaviour Adherence, and the two effects add
up to within $0.010$ (\S\ref{sec:oracle}); neither oracle moves mean decision
accuracy by more than $0.004$.
We release the resource, the annotation and judge prompts, and the per-run outputs
(Appendix~\ref{app:annprompts}). Three directions
follow (\S\ref{sec:future}): preference training on the trace stages, a controlled
unstructured-supervision baseline, and compact traces.

\section{Limitations}
\label{sec:limitations}

\paragraph{Stage-1 verdicts.} Factual Grounding counts a claim only when a
document with a gold verdict of \textit{supports} or \textit{partially supports}
backs it. These verdicts come from the annotation committee, whose Stage-1
verdicts match the human-majority verdict on $92\%$ of 100 sampled
query--document pairs ($\kappa=0.88$; human--human $\kappa=0.83$).

\paragraph{Grounding standard.} CATS scores whether the retrieved documents
justify an answer, and the grounding criterion accepts partially supporting
documents. Grounding scores therefore bound grounding under a stricter
entailment standard from above, and every model and condition is scored under the
same criterion.

\paragraph{Benchmark scope.} The benchmark is a three-track coverage design
(Appendix~\ref{app:data}), and Appendix~\ref{app:pertrack} reports results by
track.

\paragraph{Behaviour Adherence.} Behaviour Adherence is the most demanding
component to judge: humans agree with each other at $\kappa=0.67$, the committee
agrees with humans at $\kappa=0.41$, and the three judges agree among themselves
at a nominal $\alpha=0.46$. We report behaviour results by conflict type and rely
on the consistent direction across the twelve model$\times$prompt configurations
(\S\ref{sec:human}). The refusal and grounding results use components whose
judges agree closely with humans ($\kappa\geq0.78$ for Factual Grounding and
Single-Truth Recall; Grounded Refusal needs no judge).

\section{Future Work}
\label{sec:future}

The results above already support the paper's claims. Refusal F1 separates
trace-tuned from untuned models under every prompt (\S\ref{sec:refusal},
\S\ref{sec:internalization}), the oracle ablations isolate what each stage
contributes (\S\ref{sec:oracle}), and answer-only supervision on the same
examples trails trace supervision on every metric under the minimal prompt
(\S\ref{sec:answeronly}). The extensions below add to this evidence. They reuse
the benchmark, the CATS components and the noise floor of this paper, so none of
them changes the reported results.

\paragraph{Reasoning-specific preference training.} Behaviour Adherence is the
component hardest to judge (\S\ref{sec:limitations}), so a judge-based reward
would inherit that difficulty. We plan to refine the trace-tuned models with
preference optimisation (DPO or a similar method) and rewards that need no judge:
the deterministic Grounded Refusal decision, and the agreement of each trace
stage with gold annotations that the training set already contains (Stage-1
verdicts, the Stage-2 conflict type, and citations of documents with a supporting
gold verdict). For each training example we would sample several traces from the
supervised model and rank them by these signals. Stage-specific pairs, such as two
traces that are identical except for the conflict type, aim the preference signal
at one reasoning step at a time.

\paragraph{Controlled unstructured supervision.} We plan a baseline that trains on
the same examples and the same supervision signals (document verdicts, conflict
type and final answer) without the serialised three-stage trace, together with
ablations that train on each stage alone and on pairs of stages. This design tests
whether refusal calibration and prompt robustness need the full structured trace
or follow from task-specific supervision in general. We have not run it yet
because answer-only supervision (\S\ref{sec:answeronly}) already removes the trace
from the same training examples, and together with the oracle and prompt-length
ablations it supports the paper's argument. The proposed baseline refines this
comparison.

\paragraph{Efficiency and scale.} Trace-tuned outputs are about ten times longer
than answer-only outputs for Qwen2.5-7B and Llama-3.1-8B (Appendix~\ref{app:inflatency}). We plan to test whether
compact traces keep the gains, and to repeat the study on larger models and
further model families.

\section{Reproducibility Statement}

We release the training/validation split, the held-out benchmark, the
annotation, inference and judge prompts, and the per-run evaluation outputs from
which every number in this paper is computed (Appendix~\ref{app:annprompts}). The
appendices give the QLoRA hyperparameters and checkpoint rule
(Appendix~\ref{app:sft}), the deterministic decoding protocol
(Appendix~\ref{app:inference}), the annotation aggregation rule
(Appendix~\ref{app:annotation}), the judge prompts
(Appendix~\ref{app:catsprompts}) and the retrieval and sampling protocol
(Appendix~\ref{app:data}).

\section{Ethics Statement}

\paragraph{Data provenance and licensing.} We derive the released resource from
publicly available research datasets and from live web retrieval. We preserve
source provenance in every record. The dataset cards identify the
upstream collection for each source family, so downstream users can consult
the relevant licences before redistributing source-derived material. We store
retrieved snippets with their source URL and, where available, a timestamp.
Because live retrieval changes over time, the release is a snapshot of the
retrieved documents at collection time.

\paragraph{Human annotators.} Members of the research group carried out the
human review. We retain reviewer identifiers in the review artifacts for
auditability. They carry no demographic information.

\paragraph{Intended use and risks.} We intend the resource for research on
evidence-grounded generation, abstention calibration, and conflict-aware
evaluation. First, conflict labels and answerability labels describe the
\emph{supplied} evidence, not world truth, and several of the underlying questions
concern contested scientific or medical topics. Second, the source-quality field
is a coarse label derived from each document's URL and retained as annotation
metadata. It is not a credibility ranking of publishers.

\section{Use of Language Models}

\paragraph{In the research.} Language models produced the document annotations
(\S\ref{sec:data}) and serve as the CATS judges (\S\ref{sec:cats}). Human
reviewers checked samples of both (\S\ref{sec:human}).

\paragraph{In writing.} We used AI assistants to help draft and edit the text of
this paper. The authors reviewed all of that text, checked the reported numbers
against the experiment outputs, and are responsible for the content.

\bibliography{custom}

\begin{thebibliography}{41}
\providecommand{\natexlab}[1]{#1}

\bibitem[{Borgeaud et~al.(2022)Borgeaud, Mensch, Hoffmann, Cai, Rutherford,
  Millican, van~den Driessche, Lespiau, Damoc, Clark, de~Las~Casas, Guy,
  Menick, Ring, Hennigan, Huang, Maggiore, Jones, Cassirer, Brock, Paganini,
  Irving, Vinyals, Osindero, Simonyan, Rae, Elsen, and Sifre}]{borgeaud2022}
Sebastian Borgeaud, Arthur Mensch, Jordan Hoffmann, Trevor Cai, Eliza
  Rutherford, Katie Millican, George van~den Driessche, Jean-Baptiste Lespiau,
  Bogdan Damoc, Aidan Clark, Diego de~Las~Casas, Aurelia Guy, Jacob Menick,
  Roman Ring, Tom Hennigan, Saffron Huang, Loren Maggiore, Chris Jones, Albin
  Cassirer, and 9 others. 2022.
\newblock \href {https://proceedings.mlr.press/v162/borgeaud22a.html}
  {Improving language models by retrieving from trillions of tokens}.
\newblock In \emph{Proceedings of the 39th International Conference on Machine
  Learning}, volume 162 of \emph{Proceedings of Machine Learning Research},
  pages 2206--2240. PMLR.

\bibitem[{Cattan et~al.(2025)Cattan, Jacovi, Ram, Herzig, Aharoni, Goldshtein,
  Ofek, Szpektor, and Caciularu}]{cattan2025dragged}
Arie Cattan, Alon Jacovi, Ori Ram, Jonathan Herzig, Roee Aharoni, Sasha
  Goldshtein, Eran Ofek, Idan Szpektor, and Avi Caciularu. 2025.
\newblock \href {https://arxiv.org/abs/2506.08500} {{DRAG}ged into conflicts:
  Detecting and addressing conflicting sources in search-augmented {LLM}s}.
\newblock \emph{Preprint}, arXiv:2506.08500.

\bibitem[{Dettmers et~al.(2023)Dettmers, Pagnoni, Holtzman, and
  Zettlemoyer}]{dettmers2023qloraefficientfinetuningquantized}
Tim Dettmers, Artidoro Pagnoni, Ari Holtzman, and Luke Zettlemoyer. 2023.
\newblock \href {https://doi.org/10.52202/075280-0441} {{QL}o{RA}: Efficient
  finetuning of quantized {LLM}s}.
\newblock In \emph{Advances in Neural Information Processing Systems},
  volume~36, pages 10088--10115. Curran Associates, Inc.

\bibitem[{Fan et~al.(2024)Fan, Ding, Ning, Wang, Li, Yin, Chua, and
  Li}]{fan2024surveyragmeetingllms}
Wenqi Fan, Yujuan Ding, Liangbo Ning, Shijie Wang, Hengyun Li, Dawei Yin,
  Tat-Seng Chua, and Qing Li. 2024.
\newblock \href {https://doi.org/10.1145/3637528.3671470} {A survey on {RAG}
  meeting {LLM}s: Towards retrieval-augmented large language models}.
\newblock In \emph{Proceedings of the 30th ACM SIGKDD Conference on Knowledge
  Discovery and Data Mining}, pages 6491--6501, New York, NY, USA. ACM.

\bibitem[{Fang et~al.(2024)Fang, Bai, Ni, Yang, Chen, and
  Xu}]{fang2024enhancingnoiserobustnessretrievalaugmented}
Feiteng Fang, Yuelin Bai, Shiwen Ni, Min Yang, Xiaojun Chen, and Ruifeng Xu.
  2024.
\newblock \href {https://doi.org/10.18653/v1/2024.acl-long.540} {Enhancing
  noise robustness of retrieval-augmented language models with adaptive
  adversarial training}.
\newblock In \emph{Proceedings of the 62nd Annual Meeting of the {A}ssociation
  for {C}omputational {L}inguistics (Volume 1: Long Papers)}, pages
  10028--10039, Bangkok, Thailand. Association for Computational Linguistics.

\bibitem[{Gao et~al.(2024)Gao, Xiong, Gao, Jia, Pan, Bi, Dai, Sun, Wang, and
  Wang}]{gao2024retrievalaugmentedgenerationlargelanguage}
Yunfan Gao, Yun Xiong, Xinyu Gao, Kangxiang Jia, Jinliu Pan, Yuxi Bi, Yi~Dai,
  Jiawei Sun, Meng Wang, and Haofen Wang. 2024.
\newblock \href {https://arxiv.org/abs/2312.10997} {Retrieval-augmented
  generation for large language models: A survey}.
\newblock \emph{Preprint}, arXiv:2312.10997.

\bibitem[{Guu et~al.(2020)Guu, Lee, Tung, Pasupat, and Chang}]{guu2020realm}
Kelvin Guu, Kenton Lee, Zora Tung, Panupong Pasupat, and Ming-Wei Chang. 2020.
\newblock \href {https://proceedings.mlr.press/v119/guu20a.html} {{REALM}:
  Retrieval-augmented language model pre-training}.
\newblock In \emph{Proceedings of the 37th International Conference on Machine
  Learning}, volume 119 of \emph{Proceedings of Machine Learning Research},
  pages 3929--3938. PMLR.

\bibitem[{Hofst{\"a}tter et~al.(2021)Hofst{\"a}tter, Lin, Yang, Lin, and
  Hanbury}]{hofstatter2021tasb}
Sebastian Hofst{\"a}tter, Sheng-Chieh Lin, Jheng-Hong Yang, Jimmy Lin, and
  Allan Hanbury. 2021.
\newblock \href {https://doi.org/10.1145/3404835.3462891} {Efficiently teaching
  an effective dense retriever with balanced topic aware sampling}.
\newblock In \emph{Proceedings of the 44th International ACM SIGIR Conference
  on Research and Development in Information Retrieval}, pages 113--122. ACM.

\bibitem[{Hou et~al.(2024)Hou, Pascale, Carnerero-Cano, Tchrakian, Marinescu,
  Daly, Padhi, and Sattigeri}]{hou2024wikicontradict}
Yufang Hou, Alessandra Pascale, Javier Carnerero-Cano, Tigran Tchrakian, Radu
  Marinescu, Elizabeth Daly, Inkit Padhi, and Prasanna Sattigeri. 2024.
\newblock \href {https://doi.org/10.52202/079017-3481} {{W}iki{C}ontradict: A
  benchmark for evaluating {LLM}s on real-world knowledge conflicts from
  {W}ikipedia}.
\newblock In \emph{Advances in Neural Information Processing Systems 37:
  Datasets and Benchmarks Track}, pages 109701--109747. Curran Associates, Inc.

\bibitem[{Hu et~al.(2022)Hu, Shen, Wallis, Allen-Zhu, Li, Wang, Wang, and
  Chen}]{hu2021loralowrankadaptationlarge}
Edward~J. Hu, Yelong Shen, Phillip Wallis, Zeyuan Allen-Zhu, Yuanzhi Li, Shean
  Wang, Lu~Wang, and Weizhu Chen. 2022.
\newblock \href {https://openreview.net/forum?id=nZeVKeeFYf9} {{L}o{RA}:
  Low-rank adaptation of large language models}.
\newblock In \emph{The Tenth International Conference on Learning
  Representations}.

\bibitem[{Huo et~al.(2025)Huo, Li, Qin, Qu, Li, Li, Ma, and
  Cheng}]{huo2025microact}
Nan Huo, Jinyang Li, Bowen Qin, Ge~Qu, Xiaolong Li, Xiaodong Li, Chenhao Ma,
  and Reynold Cheng. 2025.
\newblock \href {https://doi.org/10.18653/v1/2025.acl-long.909} {{M}icro-{A}ct:
  Mitigate knowledge conflict in question answering via actionable
  self-reasoning}.
\newblock In \emph{Proceedings of the 63rd Annual Meeting of the Association
  for Computational Linguistics (Volume 1: Long Papers)}, pages 18550--18574,
  Vienna, Austria. Association for Computational Linguistics.

\bibitem[{Izacard et~al.(2023)Izacard, Lewis, Lomeli, Hosseini, Petroni,
  Schick, Dwivedi-Yu, Joulin, Riedel, and Grave}]{izacard2023atlas}
Gautier Izacard, Patrick Lewis, Maria Lomeli, Lucas Hosseini, Fabio Petroni,
  Timo Schick, Jane Dwivedi-Yu, Armand Joulin, Sebastian Riedel, and Edouard
  Grave. 2023.
\newblock \href {https://jmlr.org/papers/v24/23-0037.html} {Atlas: Few-shot
  learning with retrieval augmented language models}.
\newblock \emph{Journal of Machine Learning Research}, 24(251):1--43.

\bibitem[{Jain et~al.(2024)Jain, Chiang, Wen, Kirchenbauer, Chu, Somepalli,
  Bartoldson, Kailkhura, Schwarzschild, Saha, Goldblum, Geiping, and
  Goldstein}]{jain2023neftune}
Neel Jain, Ping-yeh Chiang, Yuxin Wen, John Kirchenbauer, Hong-Min Chu,
  Gowthami Somepalli, Brian~R. Bartoldson, Bhavya Kailkhura, Avi Schwarzschild,
  Aniruddha Saha, Micah Goldblum, Jonas Geiping, and Tom Goldstein. 2024.
\newblock \href {https://openreview.net/forum?id=0bMmZ3fkCk} {{NEFT}une: Noisy
  embeddings improve instruction finetuning}.
\newblock In \emph{The Twelfth International Conference on Learning
  Representations}, pages 17566--17591.

\bibitem[{Jiang et~al.(2023)Jiang, Sablayrolles, Mensch, Bamford, Chaplot,
  de~las Casas, Bressand, Lengyel, Lample, Saulnier, Lavaud, Lachaux, Stock,
  Scao, Lavril, Wang, Lacroix, and Sayed}]{jiang2023mistral7b}
Albert~Q. Jiang, Alexandre Sablayrolles, Arthur Mensch, Chris Bamford,
  Devendra~Singh Chaplot, Diego de~las Casas, Florian Bressand, Gianna Lengyel,
  Guillaume Lample, Lucile Saulnier, L{\'e}lio~Renard Lavaud, Marie-Anne
  Lachaux, Pierre Stock, Teven~Le Scao, Thibaut Lavril, Thomas Wang,
  Timoth{\'e}e Lacroix, and William~El Sayed. 2023.
\newblock \href {https://arxiv.org/abs/2310.06825} {{Mistral 7B}}.
\newblock \emph{Preprint}, arXiv:2310.06825.

\bibitem[{Jiayang et~al.(2024)Jiayang, Chan, Zhuang, Qiu, Zhang, Liu, Song,
  Zhang, Liu, and Zhang}]{jiayang2024econ}
Cheng Jiayang, Chunkit Chan, Qianqian Zhuang, Lin Qiu, Tianhang Zhang, Tengxiao
  Liu, Yangqiu Song, Yue Zhang, Pengfei Liu, and Zheng Zhang. 2024.
\newblock \href {https://doi.org/10.18653/v1/2024.emnlp-main.447} {{ECON}: On
  the detection and resolution of evidence conflicts}.
\newblock In \emph{Proceedings of the 2024 Conference on Empirical Methods in
  Natural Language Processing}, pages 7816--7844, Miami, Florida, USA.
  Association for Computational Linguistics.

\bibitem[{Lewis et~al.(2020)Lewis, Perez, Piktus, Petroni, Karpukhin, Goyal,
  K{\"u}ttler, Lewis, Yih, Rockt{\"a}schel, Riedel, and Kiela}]{lewis2020rag}
Patrick Lewis, Ethan Perez, Aleksandra Piktus, Fabio Petroni, Vladimir
  Karpukhin, Naman Goyal, Heinrich K{\"u}ttler, Mike Lewis, Wen-tau Yih, Tim
  Rockt{\"a}schel, Sebastian Riedel, and Douwe Kiela. 2020.
\newblock \href
  {https://proceedings.neurips.cc/paper_files/paper/2020/file/6b493230205f780e1bc26945df7481e5-Paper.pdf}
  {Retrieval-augmented generation for knowledge-intensive {NLP} tasks}.
\newblock In \emph{Advances in Neural Information Processing Systems},
  volume~33, pages 9459--9474. Curran Associates, Inc.

\bibitem[{Liu et~al.(2025)Liu, Ning, Halder, Qi, Xiao, Htut, Zhang, John, Min,
  Benajiba, and Roth}]{liu2025opendomainquestionanswering}
Siyi Liu, Qiang Ning, Kishaloy Halder, Zheng Qi, Wei Xiao, Phu~Mon Htut,
  Yi~Zhang, Neha~Anna John, Bonan Min, Yassine Benajiba, and Dan Roth. 2025.
\newblock \href {https://doi.org/10.18653/v1/2025.findings-naacl.99} {Open
  domain question answering with conflicting contexts}.
\newblock In \emph{Findings of the Association for Computational Linguistics:
  NAACL 2025}, pages 1838--1854, Albuquerque, New Mexico. Association for
  Computational Linguistics.

\bibitem[{Liu et~al.(2026)Liu, Zhang, Guo, Cao, Yuan, Sun, Liu, Hong, and
  Ma}]{liu2026craft}
Yu~Liu, Wenxiao Zhang, Diandian Guo, Cong Cao, Fangfang Yuan, Qiang Sun,
  Yanbing Liu, Jin~B. Hong, and Zhiyuan Ma. 2026.
\newblock \href {https://arxiv.org/abs/2602.01348} {Does faithfulness-guided
  alignment hurt accuracy? {U}nlocking accurate and faithful post-retrieval
  reasoning}.
\newblock \emph{Preprint}, arXiv:2602.01348.

\bibitem[{{Llama Team, AI at Meta}(2024)}]{dubey2024llama3}
{Llama Team, AI at Meta}. 2024.
\newblock \href {https://arxiv.org/abs/2407.21783} {The {Llama} 3 herd of
  models}.
\newblock \emph{Preprint}, arXiv:2407.21783.

\bibitem[{Ming et~al.(2025)Ming, Purushwalkam, Pandit, Ke, Nguyen, Xiong, and
  Joty}]{ming2025faitheval}
Yifei Ming, Senthil Purushwalkam, Shrey Pandit, Zixuan Ke, Xuan-Phi Nguyen,
  Caiming Xiong, and Shafiq Joty. 2025.
\newblock \href
  {https://proceedings.iclr.cc/paper_files/paper/2025/file/48404cd9ce03946c6b7177691f3267a1-Paper-Conference.pdf}
  {{F}aith{E}val: Can your language model stay faithful to context, even if
  ``{The Moon is Made of Marshmallows}''}.
\newblock In \emph{The Thirteenth International Conference on Learning
  Representations}, pages 29430--29456.

\bibitem[{Muhamed et~al.(2026)Muhamed, Ribeiro, Dreyer, Smith, and
  Diab}]{muhamed2025refusalbench}
Aashiq Muhamed, Leonardo F.~R. Ribeiro, Markus Dreyer, Virginia Smith, and
  Mona~T. Diab. 2026.
\newblock \href {https://doi.org/10.18653/v1/2026.eacl-long.321}
  {{R}efusal{B}ench: Generative evaluation of selective refusal in grounded
  language models}.
\newblock In \emph{Proceedings of the 19th Conference of the European Chapter
  of the Association for Computational Linguistics (Volume 1: Long Papers)},
  pages 6811--6856, Rabat, Morocco. Association for Computational Linguistics.

\bibitem[{{Qwen Team}(2024)}]{yang2024qwen25}
{Qwen Team}. 2024.
\newblock \href {https://arxiv.org/abs/2412.15115} {Qwen2.5 technical report}.
\newblock \emph{Preprint}, arXiv:2412.15115.

\bibitem[{Ram et~al.(2023)Ram, Levine, Dalmedigos, Muhlgay, Shashua,
  Leyton-Brown, and Shoham}]{ram-etal-2023-context}
Ori Ram, Yoav Levine, Itay Dalmedigos, Dor Muhlgay, Amnon Shashua, Kevin
  Leyton-Brown, and Yoav Shoham. 2023.
\newblock \href {https://doi.org/10.1162/tacl_a_00605} {In-context
  retrieval-augmented language models}.
\newblock \emph{Transactions of the Association for Computational Linguistics},
  11:1316--1331.

\bibitem[{Shi et~al.(2023)Shi, Chen, Misra, Scales, Dohan, Chi, Sch{\"a}rli,
  and Zhou}]{shi2023largelanguagemodelseasily}
Freda Shi, Xinyun Chen, Kanishka Misra, Nathan Scales, David Dohan, Ed~H. Chi,
  Nathanael Sch{\"a}rli, and Denny Zhou. 2023.
\newblock \href {https://proceedings.mlr.press/v202/shi23a.html} {Large
  language models can be easily distracted by irrelevant context}.
\newblock In \emph{Proceedings of the 40th International Conference on Machine
  Learning}, volume 202 of \emph{Proceedings of Machine Learning Research},
  pages 31210--31227. PMLR.

\bibitem[{Song et~al.(2025)Song, Sim, Bhardwaj, Chieu, Majumder, and
  Poria}]{song2025trustalign}
Maojia Song, Shang~Hong Sim, Rishabh Bhardwaj, Hai~Leong Chieu, Navonil
  Majumder, and Soujanya Poria. 2025.
\newblock \href
  {https://proceedings.iclr.cc/paper_files/paper/2025/file/4c88827decab6c046b881a2c3a99c76f-Paper-Conference.pdf}
  {Measuring and enhancing trustworthiness of {LLM}s in {RAG} through grounded
  attributions and learning to refuse}.
\newblock In \emph{The Thirteenth International Conference on Learning
  Representations}, pages 30783--30823.
\newblock {arXiv}:2409.11242.

\bibitem[{Su et~al.(2024)Su, Zhang, Qu, Zhu, Li, Sun, Li, Zhang, and
  Cheng}]{su2024conflictbankbenchmarkevaluatinginfluence}
Zhaochen Su, Jun Zhang, Xiaoye Qu, Tong Zhu, Yanshu Li, Jiashuo Sun, Juntao Li,
  Min Zhang, and Yu~Cheng. 2024.
\newblock \href {https://doi.org/10.52202/079017-3280} {{ConflictBank}: A
  benchmark for evaluating the influence of knowledge conflicts in {LLM}s}.
\newblock In \emph{Advances in Neural Information Processing Systems 37:
  Datasets and Benchmarks Track}, pages 103242--103268. Curran Associates, Inc.

\bibitem[{Vu et~al.(2024)Vu, Iyyer, Wang, Constant, Wei, Wei, Tar, Sung, Zhou,
  Le, and Luong}]{vu2024freshllms}
Tu~Vu, Mohit Iyyer, Xuezhi Wang, Noah Constant, Jerry Wei, Jason Wei, Chris
  Tar, Yun-Hsuan Sung, Denny Zhou, Quoc Le, and Thang Luong. 2024.
\newblock \href {https://doi.org/10.18653/v1/2024.findings-acl.813}
  {{F}resh{LLM}s: Refreshing large language models with search engine
  augmentation}.
\newblock In \emph{Findings of the Association for Computational Linguistics:
  ACL 2024}, pages 13697--13720, Bangkok, Thailand. Association for
  Computational Linguistics.

\bibitem[{Wan et~al.(2024)Wan, Wallace, and
  Klein}]{wan2024evidencelanguagemodelsconvincing}
Alexander Wan, Eric Wallace, and Dan Klein. 2024.
\newblock \href {https://doi.org/10.18653/v1/2024.acl-long.403} {What evidence
  do language models find convincing?}
\newblock In \emph{Proceedings of the 62nd Annual Meeting of the Association
  for Computational Linguistics (Volume 1: Long Papers)}, pages 7468--7484,
  Bangkok, Thailand. Association for Computational Linguistics.

\bibitem[{Wang et~al.(2026)Wang, Li, Liu, and Shu}]{wang2026conflictrag}
Chenyu Wang, Yueyuan Li, Yingmin Liu, and Yang Shu. 2026.
\newblock \href {https://arxiv.org/abs/2605.17301} {{C}onflict{RAG}: Detecting
  and resolving knowledge conflicts in retrieval augmented generation}.
\newblock \emph{Preprint}, arXiv:2605.17301.

\bibitem[{Wang et~al.(2025)Wang, Prasad, Stengel-Eskin, and
  Bansal}]{wang2025conflictingevidence}
Han Wang, Archiki Prasad, Elias Stengel-Eskin, and Mohit Bansal. 2025.
\newblock \href {https://openreview.net/forum?id=z1MHB2m3V9}
  {Retrieval-augmented generation with conflicting evidence}.
\newblock In \emph{Second Conference on Language Modeling}.
\newblock {arXiv}:2504.13079.

\bibitem[{Wei et~al.(2022)Wei, Wang, Schuurmans, Bosma, Ichter, Xia, Chi, Le,
  and Zhou}]{wei2022cot}
Jason Wei, Xuezhi Wang, Dale Schuurmans, Maarten Bosma, Brian Ichter, Fei Xia,
  Ed~Chi, Quoc~V. Le, and Denny Zhou. 2022.
\newblock \href {https://doi.org/10.52202/068431-1800} {Chain-of-thought
  prompting elicits reasoning in large language models}.
\newblock In \emph{Advances in Neural Information Processing Systems},
  volume~35, pages 24824--24837. Curran Associates, Inc.

\bibitem[{Wu et~al.(2026)Wu, Wang, Sun, Lu, Zhang, and
  Chen}]{wu2026conflictaware}
Haiyan Wu, Chenchen Wang, Chaoqun Sun, Chengxiong Lu, Zhiqiang Zhang, and
  Yanhong Chen. 2026.
\newblock \href {https://doi.org/10.1145/3774904.3792289} {Conflict-aware
  {RAG}: Multi-stage learning with conflict signals for robust
  retrieval-augmented generation}.
\newblock In \emph{Proceedings of the ACM Web Conference 2026}, pages
  2114--2125, Dubai, United Arab Emirates. ACM.

\bibitem[{Xu et~al.(2024)Xu, Qi, Guo, Wang, Wang, Zhang, and
  Xu}]{xu2024knowledgeconflictsllmssurvey}
Rongwu Xu, Zehan Qi, Zhijiang Guo, Cunxiang Wang, Hongru Wang, Yue Zhang, and
  Wei Xu. 2024.
\newblock \href {https://doi.org/10.18653/v1/2024.emnlp-main.486} {Knowledge
  conflicts for {LLM}s: A survey}.
\newblock In \emph{Proceedings of the 2024 Conference on Empirical Methods in
  Natural Language Processing}, pages 8541--8565, Miami, Florida, USA.
  Association for Computational Linguistics.

\bibitem[{Xu et~al.(2025)Xu, Gao, Yu, Bi, Shen, and
  Cheng}]{xu2024aliiceevaluatingpositionalfinegrained}
Yilong Xu, Jinhua Gao, Xiaoming Yu, Baolong Bi, Huawei Shen, and Xueqi Cheng.
  2025.
\newblock \href {https://doi.org/10.18653/v1/2025.naacl-long.23} {{ALiiCE}:
  Evaluating positional fine-grained citation generation}.
\newblock In \emph{Proceedings of the 2025 Conference of the Nations of the
  {A}mericas Chapter of the {A}ssociation for {C}omputational {L}inguistics:
  Human Language Technologies (Volume 1: Long Papers)}, pages 545--561,
  Albuquerque, New Mexico. Association for Computational Linguistics.

\bibitem[{Yang et~al.(2018)Yang, Qi, Zhang, Bengio, Cohen, Salakhutdinov, and
  Manning}]{yang2018hotpotqa}
Zhilin Yang, Peng Qi, Saizheng Zhang, Yoshua Bengio, William~W. Cohen, Ruslan
  Salakhutdinov, and Christopher~D. Manning. 2018.
\newblock \href {https://doi.org/10.18653/v1/D18-1259} {{H}otpot{QA}: A dataset
  for diverse, explainable multi-hop question answering}.
\newblock In \emph{Proceedings of the 2018 Conference on Empirical Methods in
  Natural Language Processing}, pages 2369--2380, Brussels, Belgium.
  Association for Computational Linguistics.

\bibitem[{Ye et~al.(2026)Ye, Chen, Zhong, Xiao, Zhang, Wu, and
  Shen}]{ye2026seeingthrough}
Hua Ye, Siyuan Chen, Ziqi Zhong, Canran Xiao, Haoliang Zhang, Yuhan Wu, and Fei
  Shen. 2026.
\newblock \href {https://doi.org/10.1609/aaai.v40i40.40740} {Seeing through the
  conflict: Transparent knowledge conflict handling in retrieval-augmented
  generation}.
\newblock In \emph{Proceedings of the AAAI Conference on Artificial
  Intelligence}, volume~40, pages 34423--34431.

\bibitem[{Yoran et~al.(2024)Yoran, Wolfson, Ram, and
  Berant}]{yoran2024makingretrievalaugmentedlanguagemodels}
Ori Yoran, Tomer Wolfson, Ori Ram, and Jonathan Berant. 2024.
\newblock \href
  {https://proceedings.iclr.cc/paper_files/paper/2024/file/8011b23e1dc3f57e1b6211ccad498919-Paper-Conference.pdf}
  {Making retrieval-augmented language models robust to irrelevant context}.
\newblock In \emph{The Twelfth International Conference on Learning
  Representations}, pages 29862--29883.

\bibitem[{Yu et~al.(2024)Yu, Zhang, Pan, Cao, Ma, Li, Wang, and
  Yu}]{yu2024chainofnote}
Wenhao Yu, Hongming Zhang, Xiaoman Pan, Peixin Cao, Kaixin Ma, Jian Li, Hongwei
  Wang, and Dong Yu. 2024.
\newblock \href {https://doi.org/10.18653/v1/2024.emnlp-main.813}
  {{C}hain-of-{N}ote: Enhancing robustness in retrieval-augmented language
  models}.
\newblock In \emph{Proceedings of the 2024 Conference on Empirical Methods in
  Natural Language Processing}, pages 14672--14685, Miami, Florida, USA.
  Association for Computational Linguistics.

\bibitem[{Zhang and
  Choi(2021)}]{zhang2021situatedqaincorporatingextralinguisticcontexts}
Michael J.~Q. Zhang and Eunsol Choi. 2021.
\newblock \href {https://doi.org/10.18653/v1/2021.emnlp-main.586}
  {{S}ituated{QA}: Incorporating extra-linguistic contexts into {QA}}.
\newblock In \emph{Proceedings of the 2021 Conference on Empirical Methods in
  Natural Language Processing}, pages 7371--7387, Online and Punta Cana,
  Dominican Republic. Association for Computational Linguistics.

\bibitem[{Zhang et~al.(2025)Zhang, Xiang, Xiao, Wang, Li, Wang, and
  Su}]{zhang2025faithfulrag}
Qinggang Zhang, Zhishang Xiang, Yilin Xiao, Le~Wang, Junhui Li, Xinrun Wang,
  and Jinsong Su. 2025.
\newblock \href {https://doi.org/10.18653/v1/2025.acl-long.1062}
  {{F}aithful{RAG}: Fact-level conflict modeling for context-faithful
  retrieval-augmented generation}.
\newblock In \emph{Proceedings of the 63rd Annual Meeting of the Association
  for Computational Linguistics (Volume 1: Long Papers)}, pages 21863--21882,
  Vienna, Austria. Association for Computational Linguistics.

\bibitem[{Zheng et~al.(2023)Zheng, Chiang, Sheng, Zhuang, Wu, Zhuang, Lin, Li,
  Li, Xing, Zhang, Gonzalez, and Stoica}]{zheng2023judging}
Lianmin Zheng, Wei-Lin Chiang, Ying Sheng, Siyuan Zhuang, Zhanghao Wu, Yonghao
  Zhuang, Zi~Lin, Zhuohan Li, Dacheng Li, Eric~P. Xing, Hao Zhang, Joseph~E.
  Gonzalez, and Ion Stoica. 2023.
\newblock \href {https://doi.org/10.52202/075280-2020} {Judging
  {LLM-as-a-Judge} with {MT-Bench} and {Chatbot Arena}}.
\newblock In \emph{Advances in Neural Information Processing Systems 36:
  Datasets and Benchmarks Track}, pages 46595--46623. Curran Associates, Inc.

\end{thebibliography}

\appendix
\raggedbottom

\section*{Appendix}

\section{Resource Construction}
\label{app:data}

\subsection{Benchmark sources and tracks}

The 736-example benchmark has three tracks.
\textbf{(1) Track A: web-retrieved (508).} Queries are drawn from ConflictingQA
\citep{wan2024evidencelanguagemodelsconvincing}, SituatedQA
\citep{zhang2021situatedqaincorporatingextralinguisticcontexts}, FreshQA
\citep{vu2024freshllms} and QACC \citep{liu2025opendomainquestionanswering}, with
every query overlapping the training pool excluded. Documents are retrieved live,
and each is represented by a snippet, a query-focused text window selected with a
TAS-B dense bi-encoder \citep{hofstatter2021tasb}. Appendix~\ref{app:funnel} gives the
full funnel and the five-document sampling rule. \textbf{(2) Track B: refusal (128).} Refusal examples
are derived from Trust-Align
\citep{song2025trustalign}. \textbf{(3) Track C: curated conflict (100).} Open-web
retrieval cannot reliably surface every conflict type, such as a fact whose
truth value has since changed. We therefore add 78 examples built from
dated Wikipedia revision pairs, following \citet{hou2024wikicontradict} in
sourcing real conflicts from Wikipedia,
19 multi-hop contexts from HotpotQA \citep{yang2018hotpotqa} and three
medical/misinformation cases. The track targets specific conflict types: all 78
revision-pair examples are outdated-information cases, and all 19 HotpotQA
examples are complementary-information cases (Table~\ref{tab:tracks}).

\begin{table}[htbp]
\centering
{\tabstyle\setlength{\tabcolsep}{2.5pt}
\begin{tabular*}{\linewidth}[t]{@{\extracolsep{\fill}}ll r rrrrr@{}}\toprule
\textbf{Source} & \textbf{Track} & $n$ & \textbf{NC} & \textbf{CI} & \textbf{CO} & \textbf{OI} & \textbf{MI}\\\midrule
QACC & web-retr. & 145 & 69 & 46 & 14 & 7 & 9\\
SituatedQA & web-retr. & 141 & 43 & 50 & 6 & 32 & 10\\
ConflictingQA & web-retr. & 137 & 8 & 53 & 73 & 1 & 2\\
TRUST-ALIGN & refusal & 128 & 57 & 45 & 13 & 13 & 0\\
FreshQA & web-retr. & 85 & 33 & 8 & 2 & 27 & 15\\
WikiRevision & curated & 78 & 0 & 0 & 0 & 78 & 0\\
HotpotQA & curated & 19 & 0 & 19 & 0 & 0 & 0\\
HealthContradict & curated & 2 & 1 & 0 & 1 & 0 & 0\\
Misinformation & curated & 1 & 0 & 0 & 0 & 0 & 1\\
\midrule
\textbf{Total} & & \textbf{736} & \textbf{211} & \textbf{221} & \textbf{109} & \textbf{158} & \textbf{37}\\
\bottomrule\end{tabular*}}

\caption{\textbf{Benchmark composition by source and construction track.} NC = No conflict, CI = Complementary information, CO = Conflicting opinions/research outcomes, OI = Conflict due to outdated information, MI = Conflict due to misinformation. Track~A is web-retrieved, Track~B is the held-out refusal pool and Track~C is curated. All 78 WikiRevision items are outdated-information cases, which make up 49\% of the OI cases in the benchmark, and all 19 HotpotQA items are complementary-information cases.}
\label{tab:tracks}
\end{table}

\begin{table}[htbp]
\centering
{\tabstyle\setlength{\tabcolsep}{3pt}
\begin{tabular*}{\linewidth}[t]{@{\extracolsep{\fill}}>{\raggedright\arraybackslash}p{0.67\linewidth}r@{}}\toprule
\textbf{Stage} & \textbf{Records}\\\midrule
Raw candidate rows from four source pools & 20{,}330\\
$-$ normalised overlaps with the training question set & $-598$\\
$-$ duplicate normalised queries & $-8{,}891$\\
Unique usable candidates & 10{,}841\\
Source-balanced selection (seeded) & 2{,}000\\
Retained with exactly ten documents & 1{,}878 (93.9\%)\\
\quad underfilled, excluded & 122\\
Human first-pass reviewed (canonical cleaned corpus) & 1{,}221\\
Selected non-refusal set & 800\\
Consolidated preselection/review population & 1{,}454\\
Final answerable benchmark records & 608\\\bottomrule
\end{tabular*}}
\caption{Construction funnel for the non-refusal benchmark records, with the
number of records at each stage.}
\label{tab:funnel}
\end{table}

Table~\ref{tab:tracks} breaks the benchmark down by source and track. Track~A
records carry live retrieval URLs across $1{,}423$ distinct domains, and 503 of
the 508 records have exactly five documents (the other 5 have six to eight).
Track~B records have five documents each and keep the URL format of the upstream
refusal resource. Each Wikipedia-revision item in Track~C has four documents from
a single domain, and each multi-hop item has ten documents without URLs.

\paragraph{Reading the tracks.} The tracks give the benchmark coverage of both
answer and refuse decisions and of all five conflict types, and we make no claim
about natural prevalence. Each CATS component applies to a defined subset:
Refusal F1 uses all $736$ examples, Behaviour Adherence and Factual Grounding use
every example except correct refusals, and Single-Truth Recall uses its $487$
single-truth examples. The refusal
labels are backed by document-level evidence: the annotation committee's Stage-1
verdicts, produced without the abstention label, mark \textit{supports} on $3.6\%$
of refusal-example documents (Appendix~\ref{app:relverif}), and two humans reviewed
all $128$ examples (Appendix~\ref{app:refusalquality}). Untuned models receive the same documents and,
under the strict prompt, an explicit statement of the refusal rule, yet reach at
most $0.429$ Refusal F1. Appendix~\ref{app:pertrack} reports results by track, and the
benchmark identifiers carry a source prefix, so others can do the same.

\subsection{Training set composition}

The training release holds 943 examples (862 train, 81 validation). Of these, 649 are answerable (590 train, 59 validation)
and 294 are refusal examples (272 train, 22 validation), whose gold response is
\texttt{CANNOT ANSWER, INSUFFICIENT EVIDENCE}. The examples come from the
\textsc{Conflicts} collection \citep{cattan2025dragged}, from held-out refusal
material derived from Trust-Align \citep{song2025trustalign}, and from the
benchmark construction pool (candidate examples that were not admitted to the
final benchmark). All 943 identifiers and queries were checked against the released
benchmark. The intersection is empty.

\subsection{Retrieval funnel}
\label{app:funnel}

Table~\ref{tab:funnel} lists the stages of the non-refusal track. Retrieval
requests twenty results per query and retains at most ten, each with a snippet
selected by a TAS-B dense bi-encoder \citep{hofstatter2021tasb}. The released
Track~A records typically carry a deterministic five-document view of those ten,
sampled by a seeded rule that takes two from the top five and three from the
bottom five. Track~B records also have five documents and Track~C records have
four or ten, so Table~\ref{tab:data} reports a mean of $5.03$ documents per
benchmark record. The five-document view keeps enough evidence to answer the query
and draws documents from both halves of the ranking, so a model cannot succeed by
reading only the top-ranked documents.

The $1{,}878$ candidates with exactly ten documents were assigned to seven
reviewers for first-pass review (Appendix~\ref{app:human-review}), and the cleaned
review corpus contains $1{,}221$ unique reviewed records. The preselection
review covers a consolidated population of $1{,}454$ records. A documented
selection rule over its review fields (acceptance, reviewer confidence,
retrieval quality, evidence sufficiency, conflict clarity, query specificity,
source reliability, relevant-document count and gold-answer feasibility) yields the
608 answerable benchmark records. The refusal track reduces from a 200-record
curated subset to the 128 released records, and all 128 passed a separate
refusal-quality review with complete agreement.

\subsection{Verification of the released files}
\label{app:relverif}

We computed all statistics below directly from the released JSONL files
(Table~\ref{tab:relverif}; Figures~\ref{fig:datacomp} and~\ref{fig:dataqual}).
Quotes are exact contiguous spans of their snippet for $82.6\%$ of train+val
documents and $81.1\%$ of benchmark documents. Counting quotes that join snippet
spans with an ellipsis or match them in part, $95.6\%$ and $95.9\%$ are anchored
to the snippet.

\begin{table*}[t]\centering\tabstyle
\begin{tabular*}{\textwidth}{@{\extracolsep{\fill}}lrr@{}}\toprule
\textbf{Property} & \textbf{Train+val} & \textbf{Benchmark}\\\midrule
Records & 943 & 736\\
Duplicate identifiers & 0 & 0\\
Identifier overlap (train/val $\cap$ benchmark) & \multicolumn{2}{c}{0}\\
Document/annotation identifier misalignments & 0 & 0\\
Records with documents ordered \texttt{d1..dN} & 943/943 & 736/736\\
Key fact on every non-irrelevant document & 5{,}886/5{,}886 & 3{,}393/3{,}393\\
Irrelevant documents carrying a key fact & 0 & 0\\
Documents retaining an annotator vote tally & 6{,}642/6{,}642 & 3{,}701/3{,}701\\
Unique source domains & 2{,}633 & 1{,}431\\
Mean snippet length (words) & 231 & 222\\
Quote is an exact contiguous span & 82.6\% & 81.1\%\\
Quote is snippet-anchored (exact/ellipsis/partial) & 95.6\% & 95.9\%\\
\bottomrule\end{tabular*}
\caption{Structural and quality checks on the released files.}
\label{tab:relverif}
\end{table*}

\paragraph{Validity of the refusal track.} The Stage-1 prompt of the annotation
committee (the panel of LLM annotators of Appendix~\ref{app:annotation}) contains
the query and one document, with its URL, snippet and timestamp, and no
answerability or abstention label. The committee marked \textit{supports} on
$3.6\%$ of the documents in the $128$ refusal examples ($23$ of $640$) against
$39.3\%$ of the documents in the $608$ answerable benchmark examples ($1{,}202$ of
$3{,}061$), an eleven-fold difference that indicates that refusal examples have
insufficient evidence.

\begin{figure*}[t]
\centering
\begin{minipage}[b]{2.9in}
\centering
\includegraphics[height=4.2in]{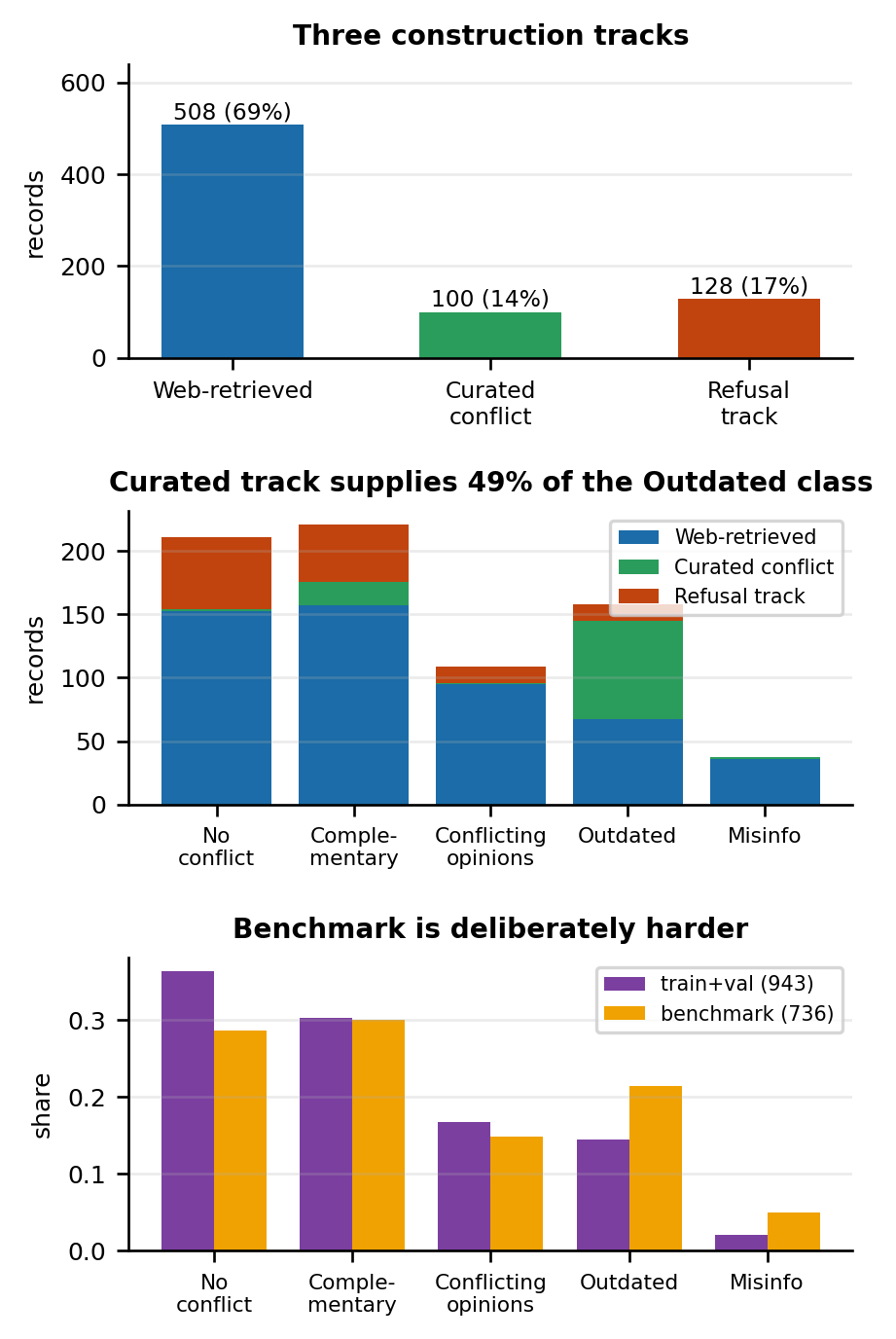}
\end{minipage}
\hfill
\begin{minipage}[b]{3.2in}
\centering
\includegraphics[height=4.2in]{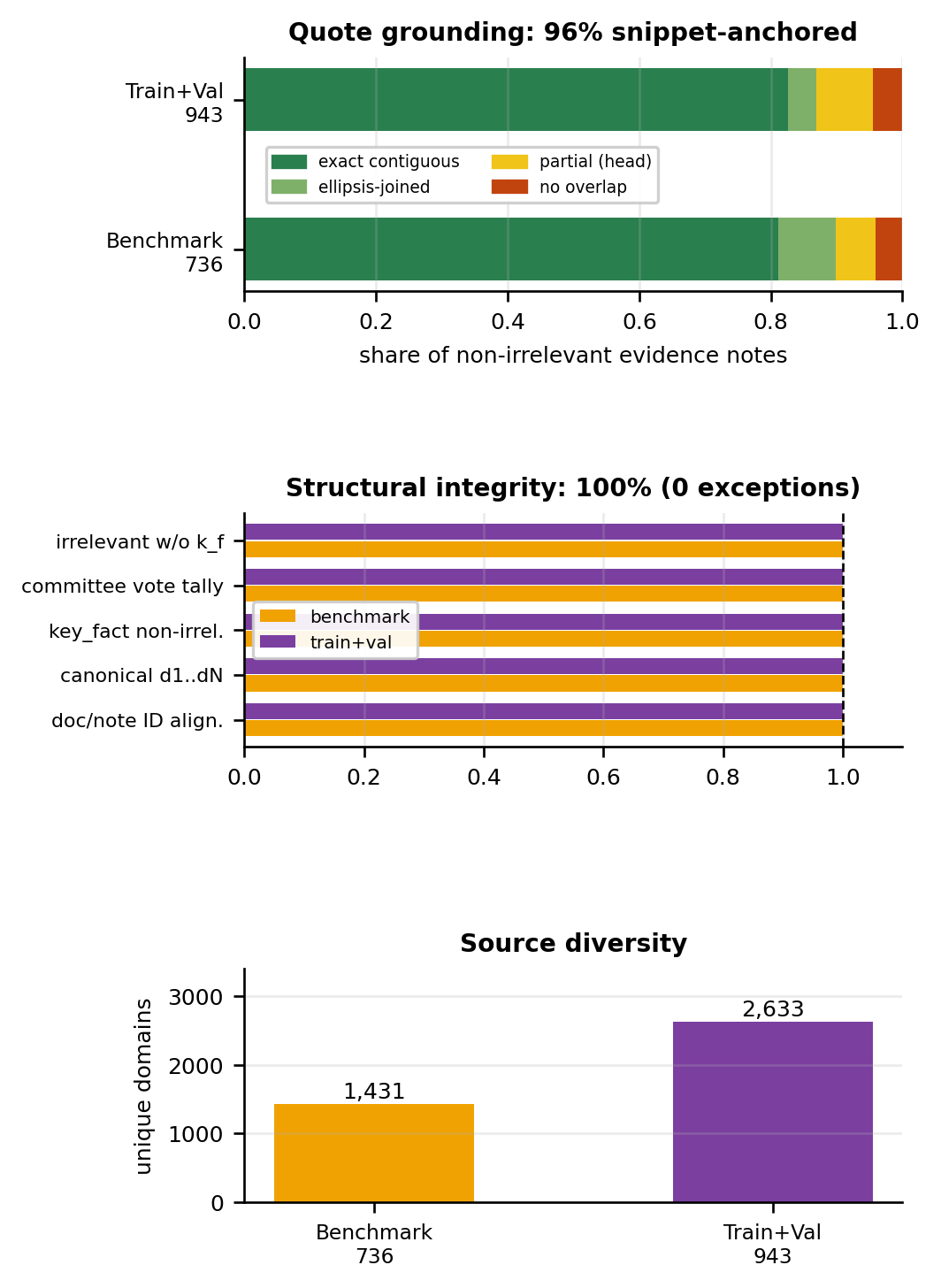}
\end{minipage}
\dualcaption{figure}{fig:datacomp}{fig:dataqual}%
{Resource composition. Top: records in each of the three benchmark construction
tracks. Middle: conflict-type counts within each track; the curated track
supplies $49\%$ of the outdated-information cases. Bottom: share of each conflict
type in the training and benchmark sets; the benchmark has fewer no-conflict and
more outdated and misinformation cases}%
{Quality audit of the released files. Top: how each supporting quote relates to
its snippet (exact contiguous span, spans joined by an ellipsis, partial match,
or no overlap). Middle: five structural checks. Bottom: number of unique source
domains.}
\end{figure*}

\subsection{Results by track}
\label{app:pertrack}

Table~\ref{tab:pertrack_dec} splits the refusal rate by benchmark track. Trace-tuned
models refuse $89.8\%$--$100\%$ of the Track~B refusal examples and $1.0\%$--$12.0\%$
of the answerable examples in Tracks~A and~C ($2.8\%$--$8.3\%$ in Track~A,
$1.0\%$--$12.0\%$ in Track~C). Untuned models refuse $0.0\%$--$63.3\%$ of the
Track~B examples.

\begin{table*}[t]\centering
{\tabstyle\setlength{\tabcolsep}{5pt}
\begin{tabular*}{\textwidth}[t]{@{\extracolsep{\fill}}ll rr rr rr@{}}\toprule
& & \multicolumn{2}{c}{\textbf{Track B (refusal, $n=128$)}} & \multicolumn{2}{c}{\textbf{Track A (web, $n=508$)}} & \multicolumn{2}{c}{\textbf{Track C (curated, $n=100$)}}\\
\cmidrule(lr){3-4}\cmidrule(lr){5-6}\cmidrule(lr){7-8}
\textbf{Model} & \textbf{Prompt} & base & \textsc{sft} & base & \textsc{sft} & base & \textsc{sft}\\\midrule
Qwen2.5-7B & strict & 0.633 & 0.961 & 0.295 & 0.033 & 0.190 & 0.020\\
 & runtime & 0.352 & 0.945 & 0.136 & 0.030 & 0.070 & 0.030\\
 & minimal & 0.016 & 0.898 & 0.002 & 0.033 & 0.000 & 0.010\\
\addlinespace[2pt]
Qwen2.5-32B & strict & 0.242 & 1.000 & 0.020 & 0.051 & 0.000 & 0.010\\
 & runtime & 0.266 & 1.000 & 0.024 & 0.053 & 0.010 & 0.010\\
 & minimal & 0.008 & 1.000 & 0.002 & 0.045 & 0.000 & 0.020\\
\addlinespace[2pt]
Llama-3.1-8B & strict & 0.234 & 0.984 & 0.134 & 0.028 & 0.090 & 0.040\\
 & runtime & 0.016 & 0.992 & 0.004 & 0.030 & 0.000 & 0.030\\
 & minimal & 0.008 & 0.992 & 0.004 & 0.028 & 0.000 & 0.030\\
\addlinespace[2pt]
Mistral-7B & strict & 0.242 & 1.000 & 0.022 & 0.083 & 0.060 & 0.120\\
 & runtime & 0.125 & 1.000 & 0.026 & 0.081 & 0.030 & 0.110\\
 & minimal & 0.000 & 1.000 & 0.004 & 0.077 & 0.000 & 0.100\\
\bottomrule\end{tabular*}}

\caption{\textbf{Refusal rate by benchmark track.} Track~B holds the 128 refusal
examples, so its rate is refusal recall. Tracks~A (web) and~C (curated) hold
answerable examples, so their rates are false refusals. \textit{base} is the
untuned model and \textsc{sft} the trace-tuned model.}
\label{tab:pertrack_dec}
\end{table*}

Within the answerable examples, the refusals of trace-tuned models concentrate
where the annotations show less support (Table~\ref{tab:pertrack_checks}). In all
twelve configurations the $17$--$54$ refused examples have a mean supports share of
$0.122$--$0.201$, against $0.412$--$0.425$ for the answered examples, and the
supports share separates the two groups with an AUC of $0.734$--$0.821$ (every
lower $95\%$ bound is above $0.5$). The models' Stage-1 document verdicts match
the annotation committee's at an accuracy of $0.74$--$0.87$ in each of the three
tracks.

\begin{table*}[t]\centering
{\tabstyle\setlength{\tabcolsep}{4pt}
\begin{tabular*}{\textwidth}[t]{@{\extracolsep{\fill}}ll r rr c rrr@{}}\toprule
& & & \multicolumn{2}{c}{\textbf{Mean supports share}} & & \multicolumn{3}{c}{\textbf{Stage-1 verdict accuracy}}\\
\cmidrule(lr){4-5}\cmidrule(lr){7-9}
\textbf{Model} & \textbf{Prompt} & \textbf{Refused/answered} & answered & refused & \textbf{AUC} [95\% CI] & A & B & C\\\midrule
Qwen2.5-7B & strict & 19/589 & 0.413 & 0.142 & 0.794 [0.703, 0.869] & 0.800 & 0.817 & 0.741\\
 & runtime & 18/590 & 0.412 & 0.167 & 0.766 [0.669, 0.850] & 0.801 & 0.811 & 0.760\\
 & minimal & 18/590 & 0.412 & 0.144 & 0.789 [0.692, 0.870] & 0.805 & 0.828 & 0.772\\
\addlinespace[2pt]
Qwen2.5-32B & strict & 27/581 & 0.417 & 0.126 & 0.814 [0.752, 0.870] & 0.839 & 0.867 & 0.815\\
 & runtime & 28/580 & 0.416 & 0.154 & 0.781 [0.702, 0.853] & 0.838 & 0.858 & 0.823\\
 & minimal & 25/583 & 0.415 & 0.160 & 0.775 [0.705, 0.843] & 0.833 & 0.853 & 0.807\\
\addlinespace[2pt]
Llama-3.1-8B & strict & 18/590 & 0.413 & 0.122 & 0.821 [0.772, 0.870] & 0.794 & 0.739 & 0.802\\
 & runtime & 18/590 & 0.412 & 0.144 & 0.793 [0.721, 0.858] & 0.795 & 0.746 & 0.784\\
 & minimal & 17/591 & 0.412 & 0.124 & 0.816 [0.748, 0.878] & 0.802 & 0.780 & 0.800\\
\addlinespace[2pt]
Mistral-7B & strict & 54/554 & 0.425 & 0.189 & 0.752 [0.701, 0.800] & 0.781 & 0.822 & 0.756\\
 & runtime & 52/556 & 0.425 & 0.188 & 0.753 [0.695, 0.808] & 0.783 & 0.816 & 0.749\\
 & minimal & 49/559 & 0.422 & 0.201 & 0.734 [0.680, 0.786] & 0.782 & 0.825 & 0.749\\
\bottomrule\end{tabular*}}

\caption{\textbf{Within-track checks for trace-tuned models.} The first three
columns use the 608 answerable examples of Tracks~A and~C: the number of refused
and answered examples, and the mean supports share of each group. The supports
share of an example is the fraction of its documents that the annotation committee
labelled \textit{supports}. AUC is the probability that an answered example has a
larger supports share than a refused one, with a 95\% bootstrap interval over
examples. The last three columns give the accuracy of the model's Stage-1 verdicts
against the committee's, by track, over the documents recognised in the trace.}
\label{tab:pertrack_checks}
\end{table*}

Under the minimal prompt, Factual Grounding rises after trace tuning in both
answerable tracks and for every model (Table~\ref{tab:pertrack_fg}).

\begin{table}[htbp]\centering
{\tabstyle\setlength{\tabcolsep}{4pt}
\begin{tabular*}{\linewidth}[t]{@{\extracolsep{\fill}}l rr rr@{}}\toprule
& \multicolumn{2}{c}{\textbf{Track A} ($n=508$)} & \multicolumn{2}{c}{\textbf{Track C} ($n=100$)}\\
\cmidrule(lr){2-3}\cmidrule(lr){4-5}
\textbf{Model} & base & \textsc{sft} & base & \textsc{sft}\\\midrule
Qwen2.5-7B & 0.019 & 0.827 & 0.030 & 0.880\\
Qwen2.5-32B & 0.096 & 0.809 & 0.165 & 0.890\\
Llama-3.1-8B & 0.080 & 0.847 & 0.040 & 0.908\\
Mistral-7B & 0.053 & 0.791 & 0.176 & 0.798\\
\bottomrule\end{tabular*}}

\caption{Factual Grounding by answerable track under the minimal prompt.}
\label{tab:pertrack_fg}
\end{table}

\section{Annotation Pipeline}
\label{app:annotation}

Annotation has three stages. At each stage every member of the annotation
committee (a panel of LLM annotators) independently receives the same structured
input and returns a structured output. The committee votes on the central
categorical decision for that stage, and the associated natural-language
explanation is taken whole from the highest-weight member that supported the
winning decision. Each label therefore comes with one internally coherent
explanation, and prose from different annotators is never averaged or spliced.

Formally, for a decision field with candidate values $v$, each annotator $m$
with a valid response contributes its configured weight $w_m$, and the pipeline
selects
\[
v^{*}=\arg\max_{v}\ \sum_{m} w_m\,\mathbf{1}[y_m=v].
\]
Exact ties are broken by a fixed candidate ordering, so identical annotator
outputs always give identical labels. Writing
$\mathcal{M}^{*}=\{m:y_m=v^{*}\}$ for the annotators on the winning side, the
explanation (verdict, quote and rationale) is taken whole from
$m^{*}=\arg\max_{m\in\mathcal{M}^{*}} w_m$, so all three come from one
annotator. Every document annotation retains
\texttt{\_vote\_tally}, \texttt{\_winner\_model} and \texttt{\_all\_verdicts},
so any label can be traced to its vote.

Stage-1 output follows a fixed format: a verdict from the closed set, a key fact
anchored to a quoted span of at most 50 words, a verdict rationale and a
URL-derived source-quality label. The Stage-1 prompt used for benchmark
annotation differs from the standard Stage-1 prompt in a single rule: older,
contradictory or otherwise conflict-bearing evidence about the target is
labelled \textit{partially supports}, so that it is kept and Stage~2 can see the
evidence it needs to identify temporal and scientific conflicts. At Stage~2 the
committee votes on the conflict label and the answerability flag. At Stage~3 it
votes on the abstention decision independently of the Stage-2 answerability flag,
so the two fields can differ and a complete analysis needs both.

Structural validators check schema conformance, the allowed label sets and
document/annotation alignment. The human reviews in Appendix~\ref{app:human-review}
assess label quality.

\section{Human Review of the Resource}
\label{app:human-review}

In this appendix, \emph{the committee} is the annotation committee of
Appendix~\ref{app:annotation}. It covers three human review workflows, each with its own target:
Stage-1 document verdicts, training conflict types, and benchmark preselection
(with a refusal-quality group). We release the raw per-reviewer files,
assignment manifests and agreement metrics for all three.

\subsection{Stage-1 document-verdict review}
\label{app:stage1review}

The two larger reviews that follow check the conflict-type label and the
preselection decision. This smaller study checks the Stage-1 per-document verdict
(\textit{supports}, \textit{partially supports}, \textit{irrelevant}), which
decides whether a document can support a claim for Factual Grounding
(\S\ref{sec:cats}).

\paragraph{Design.} We drew 100 query--document pairs that already had
Stage-1 annotation-committee verdicts, using seed \texttt{20260919}. The sample is balanced
by committee verdict (supports: 34; partially supports: 33; irrelevant: 33)
and deduplicated so each normalised query appears once. The pairs are drawn
from the held-out benchmark (46), the training set (50) and the validation
set (4). Three reviewers each assessed all 100 pairs (300 ratings). Each saw
the query, the document, and the committee's Stage-1 annotation, and recorded
agreement or a replacement verdict with a reason.

\paragraph{Human--human agreement.} Table~\ref{tab:stage1hh} gives pairwise
and three-rater agreement. Three-way unanimity holds on 83/100 pairs, with a
2--1 majority on the remaining 17 and no three-way splits.

\begin{table*}[t]\centering
{\tabstyle\setlength{\tabcolsep}{3pt}
\begin{tabular*}{\textwidth}[t]{@{\extracolsep{\fill}}l r r r@{}}\toprule
\textbf{Comparison} & \textbf{Items} & \textbf{Exact agreement} & $\kappa$\\\midrule
Reviewer 1 vs Reviewer 2 & 100 & 88.0\% & 0.820\\
Reviewer 1 vs Reviewer 3 & 100 & 85.0\% & 0.776\\
Reviewer 2 vs Reviewer 3 & 100 & 93.0\% & 0.895\\
\addlinespace[2pt]
Fleiss' $\kappa$ (3 raters) & 100 & --- & 0.830\\
Krippendorff's $\alpha$ (nominal) & 300 & --- & 0.830\\
Mean pairwise exact agreement & 300 & 88.7\% & ---\\
Unanimous verdicts (3/3) & 100 & 83.0\% & ---\\
2--1 majority (no 3-way ties) & 100 & 17.0\% & ---\\
\bottomrule\end{tabular*}}

\caption{\textbf{Stage-1 review: human--human agreement} across the three reviewers, 100 pairs (300 ratings). Items are query--document pairs ($100$) or individual reviewer ratings ($300$). $\kappa$ is chance-corrected agreement ($1$ is perfect, $0$ is chance level); the $\kappa$ column also holds Fleiss' $\kappa$ and Krippendorff's $\alpha$ where labelled. ``---'' marks a statistic that is not defined for that row.}
\label{tab:stage1hh}
\end{table*}

\paragraph{Committee--human agreement.} Table~\ref{tab:stage1llm} compares the
committee's Stage-1 verdict against each reviewer and against the human
majority. The pooled row compares all 300 ratings with the committee label
(counted once for each reviewer who rated the pair) and is only a descriptive
summary.

\begin{table*}[t]\centering
{\tabstyle\setlength{\tabcolsep}{3pt}
\begin{tabular*}{\textwidth}[t]{@{\extracolsep{\fill}}l r r r@{}}\toprule
\textbf{Comparison} & \textbf{Items} & \textbf{Exact agreement} & $\kappa$\\\midrule
Annotation committee vs Reviewer 1 & 100 & 96.0\% & 0.940\\
Annotation committee vs Reviewer 2 & 100 & 90.0\% & 0.850\\
Annotation committee vs Reviewer 3 & 100 & 87.0\% & 0.805\\
Annotation committee vs human majority & 100 & 92.0\% & 0.880\\
Annotation committee vs all individual ratings\textsuperscript{*} & 300 & 91.0\% & 0.865\\
\bottomrule\end{tabular*}}

\caption{\textbf{Stage-1 review: committee--human agreement.} \textsuperscript{*}Pooled statistic. The 300 ratings share 100 underlying pairs.}
\label{tab:stage1llm}
\end{table*}

\paragraph{Where the committee and reviewers disagree.} Table~\ref{tab:stage1confusion} gives the confusion matrix over all 300 ratings
and the direction of the 27 ratings (18 distinct pairs) in which a reviewer
replaced the committee's verdict. Every replacement moves to a neighbouring
category: no reviewer moved a pair between \textit{supports} and
\textit{irrelevant}. The reviewers' disagreements with each other show the same
pattern (\textit{supports} vs \textit{partially supports} on 11 pairs,
\textit{partially supports} vs \textit{irrelevant} on 6, none between
\textit{supports} and \textit{irrelevant}). Reviewers agree most often with the
committee's \textit{irrelevant} verdicts ($94.9\%$ of ratings, $94/99$), then
\textit{supports} ($89.2\%$, $91/102$) and \textit{partially supports}
($88.9\%$, $88/99$). Factual Grounding treats \textit{supports} and
\textit{partially supports} alike as usable, so only the $11$ replacements between
\textit{irrelevant} and \textit{partially supports} ($6+5$) can change which
documents are eligible. The other $16$ replacements move between \textit{supports}
and \textit{partially supports} and leave eligibility unchanged.

\begin{table*}[t]\centering
{\tabstyle\setlength{\tabcolsep}{3pt}
\begin{tabular*}{\textwidth}[t]{@{\extracolsep{\fill}}l r r r r@{}}\toprule
\textbf{Committee verdict} & \textbf{Human: supports} & \textbf{Human: partially} & \textbf{Human: irrelevant} & \textbf{Total}\\\midrule
Supports & 91 & 11 & 0 & 102\\
Partially supports & 5 & 88 & 6 & 99\\
Irrelevant & 0 & 5 & 94 & 99\\
\bottomrule\end{tabular*}}

\vspace{6pt}
{\tabstyle\setlength{\tabcolsep}{3pt}
\begin{tabular*}{\textwidth}[t]{@{\extracolsep{\fill}}l r r@{}}\toprule
\textbf{Committee to human verdict} & \textbf{Correction ratings} & \textbf{Distinct pairs}\\\midrule
Supports to partially supports & 11 & 7\\
Partially supports to supports & 5 & 4\\
Partially supports to irrelevant & 6 & 4\\
Irrelevant to partially supports & 5 & 3\\
\bottomrule\end{tabular*}}\par

\caption{\textbf{Stage-1 review: pooled confusion matrix} (top, 300 ratings against the repeated committee label) \textbf{and correction directions} (bottom, among the 27 ratings that changed the committee's verdict).}
\label{tab:stage1confusion}
\end{table*}

\paragraph{Summary.} The committee's Stage-1 verdict matches the human-majority
verdict on $92\%$ of pairs, and every disagreement is between neighbouring
categories, so we use it to decide which documents are eligible for Factual
Grounding (\S\ref{sec:limitations}).

\subsection{Training conflict-type review}

Each of the 943 training and validation records was reviewed by two reviewers,
assigned from a pool of three under a seeded allocation that balances reviewer
pairs. Reviewers saw the query, the retrieved snippets and the committee-assigned
label. They could keep or replace the label, recording their confidence and a
reason. The review validates labels and removes no record from the training
pool.

\begin{table*}[t]
\centering
\begin{minipage}[c]{0.48\linewidth}
\centering
{\tabstyle\setlength{\tabcolsep}{3pt}
\begin{tabular*}{\linewidth}[t]{@{\extracolsep{\fill}}l rr r@{}}\toprule
\textbf{Review field} & \textbf{agree} & \textbf{raw} & $\kappa$\\\midrule
Reviewed conflict type & 787 & 83.46\% & 0.7694\\
Retain-or-change action & 791 & 83.88\% & 0.1341\\
Changed-label (boolean) & 791 & 83.88\% & 0.1341\\
Reviewer confidence & 607 & 64.37\% & 0.0159\\
\bottomrule\end{tabular*}}

\end{minipage}%
\hspace{0.04\linewidth}%
\begin{minipage}[c]{0.48\linewidth}
\centering
{\tabstyle\setlength{\tabcolsep}{2.5pt}
\begin{tabular*}{\linewidth}[t]{@{\extracolsep{\fill}}l rr r@{}}\toprule
\textbf{Review field} & \textbf{agree} & \textbf{raw} & $\kappa$\\\midrule
Preliminary conflict type & 1378 & 94.77\% & 0.9217\\
Retention decision & 1428 & 98.21\% & 0.9228\\
Retrieval quality & 1436 & 98.76\% & 0.9550\\
Evidence sufficiency & 1436 & 98.76\% & 0.9600\\
Conflict clarity & 1441 & 99.11\% & 0.9660\\
Query specificity & 1449 & 99.66\% & 0.9875\\
Source reliability & 1446 & 99.45\% & 0.9881\\
Relevant-document count & 1426 & 98.07\% & 0.9527\\
Gold-answer feasibility & 1440 & 99.04\% & 0.9669\\
Reviewer confidence & 1445 & 99.38\% & 0.9767\\
\bottomrule\end{tabular*}}

\end{minipage}
\dualcaption{table}{tab:trainiaa}{tab:benchiaa}%
{\textbf{Training conflict-type review agreement} over all 943 records, each reviewed by two reviewers. The five-way conflict-type field is the primary result. On the other rows $\kappa$ (chance-corrected agreement) is low because one answer dominates: most reviewers kept the committee label and reported high confidence, so one answer dominates each field and $\kappa$ is low}%
{\textbf{Benchmark preselection agreement} (the review that decides which candidates enter the benchmark), all ten recorded review fields over the $1{,}454$-record consolidated population. Every field exceeds $\kappa=0.92$, and the primary result is the five-way conflict-type field. }
\end{table*}

In Table~\ref{tab:trainiaa}, the two reviewers agree on the five-way
conflict-type field for $787$ of $943$ records, giving $83.46\%$ raw agreement and
$\kappa=0.7694$.

\paragraph{Effect of review on labels.} The first reviewer kept the committee
label for $801/943$ records ($84.94\%$), the second for $895/943$
($94.91\%$), and both kept it for $772/943$ ($81.87\%$). At least one reviewer
changed the label for $171$ of the $943$ records, roughly one in five.

\paragraph{Reviewer labels and released labels.} The released
\texttt{conflict\_type} is unchanged by the review, so the reviewers' label counts
differ from the released counts. Reviewers' first-pass labels give No conflict $373$, Complementary
$289$, Outdated $130$, Conflicting opinions $120$ and Misinformation $31$,
against released counts of $343$, $286$, $136$, $158$ and $20$. The largest
relative differences are in the misinformation and conflicting-opinions classes.
We use the review as a reliability estimate for the taxonomy.

\subsection{Benchmark preselection review}

The $1{,}878$ candidates with exactly ten documents were split among seven
reviewers under seeded source balancing, in queues of $150$, $150$, $300$, $300$, $326$, $326$ and $326$ candidates. The
source mix is ConflictingQA $258$, FreshQA $394$, SituatedQA-temporal $413$,
QACC $407$ and SituatedQA-geographic $406$. Each reviewer recorded an acceptance
decision, a preliminary conflict type, confidence, retrieval quality, evidence
sufficiency and conflict clarity. They also recorded query specificity, source reliability, a
relevant-document count bin, gold-answer feasibility and free-text notes.

Every one of the ten recorded fields in Table~\ref{tab:benchiaa} exceeds
$\kappa=0.92$, with the five-way conflict-type field at $94.77\%$ raw agreement
and $\kappa=0.9217$ ($1{,}378$ of $1{,}454$ records agreeing).

\paragraph{Second-pass review.} The second reviewer saw the complete first
review and could accept, edit or reject it, so the agreement figures in
Table~\ref{tab:benchiaa} measure verification consistency. Across the 800 selected records, the second reviewer
accepted $673$ first reviews as they were and edited $127$. For three queues
($114$, $115$ and $114$ records) the second reviewers confirmed that they accepted
every first review, and the release records this in a manifest.

\subsection{Refusal-quality validation}
\label{app:refusalquality}

The $128$ released refusal examples were reviewed as a separate group against four
dedicated fields in addition to the ten preselection fields. Agreement is
\emph{unanimous on all fourteen}: both reviewers judged all $128$ examples to
require refusal, to have a valid abstention ground truth, to carry a
high-quality evidence-gap rationale, and to be a valid refusal benchmark example.
They also agreed unanimously that evidence is insufficient and that no defensible
gold answer is possible for any of them. Because $13$ of the $14$ fields have a
single value across all $128$ examples, $\kappa$ is undefined for them and we report
exact agreement; the conflict-type field has $\kappa=1.0000$.

This complements the committee check in Appendix~\ref{app:relverif}, where
\textit{supports} verdicts occur on $3.6\%$ of refusal-track documents against
$39.3\%$ elsewhere. The refusal labels are therefore supported by two checks: the
committee's document-level verdicts, produced without the abstention label, and
unanimous human review.

\setcounter{dbltopnumber}{3}
\section{Fine-Tuning Details}
\label{app:sft}

\subsection{Optimisation and checkpoint selection}

All models are fine-tuned with QLoRA
\citep{dettmers2023qloraefficientfinetuningquantized,hu2021loralowrankadaptationlarge}
at rank 32, scaling $\alpha=64$ and dropout $0.05$, with learning rate
$2\times10^{-4}$, effective batch 16, maximum sequence length $12{,}288$ and
NEFTune (training-time embedding noise) with scale $\alpha=5.0$ \citep{jain2023neftune}. Training minimises the standard
token-level cross-entropy of the serialised trace and final response given the
prompt and retrieved documents. No auxiliary loss term is added. Checkpoints are
selected on the validation split by a composite score that sums document-verdict
accuracy, response-format validity and abstention correctness. A \emph{recipe} is
one fixed training mixture: the set and proportion of task views (the full trace,
or only the document verdicts, the conflict type or the final response) and
prompt variants (strict, runtime or minimal) a run is trained on. Qwen2.5-7B and
Qwen2.5-32B use recipe~A. Llama-3.1-8B and Mistral-7B use recipe~B.

\subsection{Training mixtures}

Both recipes start from the 862 training examples. Each source example appears in
several rows (supervised views), so row counts exceed the number of source
examples. Recipe~A (Qwen2.5-7B and Qwen2.5-32B; $12{,}659$ rows) adds 27 derived
five-document answerable examples and extra examples that target document
boundaries and partial synthesis. Recipe~B (Llama-3.1-8B and Mistral-7B;
$13{,}349$ rows) adds 21 short answerable \emph{no-conflict} examples to
recipe~A, which gives $910$ source examples in total.

\section{Answer-Only Supervision Ablation}
\label{app:answeronly}

\subsection{Recipe}

This ablation covers Llama-3.1-8B and Qwen2.5-7B. It compares training on the
full output (the \texttt{<think>} block with Stages~1 and~2, followed by the
Stage-3 response) with training on the Stage-3 response alone, which tests what the
think trace contributes. Both use the same 862/81/736 train/validation/benchmark
split and the same QLoRA configuration (rank 32, $\alpha=64$, dropout $0.05$,
learning rate $2\times10^{-4}$, maximum sequence length $12{,}288$, NEFTune
$\alpha=5.0$) as the trace-supervised runs. The settings differ in the supervision
target, and the trace-supervised runs also add 27 derived examples (recipe~A,
Qwen) or 48 (recipe~B, Llama) to the 862 training examples (Appendix~\ref{app:sft}). The answer-only runs
train on $10{,}344$ rows and the trace-supervised runs on $12{,}659$ rows (recipe~A)
or $13{,}349$ rows (recipe~B), because the trace-supervised mixtures also contain
separate task views such as verdict-only and conflict-type-only rows
(\S\ref{sec:sft}).

The answer-only training target is only the final grounded answer or the refusal
string \texttt{CANNOT ANSWER, INSUFFICIENT EVIDENCE}, immediately followed by
\texttt{[[END-OF-ANSWER]]}. It contains no
\texttt{<think>} block, document-verdict array or conflict-type label. We train
on two prompt families: \textbf{final-only}, which tells the model to omit the
trace and produce only the final answer, and \textbf{minimal}, the 40-word
prompt used throughout the paper (Appendix~\ref{app:e2eprompts}). Each of the 862 source examples
yields 8 final-only and 4 minimal copies (a 2:1 mixture), giving $10{,}344$
training rows.

Generation is deterministic (temperature $0$), with the same 4-bit base-model
loading and LoRA adapter as the trace-supervised runs. For both prompt families
we compute refusal-decision metrics (precision, recall and F1 for the answer
and the refuse class, and overall classification accuracy), and, on the same
generations, Behaviour Adherence, Factual Grounding and Single-Truth Recall.
Answer Quality (AQ) is defined in Appendix~\ref{app:aggregate}: the geometric
mean of Factual Grounding and Single-Truth Recall, or Factual Grounding alone
where Single-Truth Recall does not apply.

\subsection{Results}

Tables~\ref{tab:aollama} and~\ref{tab:aoqwen} compare the two answer-only prompt
variants (final-only and minimal) with the three trace-supervised prompt variants
(strict, runtime and minimal) for each model. Table~\ref{tab:aodelta} gives the
minimal-prompt comparison. The minimal prompt gives the least instruction, so it
shows most clearly what each type of supervision taught the model
(\S\ref{sec:internalization}).

\begin{table*}[t]\centering
{\tabstyle\setlength{\tabcolsep}{2pt}
\begin{tabular*}{\linewidth}[t]{@{\extracolsep{\fill}}l r r r r r@{}}\toprule
& \multicolumn{2}{c}{Answer-only} & \multicolumn{3}{c}{Trace} \\
\cmidrule(lr){2-3}\cmidrule(lr){4-6}
\textbf{Metric (\%)} & \textbf{minimal} & \textbf{final-only} & \textbf{minimal} & \textbf{runtime} & \textbf{strict}\\\midrule
Answering Precision & 98.41 & 98.40 & \textbf{99.83} & \textbf{99.83} & \textbf{99.66}\\
Answering Recall & 91.45 & 91.28 & \textbf{97.20} & \textbf{97.04} & \textbf{97.04}\\
Answering F1 & 94.80 & 94.71 & \textbf{98.50} & \textbf{98.42} & \textbf{98.33}\\
Refusal Precision & 69.59 & 69.19 & \textbf{88.19} & \textbf{87.59} & \textbf{87.50}\\
Refusal Recall & 92.97 & 92.97 & \textbf{99.22} & \textbf{99.22} & \textbf{98.44}\\
Refusal F1 & 79.60 & 79.33 & \textbf{93.38} & \textbf{93.04} & \textbf{92.65}\\
Overall Classification Accuracy & 91.71 & 91.58 & \textbf{97.55} & \textbf{97.42} & \textbf{97.28}\\
Behaviour Adherence & 72.11 & 71.10 & \textbf{79.47} & \textbf{80.62} & \textbf{80.49}\\
Answer Quality & 57.80 & 54.90 & \textbf{69.40} & \textbf{69.00} & \textbf{70.60}\\
Single-Truth Recall & 69.64 & 68.17 & \textbf{71.36} & \textbf{70.53} & \textbf{72.69}\\
Factual Grounding & 71.83 & 70.47 & \textbf{85.74} & \textbf{84.91} & \textbf{85.16}\\
\bottomrule\end{tabular*}}

\caption{\textbf{Llama-3.1-8B: answer-only vs.\ trace supervision}, all five prompt$\times$supervision conditions (two answer-only, three trace-supervised). Values are percentages.}
\label{tab:aollama}
\end{table*}

\begin{table*}[t]\centering
{\tabstyle\setlength{\tabcolsep}{2pt}
\begin{tabular*}{\linewidth}[t]{@{\extracolsep{\fill}}l r r r r r@{}}\toprule
& \multicolumn{2}{c}{Answer-only} & \multicolumn{3}{c}{Trace} \\
\cmidrule(lr){2-3}\cmidrule(lr){4-6}
\textbf{Metric (\%)} & \textbf{minimal} & \textbf{final-only} & \textbf{minimal} & \textbf{runtime} & \textbf{strict}\\\midrule
Answering Precision & 96.35 & 97.37 & \textbf{97.84} & \textbf{98.83} & \textbf{99.16}\\
Answering Recall & 91.28 & 91.28 & \textbf{97.04} & \textbf{97.04} & \textbf{96.88}\\
Answering F1 & 93.75 & 94.23 & \textbf{97.44} & \textbf{97.93} & \textbf{98.00}\\
Refusal Precision & 66.88 & 68.07 & \textbf{86.47} & \textbf{87.05} & \textbf{86.62}\\
Refusal Recall & 83.59 & 88.28 & \textbf{89.84} & \textbf{94.53} & \textbf{96.09}\\
Refusal F1 & 74.31 & 76.87 & \textbf{88.12} & \textbf{90.64} & \textbf{91.11}\\
Overall Classification Accuracy & 89.95 & 90.76 & \textbf{95.79} & \textbf{96.60} & \textbf{96.74}\\
Behaviour Adherence & 73.18 & 73.19 & \textbf{76.81} & \textbf{77.40} & \textbf{77.98}\\
Answer Quality & 51.20 & 53.50 & \textbf{68.00} & \textbf{69.00} & \textbf{67.70}\\
Single-Truth Recall & 61.27 & 62.20 & \textbf{68.58} & \textbf{69.30} & \textbf{68.69}\\
Factual Grounding & 68.34 & 70.24 & \textbf{83.75} & \textbf{82.91} & \textbf{83.84}\\
\bottomrule\end{tabular*}}

\caption{\textbf{Qwen2.5-7B: answer-only vs.\ trace supervision}, all five prompt$\times$supervision conditions (two answer-only, three trace-supervised). Values are percentages.}
\label{tab:aoqwen}
\end{table*}

\begin{table*}[t]\centering
{\tabstyle\setlength{\tabcolsep}{2pt}
\begin{tabular*}{\linewidth}[t]{@{\extracolsep{\fill}}l r r r r r r@{}}\toprule
& \multicolumn{3}{c}{Llama-3.1-8B} & \multicolumn{3}{c}{Qwen2.5-7B} \\
\cmidrule(lr){2-4}\cmidrule(lr){5-7}
\textbf{Metric (\%), minimal prompt} & \textbf{AO} & \textbf{Trace} & \textbf{$\Delta$(pp)} & \textbf{AO} & \textbf{Trace} & \textbf{$\Delta$(pp)}\\\midrule
Answering Precision & 98.41 & 99.83 & +1.42 & 96.35 & 97.84 & +1.49\\
Answering Recall & 91.45 & 97.20 & +5.76 & 91.28 & 97.04 & +5.76\\
Answering F1 & 94.80 & 98.50 & +3.70 & 93.75 & 97.44 & +3.69\\
Refusal Precision & 69.59 & 88.19 & \textbf{+18.60} & 66.88 & 86.47 & \textbf{+19.59}\\
Refusal Recall & 92.97 & 99.22 & +6.25 & 83.59 & 89.84 & +6.25\\
Refusal F1 & 79.60 & 93.38 & \textbf{+13.78} & 74.31 & 88.12 & \textbf{+13.82}\\
Overall Classification Accuracy & 91.71 & 97.55 & +5.84 & 89.95 & 95.79 & +5.84\\
Behaviour Adherence & 72.11 & 79.47 & \textbf{+7.36} & 73.18 & 76.81 & \textbf{+3.64}\\
Answer Quality & 57.80 & 69.40 & \textbf{+11.60} & 51.20 & 68.00 & \textbf{+16.80}\\
Single-Truth Recall & 69.64 & 71.36 & \textbf{+1.72} & 61.27 & 68.58 & \textbf{+7.31}\\
Factual Grounding & 71.83 & 85.74 & \textbf{+13.92} & 68.34 & 83.75 & \textbf{+15.41}\\
\bottomrule\end{tabular*}}

\caption{\textbf{Answer-only vs.\ trace supervision under the minimal prompt}, both models, with the percentage-point (pp) difference. AO is answer-only supervision and Trace is trace supervision. Differences are computed from unrounded scores, so they can differ by $0.01$pp from the difference of the rounded values shown.}
\label{tab:aodelta}
\end{table*}

Under the minimal prompt, trace supervision exceeds answer-only supervision
on every metric for both models: Refusal F1 by $+13.78$pp (Llama-3.1-8B) and
$+13.82$pp (Qwen2.5-7B), Factual Grounding by $+13.92$pp and $+15.41$pp, and
Answer Quality by $+11.60$pp and $+16.80$pp. The largest gap is Refusal
Precision, the share of the model's refusals that were required ($+18.60$pp and
$+19.59$pp): the trace teaches a better answer/refuse decision than the answer
alone. Behaviour Adherence improves by $+7.36$pp and $+3.64$pp, and Single-Truth
Recall by $+1.72$pp and $+7.31$pp. The two supervision types share the base
training data, base model and optimiser, so the gap is mainly attributable to the
think trace in the supervision target. The 27 (Qwen2.5-7B) or 48 (Llama-3.1-8B)
additional examples in the trace-supervised runs are 3--6\% more source examples
(Appendix~\ref{app:sft}).

\section{Inference Protocol}
\label{app:inference}

Generation is deterministic (temperature $0$) and stops at the end-of-answer
marker \texttt{[[END-OF-ANSWER]]}. Each prompt variant has its own token budget,
with a larger cap for retries: strict $1{,}400$ tokens (retry cap $3{,}200$),
runtime $1{,}200$ ($2{,}200$) and minimal $900$ ($1{,}800$). A strict trace lists
every retrieved document, so a single shared budget would truncate long-context
examples and confound reasoning quality with available output length.

When the output does not meet the required format, the \emph{identical} prompt is
re-run with the larger token budget. A retry never resamples, appends corrective
feedback, changes the retrieved documents or switches the prompt variant. The
minimal prompt has no format-based retry.

Every run writes both a raw and a sanitised output file. Sanitisation puts
documents and identifiers in standard order, maps label variants to the standard
labels, clears key facts on irrelevant documents, drops out-of-range citations
and appends a missing end marker when a usable trace exists. It never adds a
missing trace, document verdicts, a conflict analysis or an answer. A record
without a parseable trace passes through unchanged.

\subsection{Output length as a cost proxy}
\label{app:inflatency}

\begin{table*}[t]\centering
{\tabstyle\setlength{\tabcolsep}{3pt}
\begin{tabular*}{\linewidth}[t]{@{\extracolsep{\fill}}l rrr rrr rrr@{}}\toprule
& \multicolumn{3}{c}{\textbf{Trace-tuned SFT}} & \multicolumn{3}{c}{\textbf{Answer-only SFT}} & \multicolumn{3}{c}{\textbf{Direct-answer baseline}}\\
\cmidrule(lr){2-4}\cmidrule(lr){5-7}\cmidrule(lr){8-10}
\textbf{Model} & mean & med. & p95 & mean & med. & p95 & mean & med. & p95\\\midrule
Qwen2.5-7B & 2495.1 & 2376.5 & 3557 & 253.9 & 88.5 & 838 & 377.2 & 242.5 & 1164\\
Llama-3.1-8B & 2424.9 & 2290.0 & 3434 & 240.5 & 91.0 & 802 & 280.9 & 91.0 & 1359\\
Mistral-7B & 2575.1 & 2431.5 & 3584 & 1545.0 & 1508.0 & 1924 & 302.1 & 202.5 & 800\\
Qwen2.5-32B & 2421.8 & 2302.0 & 3445 & -- & -- & -- & 525.8 & 410.5 & 1469\\
\bottomrule\end{tabular*}}

\caption{\textbf{Output length by supervision condition}, characters over the 736-example benchmark under the minimal prompt. Answer-only models: Qwen2.5-7B, Llama-3.1-8B and Mistral-7B.}
\label{tab:outputlength}
\end{table*}

\begin{table*}[t]\centering
{\tabstyle\setlength{\tabcolsep}{4pt}
\begin{tabular*}{0.7\linewidth}[t]{@{\extracolsep{\fill}}l rrr rrr@{}}\toprule
& \multicolumn{3}{c}{\textbf{Chain-of-Note}} & \multicolumn{3}{c}{\textbf{Few-shot Chain-of-Thought}}\\
\cmidrule(lr){2-4}\cmidrule(lr){5-7}
\textbf{Model} & mean & med. & p95 & mean & med. & p95\\\midrule
Qwen2.5-7B & 1827.7 & 1766.0 & 2637 & 2092.9 & 1960.0 & 3160\\
Mistral-7B & 1661.1 & 1579.0 & 2461 & 2144.2 & 2044.0 & 3411\\
\bottomrule\end{tabular*}}

\caption{\textbf{Output length for the two prompting-only comparators}, characters over the 736-example benchmark. The comparator models here are Qwen2.5-7B and Mistral-7B (\S\ref{sec:comparators}).}
\label{tab:outputlengthbaselines}
\end{table*}

The structured trace lengthens the output. The serialised \texttt{<think>} block
adds Stage-1 per-document verdicts, a Stage-2 conflict label and a short Stage-3
answer plan before the final answer, and the direct-answer and
prompting-only baselines produce no such structured trace. We measure generation cost as the character
length of the raw saved output over the full 736-example benchmark, reported as
mean, median and 95th percentile (nearest-rank).

Table~\ref{tab:outputlength} gives the lengths of trace-tuned and
direct-answer baseline outputs for all four models and of answer-only outputs for
three, and Table~\ref{tab:outputlengthbaselines} gives the two prompting-only
comparators for Qwen2.5-7B and Mistral-7B. Trace-tuned outputs run
roughly $10\times$ longer than answer-only outputs for Qwen2.5-7B and
Llama-3.1-8B, and $4.6$--$8.6\times$ longer than the direct-answer baseline,
depending on the model (smallest for Qwen2.5-32B, largest for Llama-3.1-8B). The
answer-only Mistral-7B model, reported here for output length only, is the one
exception: its outputs are only about $1.7\times$ shorter than the trace-tuned
ones, because it emits long, verbose answers ($1{,}545$ characters on average)
despite receiving no trace supervision.

\section{CATS Specification}
\label{app:cats}

\subsection{Component definitions}

\paragraph{Grounded Refusal (GR).} Whether the system answered exactly when the
evidence permits answering. It is computed \emph{deterministically} from the
benchmark answerability label and needs no judge. We report refusal F1, where a
true positive is the model declining to answer when the retrieved evidence does
not support an answer.

\begin{table*}[t]\centering{\tabstyle\setlength{\tabcolsep}{3pt}
\begin{tabular}{@{}>{\raggedright\arraybackslash}p{0.18\textwidth} >{\raggedright\arraybackslash}p{0.30\textwidth} >{\raggedright\arraybackslash}p{0.48\textwidth}@{}}\toprule
\textbf{Conflict type} & \textbf{Definition} & \textbf{Expected behaviour}\\\midrule
No conflict & All sources provide consistent and aligned information. & Return a unified, concise answer that synthesises the content accurately.\\\addlinespace[2pt]
Complementary information & Sources provide different but non-contradictory pieces that complete each other. & Integrate all valid details into a richer, more comprehensive response.\\\addlinespace[2pt]
Conflicting opinions or research outcomes & Sources disagree due to different viewpoints, experimental results or interpretations. & Present each viewpoint clearly, describe the nature of the disagreement, and avoid taking a side unless the evidence strongly supports one.\\\addlinespace[2pt]
Outdated information & Some sources contain older data or conclusions that may no longer be reliable. & Prioritise newer, verified information while explicitly noting which sources are outdated.\\\addlinespace[2pt]
Misinformation & A source provides factually incorrect, misleading or fabricated claims. & Reject or correct the misinformation, give the verified facts, and briefly explain why the claim is wrong.\\
\bottomrule\end{tabular}}
\caption{\textbf{The five conflict types}, their definitions and the response behaviour each requires. Stage~2 assigns exactly one label per example; Behaviour Adherence (\S\ref{sec:cats}) scores the response against the expected behaviour for the assigned type.}
\label{tab:taxonomy}\end{table*}

\paragraph{Behaviour Adherence (BA).} Whether the answer followed the response
policy its conflict type requires, judged against a per-type rubric. A
no-conflict item expects a direct answer. Complementary evidence expects
synthesis across documents. Conflicting opinions expect both positions
presented neutrally. Outdated information expects the current value
with the superseded one flagged, and misinformation expects explicit correction.
The rubric is reproduced verbatim in Appendix~\ref{app:catsprompts}.

\paragraph{Factual Grounding (FG).} Whether each extracted claim is both
supported by an eligible retrieved document \emph{and} cited. Only documents
whose \emph{gold} per-document verdict is \textit{supports} or \textit{partially
supports} are eligible. FG therefore measures grounding in the retrieved
documents. A claim can also be supported by two documents jointly.

\paragraph{Single-Truth Recall (STR).} Whether the response explicitly adopts the
target answer as its own conclusion. A response fails this test if it only quotes
the answer, presents it as one of several possibilities, or attributes it to a
source without endorsing it. STR is evaluated only on the 487 benchmark examples that
are answerable, have a gold answer, and have a single correct conclusion: no
conflict ($154$),
complementary information ($156$), outdated information ($140$) and
misinformation ($37$). Of the $608$ answerable examples, $96$ have conflicting opinions, which have no
such single conclusion, and $25$ more (20 complementary, 5 outdated) have no gold
answer, which leaves $487$. Refusal examples have no gold answer, so
every model is scored on the same $487$ examples.

\subsection{Applicability and completeness}

For example $i$, let $A_i\in\{0,1\}$ be gold answerability, $\hat A_i$ the parsed
answer/refuse decision and $t_i\in\{1,\dots,5\}$ the conflict type. For a metric
$M$, write $\mathcal{I}_M$ for its applicable examples and $n_M=|\mathcal{I}_M|$.
Every dataset-level component below is the mean of per-example scores over its
applicable set (each example counts once). Correct refusals carry no
answer-content metrics, so
\[
\begin{gathered}
\mathcal{I}_{\mathrm{BA}}=\mathcal{I}_{\mathrm{FG}}
=\{i:\neg(A_i=0 \wedge \hat A_i=0)\},\\
\mathcal{I}_{\mathrm{AQ}}=\mathcal{I}_{\mathrm{FG}},
\end{gathered}
\]
and $\mathcal{I}_{\mathrm{STR}}\subseteq\mathcal{I}_{\mathrm{FG}}$ additionally
requires a gold answer and a single-truth conflict type ($n_{\mathrm{STR}}=487$
on the benchmark). A false
refusal stays in these denominators with zero extracted content. A CATS
aggregate is computed only when every applicable example has a score and every
conflict type present has a value; otherwise the run is marked incomplete.

\subsection{Grounded Refusal}

The per-example decision gate is $g_i=\mathbf{1}[\hat A_i=A_i]$. With
``answered'' as the positive class,
\[
\begin{gathered}
\mathrm{TP}=\textstyle\sum_i\mathbf{1}[\hat A_i{=}1,A_i{=}1],\\
\mathrm{FP}=\textstyle\sum_i\mathbf{1}[\hat A_i{=}1,A_i{=}0],\\
\mathrm{FN}=\textstyle\sum_i\mathbf{1}[\hat A_i{=}0,A_i{=}1],\\
\mathrm{TN}=\textstyle\sum_i\mathbf{1}[\hat A_i{=}0,A_i{=}0],
\end{gathered}
\]
giving $P_{\mathrm{ans}}=\mathrm{TP}/(\mathrm{TP}+\mathrm{FP})$,
$R_{\mathrm{ans}}=\mathrm{TP}/(\mathrm{TP}+\mathrm{FN})$ and
$F_{1,\mathrm{ans}}=2P_{\mathrm{ans}}R_{\mathrm{ans}}/(P_{\mathrm{ans}}+R_{\mathrm{ans}})$,
with a zero denominator yielding zero. The refusal-positive orientation swaps
$(\mathrm{TP},\mathrm{FP},\mathrm{FN},\mathrm{TN})\!\to\!(\mathrm{TN},\mathrm{FN},\mathrm{FP},\mathrm{TP})$:
\[
P_{\mathrm{ref}}=\frac{\mathrm{TN}}{\mathrm{TN}+\mathrm{FN}},\qquad
R_{\mathrm{ref}}=\frac{\mathrm{TN}}{\mathrm{TN}+\mathrm{FP}},
\]
and analogously $F_{1,\mathrm{ref}}$. Decision accuracy is
$(\mathrm{TP}+\mathrm{TN})/N$. We report both orientations. A model can have
near-perfect answer recall and still refuse too rarely, and a model that answers
every question already reaches $608/736=0.826$ accuracy because $82.6\%$ of the
benchmark is answerable (\S\ref{sec:refusal}).

\subsection{Committee voting rule for BA and STR}

The judge committee consists of Qwen3.5-397B-A17B, Mistral Small 4 and
DeepSeek-R1-Distill-32B, run locally, with priority weights $p_j=(6,3,2)$. On a
binary question,
each valid judge $j$ emits vote $z_{ij}\in\{0,1\}$ and confidence $c_{ij}\in[0,1]$,
contributing effective weight and vote masses
\[
a_{ij}=p_j\max(c_{ij},0.01),
\]
\[
W_i^{+}=\!\!\sum_{j:z_{ij}=1}\!\!a_{ij},\qquad
W_i^{-}=\!\!\sum_{j:z_{ij}=0}\!\!a_{ij}.
\]
The decision and the continuous positive support are
\[
d_i=\mathbf{1}[W_i^{+}>W_i^{-}],\qquad
u_i=\frac{W_i^{+}}{W_i^{+}+W_i^{-}}.
\]
A weighted tie is negative. Timeout, API and parse failures are removed before
forming these sums. If every judge fails on an example, the example has no score and the run is marked
incomplete (Appendix~\ref{app:cats}, Applicability and completeness).

\subsection{Behaviour Adherence}

Here $z_{ij}$ asks whether the response follows the type-specific behaviour
rubric (Appendix~\ref{app:catsprompts}). The reported component is binary,
\[
h_i=d_i,\qquad
\mathrm{BA}=\frac{1}{n_{\mathrm{BA}}}\sum_{i\in\mathcal{I}_{\mathrm{BA}}}h_i.
\]
The aggregate uses the continuous support $b_i=u_i$, which also reflects how
close the vote was: a narrow weighted majority and a unanimous panel both give
$h_i=1$ but different $b_i$.

\subsection{Factual Grounding}

Let $\mathcal{D}_i^{+}$ be the documents whose gold per-document verdict is
\textit{supports} or \textit{partially supports}. A deterministic claim extractor
returns at most eight claims per response. For claim $k$, let $\mathcal{C}_{ik}$ be its
deduplicated cited-document set. FG weights each judge by its raw
priority, since its judge schema emits no per-document confidence. With
$\mathcal{J}_{ik}$ the valid judges,
\[
P_{ik}=\!\!\sum_{j\in\mathcal{J}_{ik}}\!\!\max(1,p_j),\qquad
\tau_{ik}=\frac{P_{ik}}{2},
\]
\[
m_{ik}=\begin{cases}2,&|\mathcal{J}_{ik}|>1,\\ 1,&|\mathcal{J}_{ik}|=1.\end{cases}
\]
For eligible document $d$, let
\[
\begin{aligned}
V_{ikd}&=\sum_j\mathbf{1}[j\text{ names }d]\quad\text{and}\\
P_{ikd}&=\sum_j\max(1,p_j)\mathbf{1}[j\text{ names }d].
\end{aligned}
\]
The accepted
claim-specific support set is
\[
\mathcal{S}_{ik}=\{d\in\mathcal{D}_i^{+} : P_{ikd}>\tau_{ik} \wedge V_{ikd}\geq m_{ik}\}.
\]
For cross-document support, let $x_{ikj}$ be judge $j$'s binary vote that two
documents jointly support the claim, with mass $X_{ik}=\sum_j\max(1,p_j)x_{ikj}$
and count $U_{ik}=\sum_j x_{ikj}$. The same thresholds give
$\chi_{ik}=\mathbf{1}[X_{ik}>\tau_{ik}\wedge U_{ik}\geq m_{ik}]$ and accepted
combination set $\mathcal{D}^{\times}_{ik}$. A claim earns credit only when
accepted support is cited:
\[
y_{ik}=\mathbf{1}\big[(\mathcal{C}_{ik}\cap\mathcal{S}_{ik}\neq\emptyset)
\ \vee
\]
\[
(\chi_{ik}=1 \wedge \mathcal{C}_{ik}\cap\mathcal{D}^{\times}_{ik}\neq\emptyset)\big].
\]
For $K_i$ extracted claims,
\[
f_i=\begin{cases}K_i^{-1}\sum_{k=1}^{K_i}y_{ik},&K_i>0,\\ 0,&K_i=0,\end{cases}
\]
\[
\mathrm{FG}=\frac{1}{n_{\mathrm{FG}}}\sum_{i\in\mathcal{I}_{\mathrm{FG}}}f_i.
\]
The outer mean over examples keeps a long response from gaining weight by
yielding more claims. With all three judges, $\tau=5.5$. The priority-6 judge
alone exceeds the mass threshold but fails the two-judge corroboration rule, and
the priority-3 and priority-2 judges together (mass $5$) fall below the
threshold. Accepting a document therefore needs the priority-6 judge plus at
least one other judge, and no single judge can certify a document alone.

\begin{figure*}[t]
    \centering
    \includegraphics[width=0.9\textwidth]{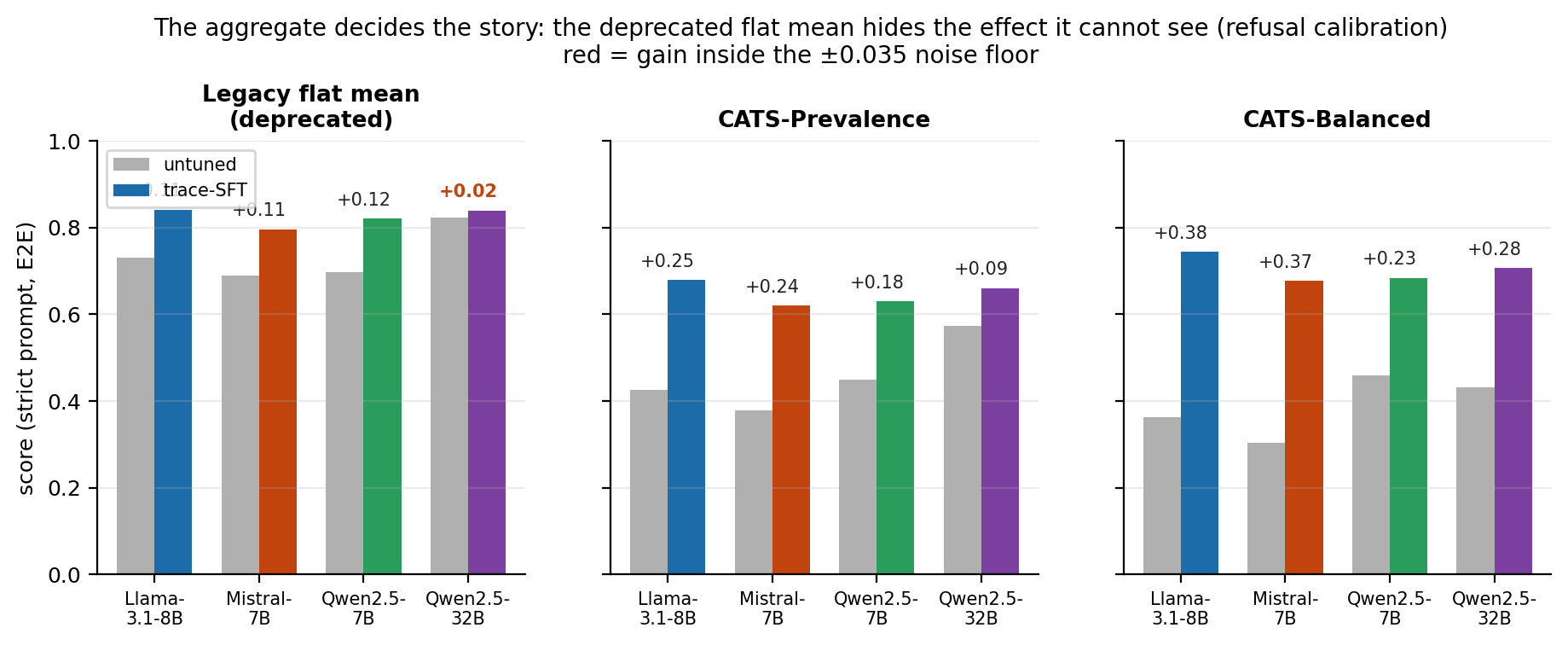}
    \caption{The choice of aggregate changes the conclusion. An unweighted mean of
    the four components almost entirely hides the refusal-calibration effect for
    Qwen2.5-32B, while CATS-Prevalence and CATS-Balanced both show it. We
    therefore report the four components separately.}
    \label{fig:aggchoice}
\end{figure*}

\begin{figure*}[t]
    \centering
    \includegraphics[width=0.9\textwidth]{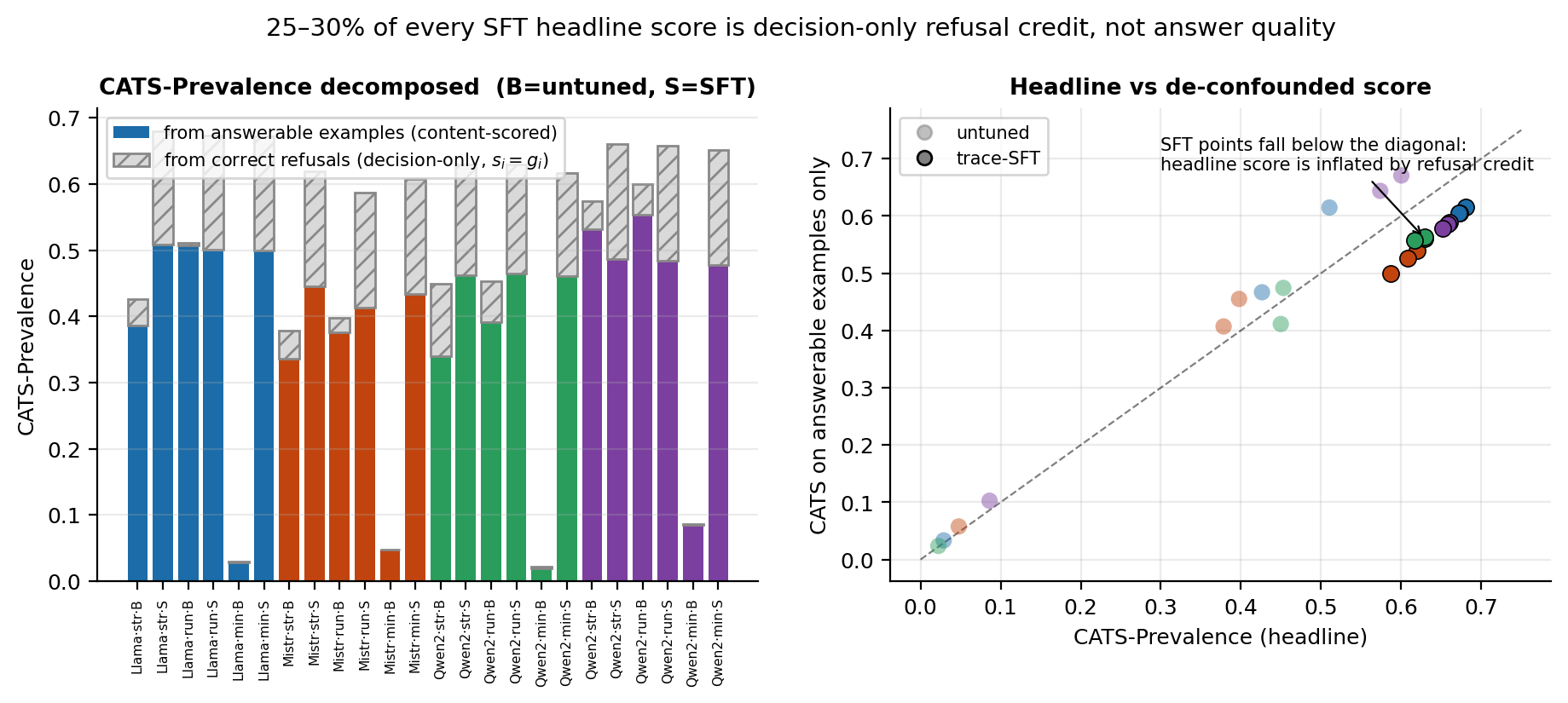}
    \caption{Refusal credit inside the aggregate. Left: CATS-Prevalence split
    into the part earned on answerable examples and the part earned on correct
    refusals, which are scored on their decision alone. Right: the answerable-only
    score plotted against the full CATS-Prevalence score. Every trace-tuned point
    lies below the diagonal, because the full score includes credit for correct
    refusal decisions.}
    \label{fig:refcredit}
\end{figure*}

\begin{figure*}[t]
\centering
\begin{minipage}[b]{2.873in}
\centering
\includegraphics[width=2.873in]{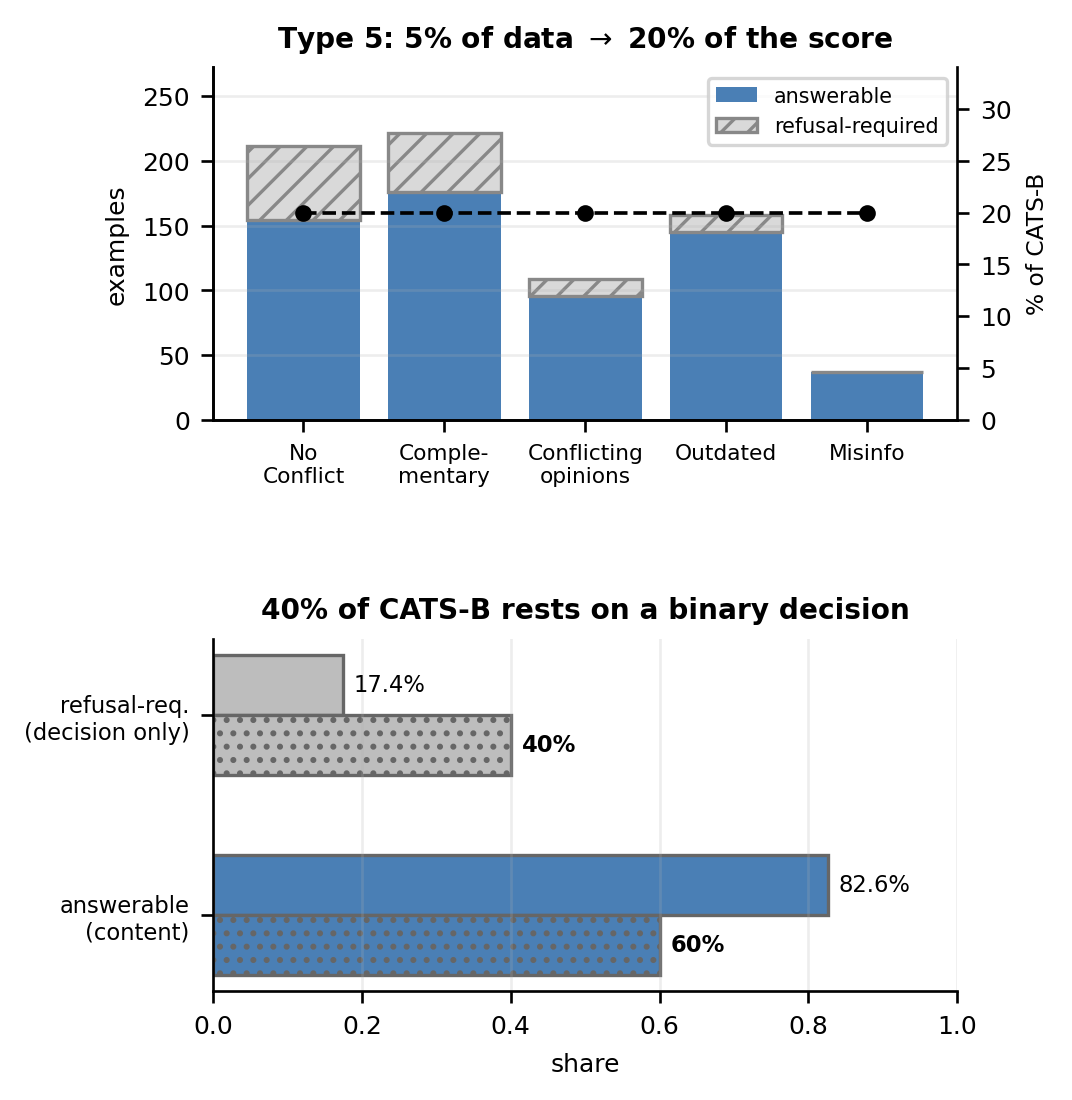}
\end{minipage}
\hfill
\begin{minipage}[b]{2.857in}
\centering
\includegraphics[width=2.857in]{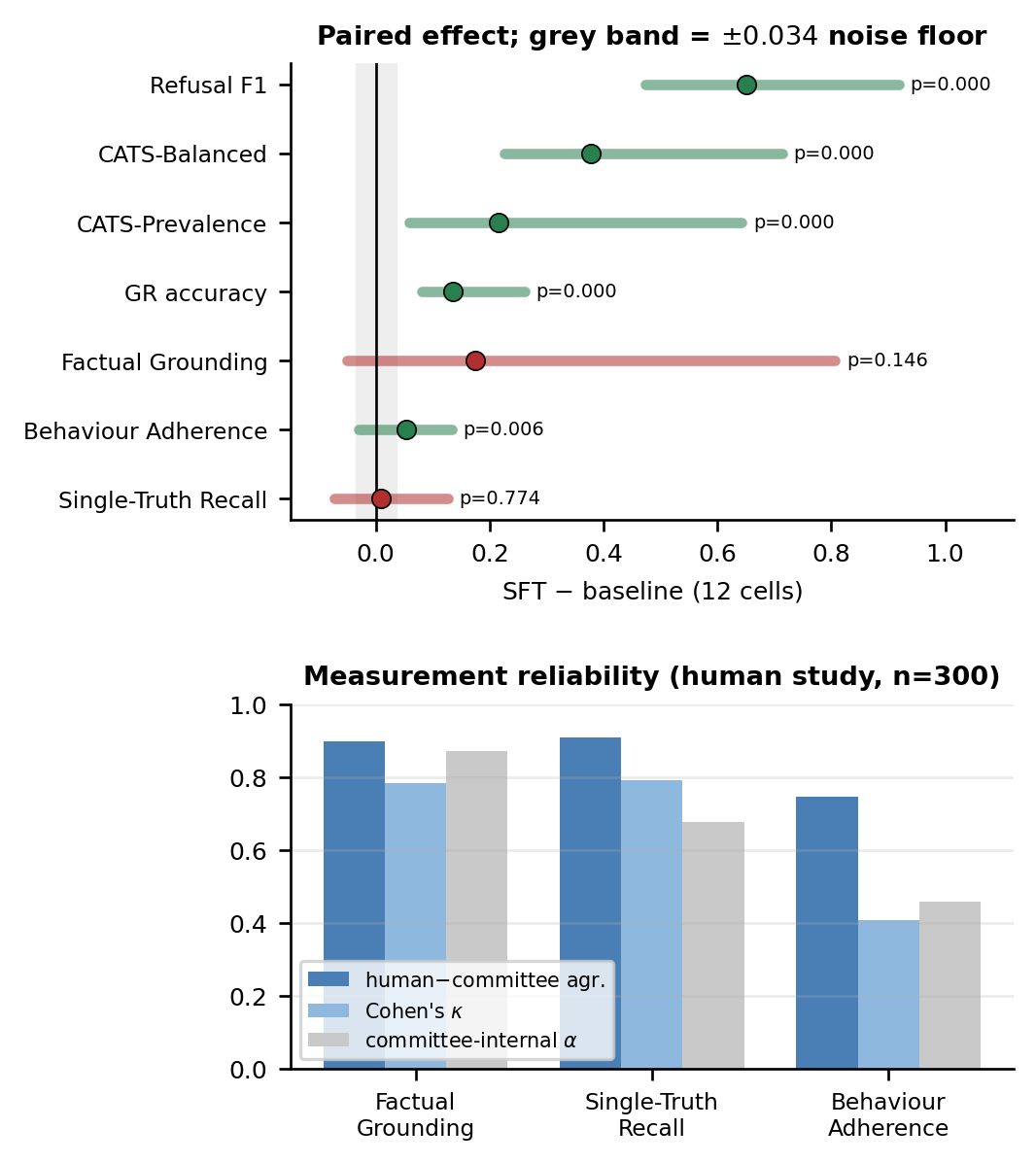}
\end{minipage}
\dualcaption{figure}{fig:weights}{fig:sigrel}%
{How CATS-B distributes its weight: misinformation is $5\%$ of the data and carries $20\%$ of the score, and refusal examples carry $40\%$}%
{Top: paired effect of trace supervision across the twelve
model$\times$prompt configurations, with the distribution-free noise floor shaded. Bottom:
per-component reliability in the human study.}
\end{figure*}

\subsection{Single-Truth Recall}

STR applies when a gold answer exists and $t_i\in\{1,2,4,5\}$. Type 3 is excluded
because conflicting-opinion cases need not have one canonical proposition. Let
$\mathcal{G}_i$ contain $M_i$ gold answer items. Applying the same weighted rule
to each item $\ell$, let $e_{i\ell}$ be the committee decision that the response
asserts that item as its own conclusion. For a negative decision the positive
side is the minority, so define a partial-credit indicator
\[
p_{i\ell}=\mathbf{1}\!\left[e_{i\ell}=0 \wedge
\frac{W_{i\ell}^{+}}{W_{i\ell}^{+}+W_{i\ell}^{-}}\geq 0.30\right],
\]
\[
E_i=\sum_{\ell}e_{i\ell},\qquad P_i=\sum_{\ell}p_{i\ell}.
\]
The per-example and dataset scores are
\[
\resizebox{\linewidth}{!}{$\displaystyle r_i=\begin{cases}
0,& M_i=0\text{ or empty candidate},\\
1,& E_i>0,\\
\min\!\left(1,\tfrac{1}{2}\tfrac{P_i}{M_i}\right),& E_i=0,\ P_i>0,\\
0,&\text{otherwise},
\end{cases}$}
\]
\[
\mathrm{STR}=\frac{1}{n_{\mathrm{STR}}}\sum_{i\in\mathcal{I}_{\mathrm{STR}}}r_i.
\]
Quoting the gold item, listing it as one possibility among several, or
attributing it to a source without adopting it does not satisfy $e_{i\ell}=1$.

\subsection{Aggregation and its structural properties}
\label{app:aggregate}

Besides the four components, CATS defines one score per example that combines
them, and two benchmark-level averages of that score, CATS-Prevalence and
CATS-Balanced. The main text reports the four components alone.

The choice of aggregate changes the conclusion. Figure~\ref{fig:aggchoice} shows
that an unweighted mean of the four components hides the effect of better refusal
calibration (a better decision on when to refuse), which both CATS aggregates
show, and Figure~\ref{fig:refcredit} splits CATS-Prevalence into the
part earned on answerable content and the part earned on the refusal decision
alone.

Each example's score is built in three steps. First, the geometric mean of
Factual Grounding and Single-Truth Recall gives the answer-quality term (Answer
Quality, AQ); for examples without a Single-Truth Recall score, AQ is Factual
Grounding alone. Second, the harmonic mean of AQ and Behaviour Adherence gives a
combined score. Third, the combined score is multiplied by the Grounded Refusal
decision, which is $1$ when the answer/refuse choice was correct and $0$
otherwise. Both means penalise imbalance, so a high score on one component cannot
make up for a low score on another, and a wrong answer/refuse decision sets the
example score to zero. CATS-Prevalence (CATS-P) is the plain mean over the
benchmark. CATS-Balanced (CATS-B) weights each conflict type equally.

Per example, the two components first form the answer-quality term
\[
q_i=\begin{cases}
\sqrt{f_i r_i},& i\in\mathcal{I}_{\mathrm{FG}}\cap\mathcal{I}_{\mathrm{STR}},\\
f_i,& i\in\mathcal{I}_{\mathrm{FG}}\setminus\mathcal{I}_{\mathrm{STR}},\\
\text{N/A},& i\notin\mathcal{I}_{\mathrm{FG}},
\end{cases}
\]
\[
\mathrm{AQ}=\frac{1}{n_{\mathrm{AQ}}}\sum_{i\in\mathcal{I}_{\mathrm{AQ}}}q_i.
\]
Let $H(x,y)=2xy/(x+y)$ for $x,y>0$ and $H(x,y)=0$ otherwise. The complete
per-example score is
\[
s_i=\begin{cases}
g_i,& A_i=0,\\
g_i\,H(b_i,q_i),& A_i=1,\ b_i,q_i\text{ available},\\
g_i b_i,& A_i=1,\ b_i\text{ only},\\
g_i q_i,& A_i=1,\ q_i\text{ only},\\
g_i,& A_i=1,\text{ neither available}.
\end{cases}
\]
The last three cases handle runs where a component is unavailable, and complete
benchmark runs use only the first two. The harmonic mean gives a low score to an
answer that is weak on either behaviour or answer quality.

To define the two aggregates, let
$\mathcal{I}^{A}_t=\{i:t_i=t,A_i=1\}$ and $\mathcal{I}^{R}_t=\{i:t_i=t,A_i=0\}$,
with subgroup means
\[
T^{A}_t=\frac{1}{|\mathcal{I}^{A}_t|}\sum_{i\in\mathcal{I}^{A}_t}s_i,
\qquad
T^{R}_t=\frac{1}{|\mathcal{I}^{R}_t|}\sum_{i\in\mathcal{I}^{R}_t}s_i.
\]
The decision-balanced score for type $t$ is
\[
T^{\mathrm{DB}}_t=\begin{cases}
\tfrac{1}{2}(T^{A}_t+T^{R}_t),& |\mathcal{I}^{A}_t|>0,\ |\mathcal{I}^{R}_t|>0,\\
T^{A}_t,& |\mathcal{I}^{A}_t|>0,\ |\mathcal{I}^{R}_t|=0,\\
T^{R}_t,& |\mathcal{I}^{A}_t|=0,\ |\mathcal{I}^{R}_t|>0,
\end{cases}
\]
and the two reported aggregates are
\[
\boxed{\ \mathrm{CATS\text{-}P}=\frac{1}{N}\sum_{i=1}^{N}s_i\ }
\]
\[
\boxed{\ \mathrm{CATS\text{-}B}=\frac{1}{5}\sum_{t=1}^{5}T^{\mathrm{DB}}_t\ }
\]
CATS-P preserves benchmark prevalence. CATS-B gives each conflict type equal
weight and, within each type where both decisions occur, gives the answerable and
refusal subgroups equal weight.

Conflict types are numbered 1 to 5 in the order of Table~\ref{tab:taxonomy}: no
conflict, complementary information, conflicting opinions or research outcomes,
outdated information and misinformation. Three properties follow from this
construction.

\paragraph{Refusal examples are scored on the answer/\allowbreak refuse decision
alone.} Grounded Refusal is computed deterministically from the
answer/refuse decision.

\paragraph{CATS-B gives refusal examples $40\%$ of the score.} CATS-B
weights answerable and refusal examples equally within each conflict type.
Refusal examples ($17.4\%$ of the data) carry $40\%$ of the score: the
misinformation type has no refusal examples, and each of the other four types
gives half its weight to its refusal examples, so the refusal share is
$4\times\tfrac12/5=40\%$. One
type-3 refusal example carries roughly twelve times the weight of one type-1
answerable example, and the misinformation class ($5\%$ of the data) carries
$20\%$ of the score.

\paragraph{CATS-B reaches $0.400$ with perfect refusals and no usable answer
content.} A model with no usable answer content and perfect refusal calibration
scores exactly $0.400$ on CATS-B (Figure~\ref{fig:weights}).

\section{Human Evaluation Study}
\label{app:human}

\subsection{Per-component reliability}

In this appendix, \emph{committee} means the judge committee of three LLM judges;
the panel of LLM annotators is always called the annotation committee. Four
reviewers judged $350$ responses generated by two of the four models. Of these,
$300$ were each reviewed independently by two of the four reviewers (the
\emph{agreement frame}) and $50$ by a single reviewer, which gives $650$ reviews
($300\times2$ plus $50$). The agreement frame is drawn from a $2\times3\times2$ design (two models
$\times$ three prompt variants $\times$ untuned/fine-tuned), stratified by
conflict type across all five types, with correct refusals excluded before
assignment. All human--human statistics are computed on this frame
(Table~\ref{tab:human}).

\begin{table*}[t]
\centering
\begin{minipage}[t]{0.485\linewidth}\vspace{0pt}
\centering
{\tabstyle\renewcommand{\arraystretch}{2.05}\setlength{\tabcolsep}{3.5pt}
\begin{tabular*}{\linewidth}[t]{@{\extracolsep{\fill}}l rr r r@{}}\toprule
& \multicolumn{2}{c}{\textbf{human--comm.}} & \textbf{h.--h.} & \textbf{judges}\\
\cmidrule(lr){2-3}\cmidrule(lr){4-4}\cmidrule(lr){5-5}
\textbf{Component} & agr. & $\kappa$ & $\kappa$ & $\alpha$\\\midrule
FG  & 0.899 & \textbf{0.783} & 0.780 & \textbf{0.871}\\
STR & 0.908 & \textbf{0.790} & 0.773 & 0.678\\
BA  & 0.745 & 0.407 & 0.669 & 0.457\\
\bottomrule\end{tabular*}}

\end{minipage}%
\hspace{0.03\linewidth}%
\begin{minipage}[t]{0.485\linewidth}\vspace{0pt}
\centering
{\tabstyle\setlength{\tabcolsep}{4.5pt}
\begin{tabular*}{\linewidth}[t]{@{\extracolsep{\fill}}l r rr@{}}\toprule
\textbf{Judge (behaviour)} & \textbf{weight} & \textbf{agr.} & $\kappa$\\\midrule
Qwen3.5-397B-A17B & 6 & 0.743 & 0.405\\
DeepSeek-R1-Distill-32B & 2 & 0.735 & 0.339\\
Mistral Small 4 & 3 & 0.643 & 0.229\\
\midrule \textbf{full committee} & & \textbf{0.745} & \textbf{0.407}\\
\midrule\multicolumn{4}{@{}p{\linewidth}@{}}{\textit{Committee-internal agreement (300-response frame)}}\\
Factual Grounding & & $\alpha$=0.871 & 0.871\\
Single-Truth Recall & & $\alpha$=0.678 & 0.687\\
Behaviour Adherence & & $\alpha$=0.457 & 0.470\\
\bottomrule\end{tabular*}}

\end{minipage}
\dualcaption{table}{tab:human}{tab:judges}%
{\textbf{Evaluator validation against human review.} Rows are Factual Grounding, Single-Truth Recall and Behaviour Adherence. Columns are human--committee agreement and $\kappa$, human--human $\kappa$, and \emph{judges}, the nominal $\alpha$ (a chance-corrected agreement coefficient for more than two raters) among the three committee models. Human--committee uses all $650$ accepted reviews. Human--human and \emph{judges} use the 300-response agreement frame. Grounded Refusal is computed by a fixed rule and needs no judges}%
{\textbf{Per-judge agreement with human reviewers, and committee-internal agreement.} Top: each judge's behaviour agreement with human reviewers over 650 human reviews (649 for Qwen3.5-397B-A17B; two per response for the 300 double-reviewed responses, one each for 50 single-reviewed responses), alongside the judge's weight in the committee vote. Qwen3.5-397B-A17B, the highest-weight judge, agrees best with humans of the three, and the full committee is slightly higher still. Bottom: agreement among the three judges on the 300-response frame. The final column is mean pairwise $\kappa$.}
\end{table*}

Human--committee agreement is high for Factual Grounding and Single-Truth Recall
($\kappa=0.78$ and $0.79$; $89.9\%$ and $90.8\%$ raw agreement), and the three
committee judges agree among themselves at $\alpha=0.87$ for grounding and
$0.68$ for recall. Behaviour Adherence has moderate agreement: the human panel agrees
with the committee on $74.5\%$ of behaviour judgments ($\kappa=0.41$), and the
three judges agree among themselves at $\alpha=0.46$.

We therefore report the Behaviour Adherence (BA) result as a direction of
improvement. BA is higher after trace supervision in $11$ of $12$
model$\times$prompt configurations ($p=0.006$; Figure~\ref{fig:sigrel}). The human panel and the committee
disagree on $25.5\%$ of behaviour labels, the complement of the $74.5\%$
agreement, and each of the twelve BA differences is smaller than that rate. No
single difference is therefore interpretable on its own, and we draw conclusions
only from the consistent direction across configurations.

Reviewers judged behaviour adherence against the per-type rubric, selected
supporting documents for each claim of the answer and judged whether the response
asserted the gold target. Each metric was judged on its own page of the review
interface, so that a factual impression could not change a behaviour label.

The $50$ singly reviewed responses came from one reviewer who completed half of an
assigned queue. They are excluded from the human--human statistics and included in
the human--committee comparison. Each statistic below uses one of three
denominators. Human--human statistics use the $300$ doubly reviewed responses.
Human--committee statistics use all $650$ reviews. Claim-level grounding is
counted in claims: $477$ for human--human and $1{,}029$ for human--committee. The
consensus comparison keeps only the responses on which both reviewers agree. It
has $256$ behaviour samples and $194$ Single-Truth Recall samples (strict
matching rule).

\paragraph{Scope.} The study covers two of the four models in the end-to-end
condition, and its reviewed responses are answers (correct refusals were excluded
before assignment). Human reviewers extracted up to twelve claims per answer; the
benchmark scorer extracts at most eight.

\begin{table*}[t]\centering{\tabstyle\setlength{\tabcolsep}{2pt}
\begin{tabular*}{\linewidth}{@{\extracolsep{\fill}}l cc cc cc cc@{}}\toprule
& \multicolumn{4}{c}{\textbf{human--human}} & \multicolumn{4}{c}{\textbf{human--committee}}\\
\cmidrule(lr){2-5}\cmidrule(lr){6-9}
& \multicolumn{2}{c}{Behaviour} & \multicolumn{2}{c}{FG (claims)} & \multicolumn{2}{c}{Behaviour} & \multicolumn{2}{c}{FG (claims)}\\
\cmidrule(lr){2-3}\cmidrule(lr){4-5}\cmidrule(lr){6-7}\cmidrule(lr){8-9}
\textbf{Conflict type} & agr. & $\kappa$ & agr. & $\kappa$ & agr. & $\kappa$ & agr. & $\kappa$\\\midrule
No conflict & 0.937 & 0.714 & 0.987 & 0.972 & 0.865 & \textbf{0.325} & 0.945 & 0.880\\
Complementary & 0.783 & 0.536 & 0.908 & 0.803 & 0.708 & \textbf{0.279} & 0.910 & 0.805\\
Conflicting opinions & 0.862 & 0.725 & 0.887 & 0.749 & 0.820 & 0.641 & 0.888 & 0.759\\
Outdated & 0.817 & 0.505 & 0.800 & 0.566 & 0.746 & \textbf{0.338} & 0.825 & 0.606\\
Misinformation & 0.864 & 0.725 & 0.912 & 0.821 & 0.581 & \textbf{0.149} & 0.929 & 0.855\\
\midrule \textbf{Overall} & 0.853 & 0.669 & 0.897 & 0.780 & 0.745 & 0.407 & 0.899 & 0.783\\
\bottomrule\end{tabular*}}
\caption{\textbf{Agreement by conflict type.} Bold marks $\kappa<0.40$. Human reviewers agree least on Complementary and Outdated behaviour. Committee--human behaviour agreement is lowest on Misinformation ($\kappa=0.149$) and Complementary ($\kappa=0.279$) and highest on Conflicting opinions ($\kappa=0.641$). Claim-level grounding agreement is high on every conflict type except Outdated. Conflicting opinions is excluded from Single-Truth Recall, because that metric needs a single correct conclusion, and so does not appear in that metric's rows.}
\label{tab:bytype}\end{table*}

\paragraph{Agreement by conflict type.} Table~\ref{tab:bytype} gives the per-type
values. Claim-level grounding agreement is highest on No-conflict
items ($\kappa=0.972$ human--human) and lowest on Outdated items
($\kappa=0.566$), where a superseded document may or may not still support a
claim. Human behaviour agreement is lowest on Complementary ($\kappa=0.536$) and
Outdated ($\kappa=0.505$). Committee--human behaviour agreement is highest on
Conflicting opinions ($\kappa=0.641$) and lowest on Misinformation
($\kappa=0.149$) and Complementary ($\kappa=0.279$).

\paragraph{Per-judge and committee-internal agreement.} Table~\ref{tab:judges}
compares each judge separately against human reviewers over the 650 human
reviews. Qwen3.5-397B-A17B (the highest-weight judge in the vote) has the highest
agreement ($0.743$ agreement, $\kappa=0.405$), DeepSeek-R1-Distill-32B is next
($0.735$, $\kappa=0.339$) and Mistral Small 4 is lowest ($0.643$,
$\kappa=0.229$). The full weighted committee reaches $\kappa=0.407$, marginally
above its best individual member. Committee-internal agreement on the
300-response frame is $\alpha=0.871$ for grounding, $0.678$ for recall and $0.457$ for
behaviour. Behaviour agreement differs between judge pairs (pairwise $\kappa$ from
$0.368$ to $0.611$).

\paragraph{Consensus frame.} Restricting to samples where both reviewers agree
and comparing that consensus with the committee raises agreement as expected.
Behaviour reaches $0.789$ ($\kappa=0.483$, $n=256$), Single-Truth Recall
(strict rule) $0.948$ ($\kappa=0.876$, $n=194$) and claim-level grounding $0.949$ ($\kappa=0.888$,
$n=428$). Consensus filtering retains easier cases disproportionately, so we
report it together with the comparison over all reviews. On Misinformation items,
behaviour agreement stays low under consensus ($0.588$, $\kappa=0.165$, against
$0.581$ and $\kappa=0.149$ over all reviews). The committee and the humans
therefore still disagree on these items when the two reviewers agree with each
other.

\paragraph{Alignment integrity.} Claim-level comparison pairs each claim
extracted by a human reviewer with the committee's judgment of the same claim.
All 650 accepted human reviews matched a judge-committee judgment, and no claim was
dropped or aligned only approximately.

\section{Comparators and Citation Metrics}
\label{app:citation}

\subsection{Prompting comparators}

\begin{figure*}[t]
    \centering
    \includegraphics[width=0.85\textwidth]{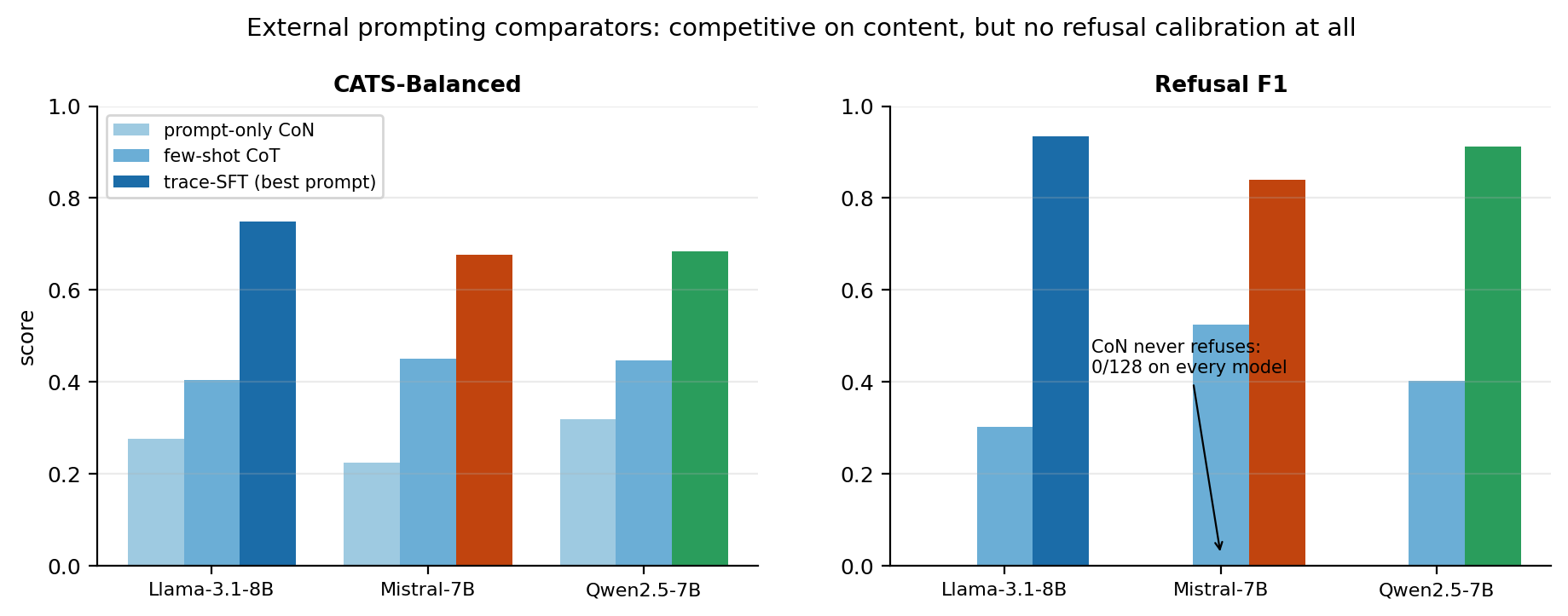}
    \caption{External prompting comparators. Both prompting-only strategies reach
    competitive answer content. Chain-of-Note records $0/128$ correct refusals on
    every model. Trace supervision teaches the answer/refuse decision
    (Table~\ref{tab:comparators}).}
    \label{fig:comparators}
\end{figure*}

We compare against two prompting-only methods, Chain-of-Note and few-shot
Chain-of-Thought (\S\ref{sec:comparators}). Few-shot Chain-of-Thought refuses on
$32/128$ refusal examples for Llama-3.1-8B, $38/128$ for Qwen2.5-7B and
$85/128$ for Mistral-7B ($25$--$66\%$). Chain-of-Note has the
poorest citation quality of any method examined: its hard-negative citation rate
(the share of cited documents that are not gold-positive) is roughly five times
that of the trace-tuned models. Figure~\ref{fig:comparators} compares the two
prompting strategies with trace supervision on answer content and on the refusal
decision.

\begin{table*}[t]
\centering
{\tabstyle\renewcommand{\arraystretch}{1.9}\setlength{\tabcolsep}{3pt}\let\frac\dfrac
\begin{tabular}[t]{@{}m{0.40\linewidth} m{0.33\linewidth} m{0.24\linewidth}@{}}\toprule
\textbf{Metric} & \textbf{Formula} & \textbf{Unit}\\\midrule
Claim-spec.\ citation prec.\ $\uparrow$ &
$\frac{\text{approved}}{\text{all cited}}$ &
cited doc\\[5pt]
Gold-positive precision $\uparrow$ &
$\frac{\text{gold-pos.}}{\text{all cited}}$ &
cited doc\\[5pt]
Citation sufficiency $\uparrow$ &
$\frac{\text{claims w/}\geq1\text{ cite}}{\text{claims w/ support}}$ &
supported claim\\[5pt]
Soft-extra citation rate $\downarrow$ &
$\frac{\text{gold-pos., not appr.}}{\text{all cited}}$ &
cited doc\\[5pt]
Hard-negative citation rate $\downarrow$ &
$\frac{\text{not gold-pos.}}{\text{all cited}}$ &
cited doc\\[5pt]
Strict grounded, gold-clean $\uparrow$ &
$\frac{\text{gr.}\wedge\text{cit.}\wedge\text{no hard-neg.}}{\text{all claims}}$ &
claim (soft extras allowed)\\[5pt]
Strict grounded, committee-clean $\uparrow$ &
$\frac{\text{gr.}\wedge\text{cit.}\wedge\text{clean}}{\text{all claims}}$ &
claim (no soft extras or hard negatives)\\
\bottomrule\end{tabular}}
\caption{\textbf{Citation-metric definitions.} ``Approved'' means the cited
document is in the committee's claim-specific support set. ``Gold-positive''
means its gold per-document verdict is \textit{supports} or \textit{partially
supports}, independent of which specific claim is being checked. ``Clean''
means no soft-extra and no hard-negative citation. A soft extra is a cited
document that is gold-positive but not approved for the claim; a hard negative is
a cited document that is not gold-positive. The two strict-grounded rates share a
base condition (committee-grounded and cited). The gold-clean rate additionally
requires no hard-negative citations; the committee-clean rate additionally
requires no soft-extra and no hard-negative citations, so it is the most
demanding metric in this section. $\uparrow$/$\downarrow$ after the metric
name marks whether higher or lower is the better direction. All fractions
count distinct cited documents or evaluated claims as noted in the Unit
column.}
\label{tab:citdefs}
\end{table*}

\begin{table}[tb]
\centering
{\tabstyle\scriptsize\setlength{\tabcolsep}{3pt}
\begin{tabular*}{\linewidth}[t]{@{\extracolsep{\fill}}lrrr@{}}\toprule
\textbf{Metric} & \textbf{mean $\Delta$} & \textbf{min} & \textbf{max}\\\midrule
Citation sufficiency (support recall) & $\mathbf{+0.363}$ & $-0.008$ & $+0.869$\\
Strict grounded, gold-clean & $\mathbf{+0.358}$ & $-0.002$ & $+0.811$\\
Strict grounded, committee-clean & $\mathbf{+0.339}$ & $\mathbf{+0.119}$ & $+0.651$\\
Gold-positive precision & $+0.003$ & $-0.004$ & $+0.014$\\
Soft-extra citation rate & $+0.006$ & $-0.058$ & $+0.074$\\
Hard-negative citation rate & $-0.003$ & $-0.014$ & $+0.004$\\
Claim-specific citation precision & $-0.003$ & $-0.077$ & $+0.055$\\
\bottomrule\end{tabular*}}
\caption{Change in each citation metric under trace supervision (trace-tuned minus
baseline): mean, minimum and maximum over the twelve model$\times$prompt
configurations.}
\label{tab:citresults}
\end{table}

\begin{figure*}[t]
\centering
\begin{minipage}[b]{1.830in}
\centering
\includegraphics[height=3.1in]{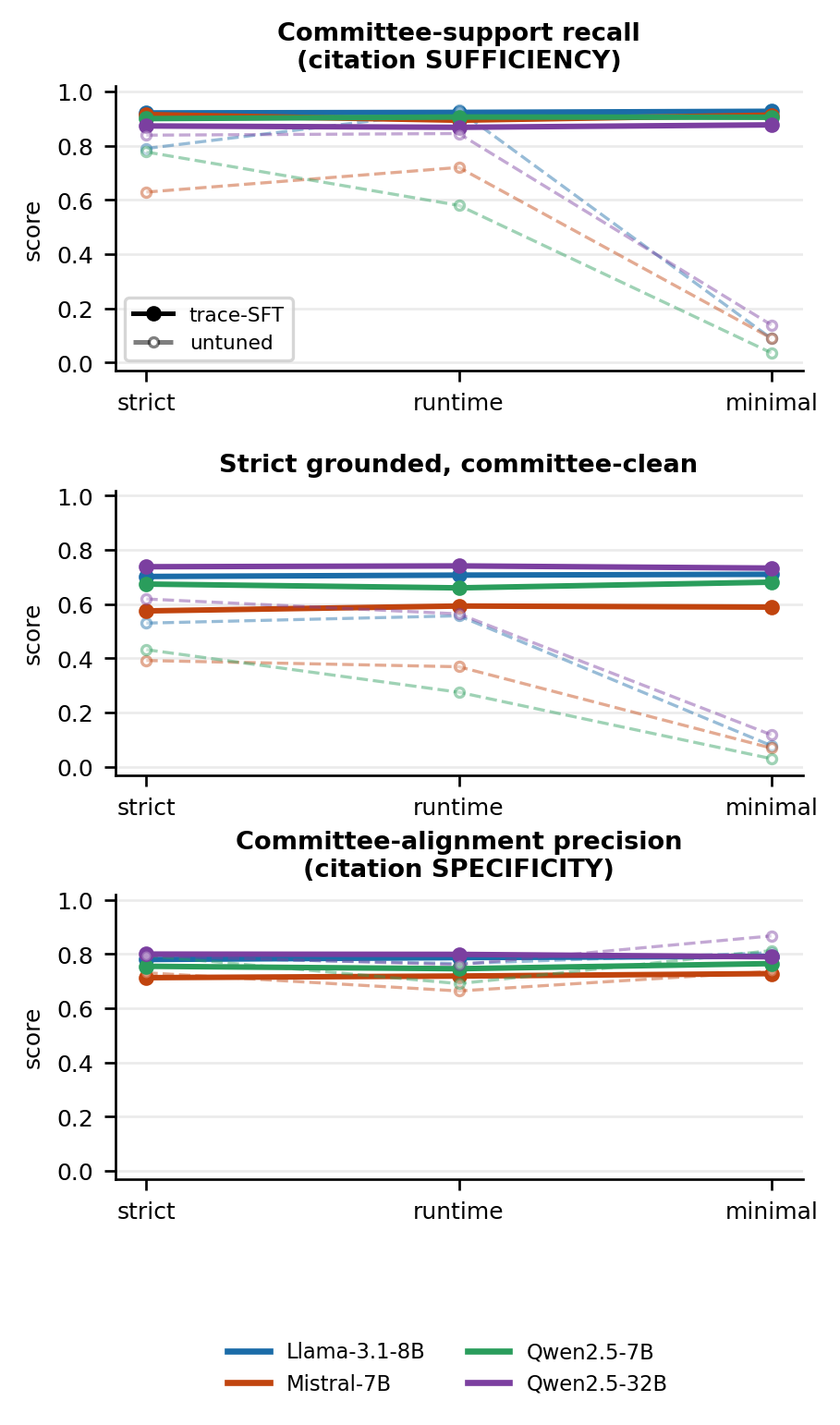}
\end{minipage}
\hfill
\begin{minipage}[b]{3.440in}
\centering
\includegraphics[height=3.1in]{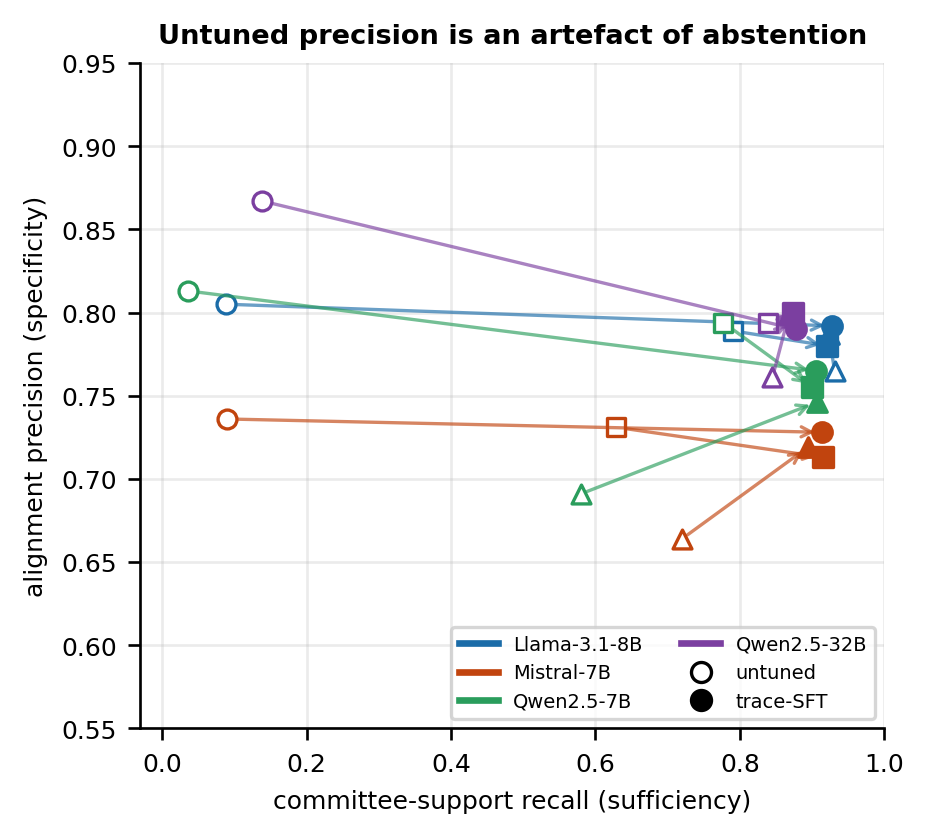}
\end{minipage}
\dualcaption{figure}{fig:cit1}{fig:cit2}%
{Trace supervision raises citation sufficiency, and mean citation precision is
essentially unchanged}%
{Axes: citation sufficiency (x) and claim-specific citation precision (y).
Open markers are baseline models and filled markers trace-tuned models, joined by a
line. Baseline models have high citation precision because they cite few
documents, and nearly every model$\times$prompt pair moves right, to higher
sufficiency, when trace-tuned.}
\end{figure*}

\subsection{Citation metrics}

We analyse cited-document behaviour with seven supplementary metrics. They are not
part of CATS, are computed after the main evaluation from the same claim-level
extraction that Factual Grounding uses, and do not change any CATS score. The
metrics split into a recall side (does the model cite the support that exists?),
made up of citation sufficiency and the two strict-grounded rates, and a precision
side (is what it cites the right evidence?), made up of claim-specific citation
precision, gold-positive precision and the soft-extra and hard-negative citation
rates.

Table~\ref{tab:citdefs} defines each metric. Table~\ref{tab:citresults}
reports the changes.

\paragraph{Unit of analysis.} The document-level metrics count \emph{distinct
cited document identifiers per claim}: if a claim cites the same document twice,
the document counts once for that claim. Each metric is a micro-average, so counts are pooled
across all evaluated claims in a run before dividing, and claims with more
citations carry proportionally more weight.

Trace supervision changes only the recall-side metrics. The strict
committee-clean metric requires a claim to be committee-supported, cited and free
of soft-extra and hard-negative citations. It improves in all twelve
model$\times$prompt configurations (minimum $+0.119$).

Baseline models under the minimal prompt have \emph{higher} claim-specific
citation precision than their trace-tuned counterparts in all four cases
($0.736$--$0.867$ against $0.713$--$0.800$), but they cite few documents
(citation sufficiency $0.036$--$0.139$), so their precision is computed over few
citations. The recall-side metrics track Factual Grounding closely
($\rho=0.93$--$0.98$). Every precision-side metric has $|\rho|\leq0.46$ against
every CATS component, so it adds information that the CATS components do not
capture. Behaviour Adherence is the CATS component least correlated with citation
behaviour, consistent with it measuring a separate property of the response.
Figure~\ref{fig:cit1} plots the sufficiency and precision trajectories across the
strict, runtime and minimal prompts, and Figure~\ref{fig:cit2} plots precision
against sufficiency for each model.

\section{Bounding Factual Grounding}
\label{app:deconfound}

\begin{figure*}[t]
    \centering
    \includegraphics[width=0.85\textwidth]{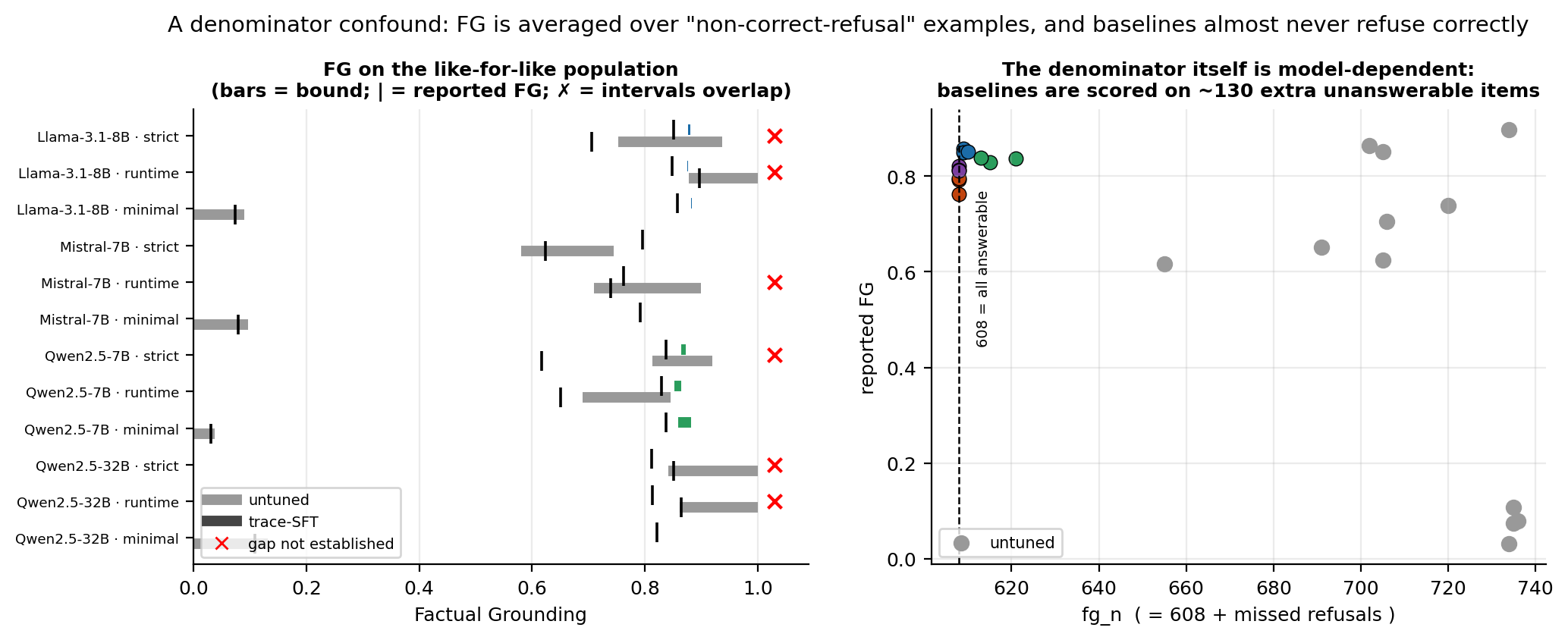}
    \caption{Factual Grounding (FG) and its denominator. Left: like-for-like FG
    intervals (bars) and reported FG (vertical marks). Crosses mark model--prompt
    pairs whose baseline and trace-tuned intervals overlap. Right: the denominator
    (the $608$ answerable examples plus the refusal examples the model
    answered) varies by model, so baseline models are scored on up to $128$
    additional refusal examples, for which the evidence is insufficient
    by design.}
    \label{fig:fgdenom}
\end{figure*}

Reported Factual Grounding (FG) is averaged over $\textit{fg\_n}=608+m$
examples: the $608$ answerable examples plus the $m$ missed refusals, meaning the
refusal examples that the model answers. Models that refuse at different rates
are therefore scored on different sets of examples.
Table~\ref{tab:fgdenom} gives, for each model and prompt, the reported FG and the
like-for-like FG interval. Figure~\ref{fig:fgdenom} plots both and shows how the
denominator varies across models.

The like-for-like interval concerns the $T$ answerable examples that the model
answered. Let $S$ be the sum of per-example FG over the $608+m$ examples above.
Missed refusals contribute between $0$ and $m$ to $S$, so the mean FG on the $T$
examples lies between $(S-m)/T$ and $S/T$ (clipped to $[0,1]$), which are the two
ends of the interval. The ends are allocation bounds and have no confidence-level interpretation. The
reported value divides $S$ by $608+m$ and counts false refusals as zero, so it can
lie below the interval: the Qwen2.5-7B baseline under the strict prompt has
$\mathrm{FG}=0.617$ against $[0.81,0.92]$. The
trace-tuned interval lies entirely above the baseline interval in all four
minimal-prompt pairs and in two others (Qwen2.5-7B runtime, Mistral-7B strict). For baseline models, FG is higher under
the 195-word runtime prompt than under the 1,965-word strict prompt in all four
families. Baseline Llama-3.1-8B under the runtime prompt reaches
$\mathrm{FG}=0.897$ while correctly refusing only $2$ of $128$ refusal items, so
its reported FG is averaged over $734$ examples and lies near the lower end of its
like-for-like interval ($[0.88,1.00]$).

\begin{table*}[t]
\centering
{\tabstyle\setlength{\tabcolsep}{1pt}
\begin{tabular*}{\linewidth}[t]{@{\extracolsep{\fill}}>{\raggedright\arraybackslash}p{0.85in}>{\raggedright\arraybackslash}p{0.50in}>{\centering\arraybackslash}p{0.35in}>{\centering\arraybackslash}p{0.35in}>{\centering\arraybackslash}p{0.75in}>{\centering\arraybackslash}p{0.75in}>{\centering\arraybackslash}p{0.40in}@{}}\toprule
& & \multicolumn{2}{c}{\textbf{reported FG}} & \multicolumn{2}{c}{\textbf{like-for-like FG}} & \\
\cmidrule(lr){3-4}\cmidrule(lr){5-6}
\textbf{Model} & \textbf{Prompt} & base & \textsc{sft} & base & \textsc{sft} & \textbf{Above}\\\midrule
Qwen2.5-7B & strict & 0.617 & 0.838 & [0.81,0.92] & [0.86,0.87] & --\\
 & runtime & 0.651 & 0.829 & [0.69,0.85] & \textbf{[0.85,0.86]} & \ding{51}\\
 & minimal & 0.031 & \textbf{0.837} & [0.00,0.04] & \textbf{[0.86,0.88]} & \ding{51}\\
\addlinespace[2pt]
Qwen2.5-32B & strict & 0.851 & 0.812 & [0.84,1.00] & [0.85,0.85] & --\\
 & runtime & 0.864 & 0.814 & [0.86,1.00] & [0.85,0.85] & --\\
 & minimal & 0.108 & \textbf{0.822} & [0.00,0.13] & \textbf{[0.86,0.86]} & \ding{51}\\
\addlinespace[2pt]
Llama-3.1-8B & strict & 0.705 & 0.852 & [0.75,0.94] & [0.88,0.88] & --\\
 & runtime & 0.897 & 0.849 & [0.88,1.00] & [0.87,0.88] & --\\
 & minimal & 0.074 & \textbf{0.857} & [0.00,0.09] & \textbf{[0.88,0.88]} & \ding{51}\\
\addlinespace[2pt]
Mistral-7B & strict & 0.624 & 0.796 & [0.58,0.74] & \textbf{[0.87,0.87]} & \ding{51}\\
 & runtime & 0.739 & 0.762 & [0.71,0.90] & [0.83,0.83] & --\\
 & minimal & 0.079 & \textbf{0.792} & [0.00,0.10] & \textbf{[0.86,0.86]} & \ding{51}\\
\addlinespace[2pt]
\bottomrule\end{tabular*}}

\caption{\textbf{The Factual-Grounding denominator is model-dependent.} The denominator is the $608$ answerable examples plus every refusal example the model answered, so a baseline model that rarely refuses is additionally scored on up to $128$ refusal examples for which the evidence is insufficient by design, while its false refusals (refusals on answerable examples) enter the denominator with $\mathrm{FG}=0$. The \emph{like-for-like} columns bound FG on the examples where the model answered \emph{and} the gold label is answerable. A check mark in the \emph{Above} column means the trace-tuned interval lies strictly above the baseline interval: this holds for every minimal-prompt row and for two of the eight strict and runtime rows (Qwen2.5-7B runtime and Mistral-7B strict).}
\label{tab:fgdenom}
\end{table*}

\begin{table}[htbp]
\centering
{\tabstyle\setlength{\tabcolsep}{1pt}
\begin{tabular*}{\linewidth}[t]{@{\extracolsep{\fill}}l cc cc@{}}\toprule
& \multicolumn{2}{c}{\textbf{FG (minimal)}} & \multicolumn{2}{c}{\textbf{Spread}}\\
\cmidrule(lr){2-3}\cmidrule(lr){4-5}
\textbf{Model} & base & \textsc{sft} & base & \textsc{sft}\\\midrule
Qwen2.5-7B & 0.031 & \textbf{0.837} & 0.442 & \textbf{0.026}\\
Qwen2.5-32B & 0.108 & \textbf{0.822} & 0.398 & \textbf{0.009}\\
Llama-3.1-8B & 0.074 & \textbf{0.857} & 0.345 & \textbf{0.016}\\
Mistral-7B & 0.079 & \textbf{0.792} & 0.278 & \textbf{0.020}\\
\bottomrule\end{tabular*}}

\caption{\textbf{Effect of shortening the prompt.} Left: Factual Grounding (FG) under the minimal prompt, which gives only the query, the documents and a 40-word evidence policy. Right: spread (highest minus lowest value) of the CATS-Balanced aggregate across the strict, runtime and minimal prompts. Baseline FG falls to $0.031$--$0.108$ under the minimal prompt, and baseline CATS-Balanced varies by $0.278$--$0.442$ across prompts, against $0.009$--$0.026$ for trace-tuned models.}
\label{tab:internalization}
\end{table}

\section{Additional Figures and Tables}
\label{app:figs}

\begin{figure*}[t]
\centering
\begin{minipage}[b]{1.801in}
\centering
\includegraphics[height=3.0in]{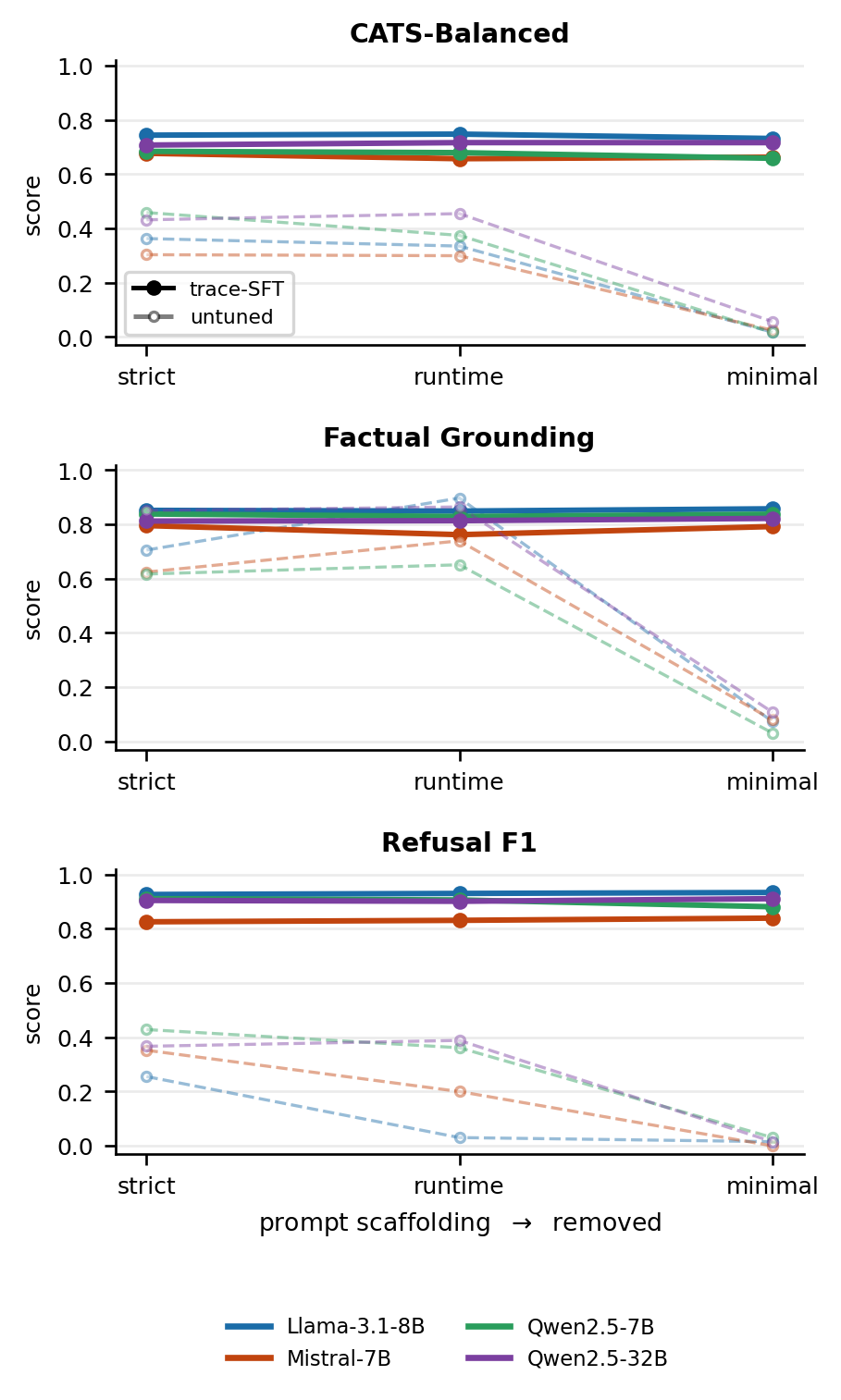}
\end{minipage}
\hfill
\begin{minipage}[b]{3.635in}
\centering
\includegraphics[height=3.0in]{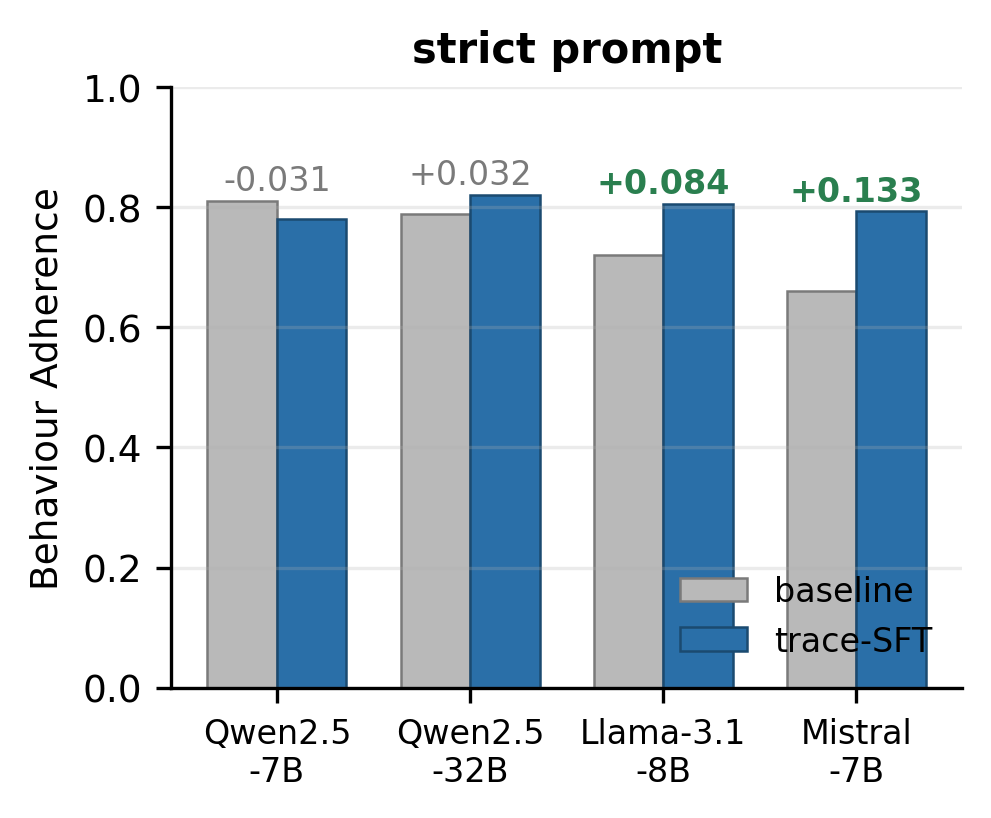}
\end{minipage}
\dualcaption{figure}{fig:internal}{fig:bastrict}%
{Performance across the strict, runtime and minimal prompts, from longest to
shortest. Solid lines are trace-tuned models, dashed lines untuned models.
Trace-tuned performance is essentially flat as the instruction is shortened,
while untuned grounding and refusal collapse}%
{Behaviour Adherence under the \textbf{strict} prompt, baseline
against trace-tuned. Differences (trace-tuned minus baseline) are annotated; bold
green marks a gain larger than the $\pm0.034$ noise floor (the 95\% interval
half-width, \S\ref{sec:setup}).}
\end{figure*}

\subsection{Behaviour Adherence by prompt}

Figures~\ref{fig:bastrict} and~\ref{fig:baminimal} give the per-model Behaviour
Adherence scores behind the summary in \S\ref{sec:refusal}, separated by
prompt. Two patterns appear.
Under the strict prompt the gain is strongly ordered by how well the baseline
already follows instructions: Mistral-7B starts lowest ($0.660$) and gains most
($+0.133$), Llama-3.1-8B gains $+0.084$, while the two Qwen baselines already sit
at $0.789$ and $0.811$ and move $+0.032$ and $-0.031$. Only the first two clear
the $\pm0.034$ noise floor. Under the strict prompt, trace supervision therefore
helps most the baselines that follow instructions least well.

Under the minimal prompt the picture is more uniform. Every model gains
($+0.031$ to $+0.062$, although the smallest gain lies within the $\pm0.034$ noise
floor) and the baselines compress into a narrow band ($0.727$--$0.747$), which
suggests that a forty-word instruction that never names the taxonomy removes the
advantage the stronger instruction-followers had.
Qwen2.5-7B, whose Behaviour Adherence drops under the strict prompt, gains here.
This matches the prompt-invariance result in \S\ref{sec:internalization} for
Behaviour Adherence. The strict prompt gives baseline models the conflict taxonomy in the
instruction, and trace supervision gives it to the model during training, which
would explain why trace-tuned models score similarly under the strict and minimal
prompts while baselines differ between them. Behaviour Adherence agrees with human review at
$\kappa=0.41$ (\S\ref{sec:human}), so we read these per-model margins as showing a
consistent direction.

\begin{figure*}[t]
\centering
\begin{minipage}[b]{2.908in}
\centering
\includegraphics[height=2.4in]{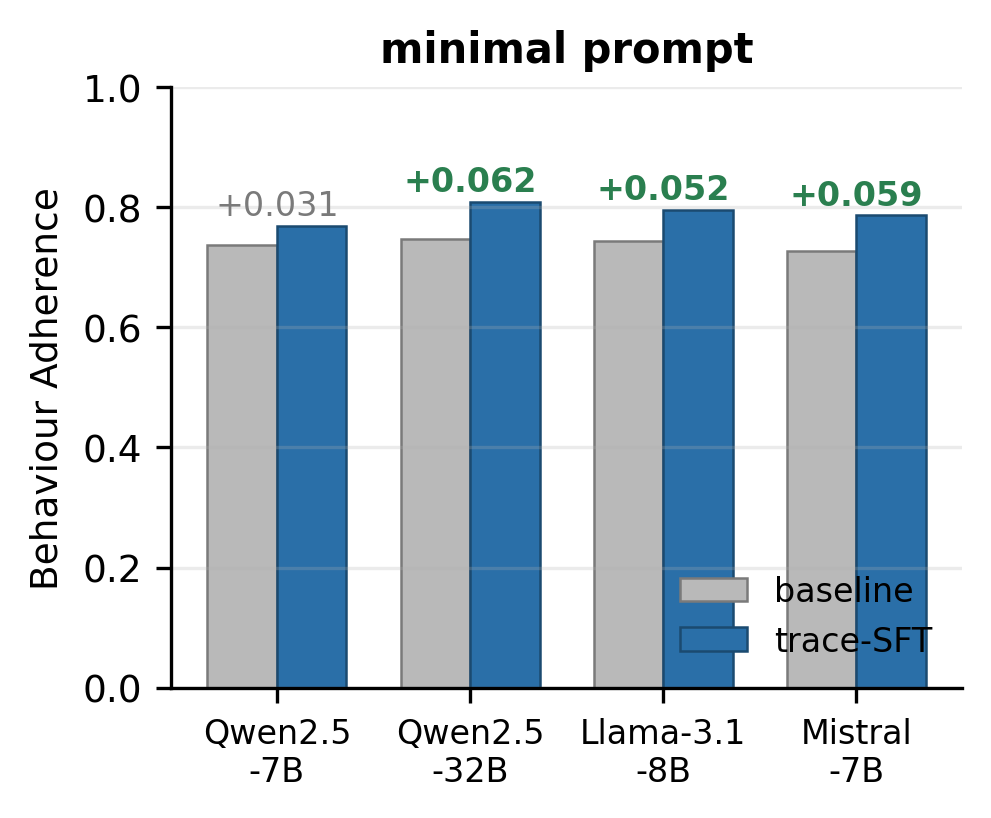}
\end{minipage}
\hfill
\begin{minipage}[b]{2.663in}
\centering
\includegraphics[height=2.4in]{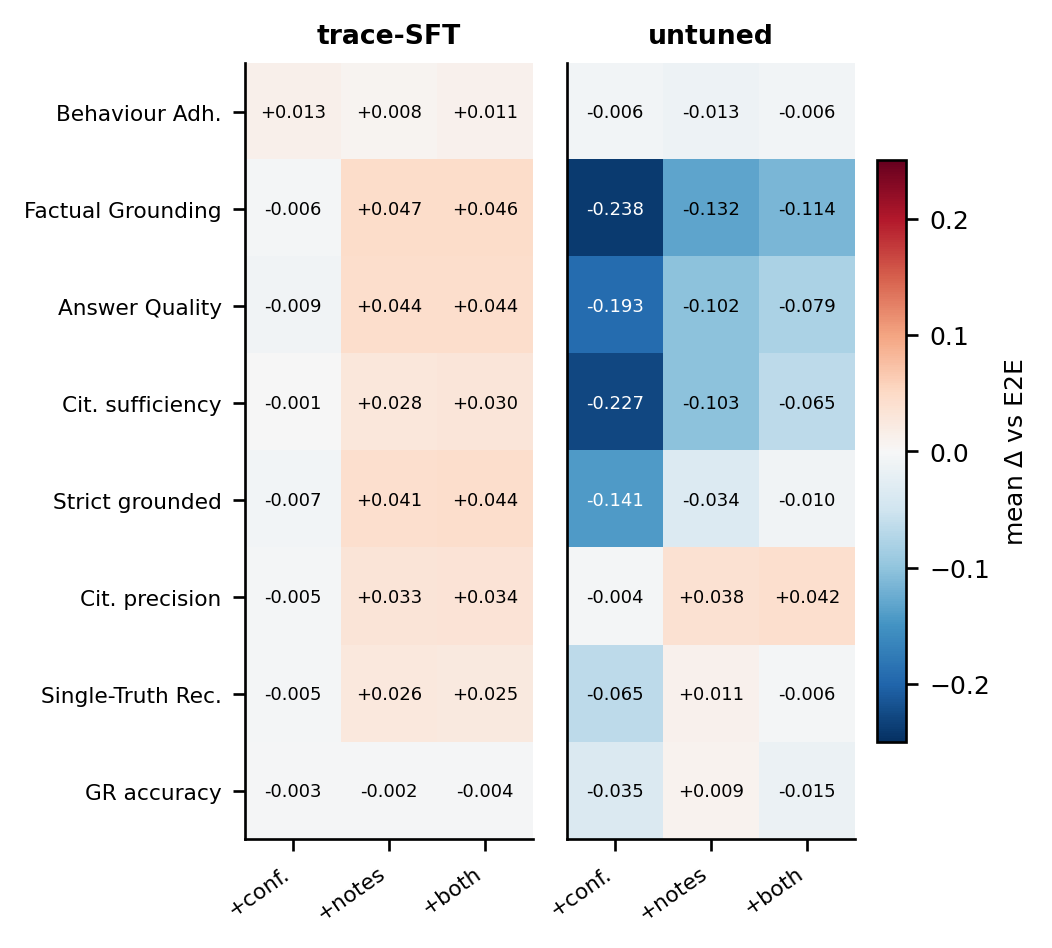}
\end{minipage}
\dualcaption{figure}{fig:baminimal}{fig:oracleheat}%
{Behaviour Adherence under the \textbf{minimal} prompt. Baselines
compress into a narrow band once the instruction no longer names the
taxonomy, and every model gains}%
{Mean change in every metric when gold information is supplied (an
\emph{oracle}: the gold conflict label, the gold per-document notes, or both).
Trace-tuned models gain on the metrics each oracle targets. Baseline models mostly decline,
most on grounding under the conflict-label oracle.}
\end{figure*}

The runtime prompt confirms the pattern for trace-tuned models. In all twenty
model$\times$metric combinations, the runtime score lies within $0.030$ of the range spanned by the strict and minimal scores, inside
the $\pm0.034$ noise floor. Single-Truth Recall shows the same pattern for
trace-tuned models, while baseline scores vary by $0.047$--$0.196$ across the
prompts. Citation sufficiency, a supplementary metric defined separately from CATS
(Appendix~\ref{app:citation}), follows it too: baseline sufficiency varies by $0.730$ across the three prompts and
fine-tuned sufficiency by $0.011$, about sixty-five times less.

\section{Oracle Ablations: Full Results}
\label{app:oracle}

\begin{table*}[t]\centering{\tabstyle\setlength{\tabcolsep}{5.5pt}
\begin{tabular*}{\linewidth}{@{\extracolsep{\fill}}l ccc ccc ccc@{}}\toprule
& \multicolumn{3}{c}{\textbf{+ gold conflict label}} & \multicolumn{3}{c}{\textbf{+ gold per-doc notes}} & \multicolumn{3}{c}{\textbf{+ both}}\\
\cmidrule(lr){2-4}\cmidrule(lr){5-7}\cmidrule(lr){8-10}
\textbf{Metric} & mean $\Delta$ & $+/n$ & $p$ & mean $\Delta$ & $+/n$ & $p$ & mean $\Delta$ & $+/n$ & $p$\\\midrule
Behaviour Adherence & \textbf{+0.013} & 8/9 & \textbf{0.039} & +0.008 & 10/12 & \textbf{0.039} & +0.011 & 9/12 & 0.146\\
Factual Grounding & -0.006 & 3/9 & 0.508 & \textbf{+0.047} & 12/12 & \textbf{0.000} & \textbf{+0.046} & 12/12 & \textbf{0.000}\\
Answer Quality & -0.009 & 2/9 & 0.180 & \textbf{+0.044} & 12/12 & \textbf{0.000} & \textbf{+0.044} & 12/12 & \textbf{0.000}\\
Citation sufficiency & -0.001 & 3/9 & 0.508 & \textbf{+0.028} & 12/12 & \textbf{0.000} & \textbf{+0.030} & 12/12 & \textbf{0.000}\\
Strict grounded (cit.) & -0.007 & 3/9 & 0.508 & \textbf{+0.041} & 12/12 & \textbf{0.000} & \textbf{+0.044} & 12/12 & \textbf{0.000}\\
Citation precision & -0.005 & 2/9 & 0.180 & \textbf{+0.033} & 12/12 & \textbf{0.000} & \textbf{+0.034} & 12/12 & \textbf{0.000}\\
Single-Truth Recall & -0.005 & 3/9 & 0.508 & +0.026 & 9/12 & 0.146 & +0.025 & 8/12 & 0.388\\
Decision accuracy & -0.003 & 2/9 & 0.180 & -0.002 & 5/12 & 0.774 & -0.004 & 5/12 & 0.774\\
\bottomrule\end{tabular*}}
\caption{\textbf{Oracle ablations on the fine-tuned models.} Each intervention supplies gold information for one reasoning stage; a metric is \emph{targeted} if it depends on that stage. The gold conflict label moves Behaviour Adherence and nothing else; the gold per-document notes move every grounding and citation metric in $12/12$ model$\times$prompt configurations and change mean decision accuracy by at most $0.004$. $p$ is an exact two-sided sign test over model$\times$prompt configurations. The conflict-oracle columns cover Llama-3.1-8B, Mistral-7B and Qwen2.5-7B; the $+/n$ column reports the exact number of configurations contributing to each test ($n=9$ for the conflict oracle, $n=12$ otherwise).}
\label{tab:oracle}\end{table*}

\subsection{Additivity, exceptions and baselines}

If the two oracles act independently, the gain from supplying both should equal
the sum of the gains from each. Using the mean $\Delta$ values in
Table~\ref{tab:oracle}, the difference
$|\Delta_{\text{both}}-(\Delta_{\text{conflict}}+\Delta_{\text{notes}})|$ is at
most $0.010$ on every grounding metric and on Behaviour Adherence alike. The
conflict-oracle column averages over nine model$\times$prompt configurations
(Llama-3.1-8B, Mistral-7B and Qwen2.5-7B), and the notes and both columns over
twelve.

The oracle effects on the metrics each oracle targets are consistent across
model$\times$prompt configurations (the $+/n$ columns of Table~\ref{tab:oracle}).
With both oracles supplied, the model still produces Stage~3 (grounded
synthesis) on its own.

For baseline models the oracles lower most metrics, and the conflict label lowers
grounding by $0.238$ (Figure~\ref{fig:oracleheat}). The trace-tuned models, which
were trained on the stage outputs, gain consistently on the targeted metrics.

\begin{figure*}[t]
    \centering
    \includegraphics[width=\textwidth]{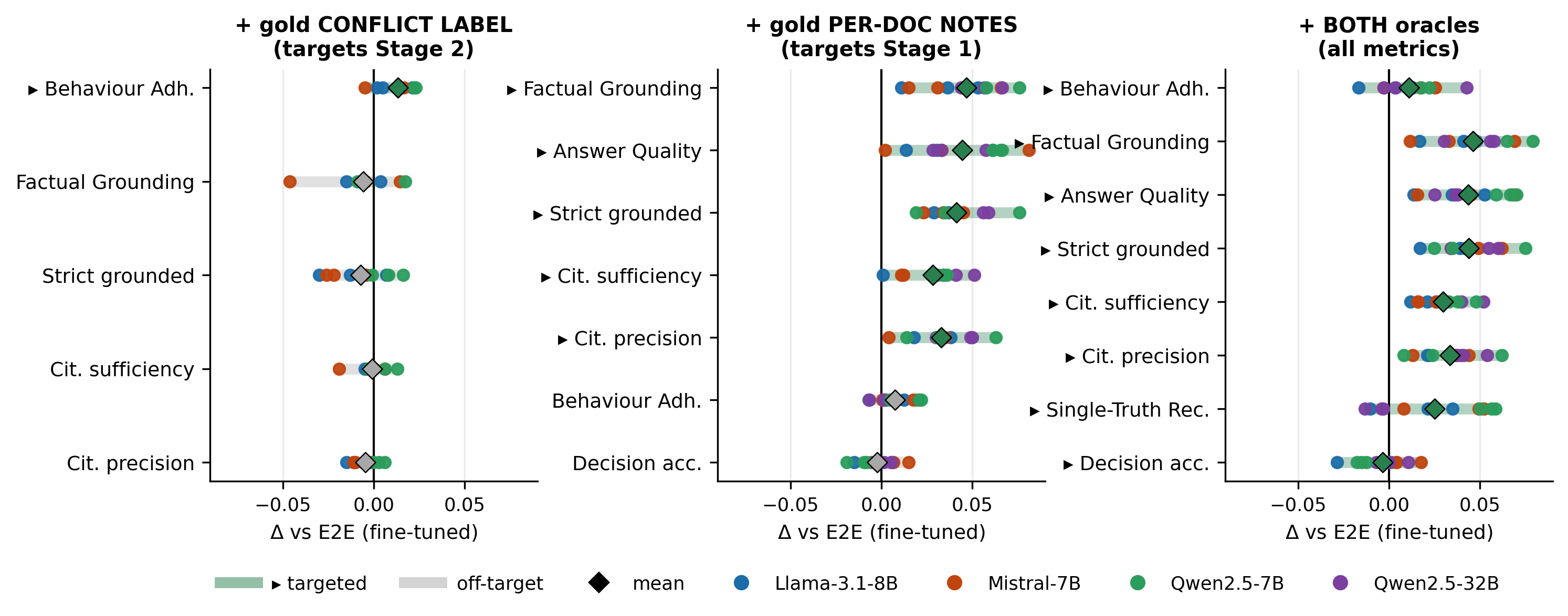}
    \caption{Each oracle (gold information supplied to the model) changes the metric
    it targets and leaves the others essentially unchanged. Green bars are
    targeted metrics, the metrics that depend on the stage the oracle supplies.
    Grey bars are control metrics that the oracle should not affect. Diamonds mark
    means and points mark individual model$\times$prompt runs.}
    \label{fig:oraclefig}
\end{figure*}

\section{Released Artifacts}
\label{app:annprompts}

We release the full resource and the prompt suite used to build and evaluate it.
The release comprises the \textbf{training/validation split} (943 examples) and
the \textbf{held-out benchmark} (736 examples), together carrying $10{,}343$
labeled documents. Each document has a per-document verdict, a source-quality label and the
complete annotator vote tally; non-irrelevant documents also have an extracted
key fact, a supporting quote taken from the snippet and a verdict rationale. The release also
includes the \textbf{stagewise annotation prompts} (a system prompt and user
template for each of the three annotation stages),
the \textbf{end-to-end inference prompts} for the strict, runtime and minimal
settings, the \textbf{CATS judge prompts}, and the per-run evaluation outputs
from which every number in this paper is computed.

Records are distributed as JSONL with dataset cards documenting per-source
provenance and the upstream licence for each source family. All artifacts are
available at \nolinkurl{https://anonymous.4open.science/r/rag_reason-2F8F/}.

\makeatletter
\let\PromptFileMain\PromptFile
\newcommand{\PFNeed}[1]{\par\begingroup\dimen@=#1\relax\dimen@ii=\pagegoal\advance\dimen@ii-\pagetotal
  \ifdim\dimen@>\dimen@ii\break\fi\endgroup}
\newcommand{\PF@box}[3]{%
  {\raggedright\paragraph{#2}\leavevmode\par}\nobreak\vspace{-\parskip}\nobreak
  \begin{tcolorbox}[mypromptstyle,top=2pt,bottom=2pt,boxsep=1pt,#1,before skip=10pt plus 2pt,after skip=8pt plus 8pt minus 2pt]
  \VerbatimInput[fontsize=\scriptsize,baselinestretch=0.94,breaklines=true,breakanywhere=false,breakafter=\#,%
    breaksymbolleft={},breaksymbolright={},breaksymbolsepleft=0pt,breaksymbolsepright=0pt]{#3}
  \end{tcolorbox}}
\renewcommand{\PromptFile}[2]{\PF@box{}{#1}{#2}}
\newcommand{\PromptFileShort}[2]{\PF@box{unbreakable}{#1}{#2}}
\makeatother

\section{End-to-End Inference Prompts}
\label{app:e2eprompts}

\subsection{Why the prompt is varied}

Varying the prompt tests whether conflict-aware behaviour has been
\emph{learned} or follows from the instruction. A model that produces the trace
only when the strict prompt describes it may be following the prompt. A model
that still produces it under the minimal prompt, which never mentions the trace,
the taxonomy or the citation format, has learned a mapping from retrieval context
to behaviour. Training and evaluating across all three variants therefore
measures this directly, and \S\ref{sec:internalization} reports the resulting
gap.

The three end-to-end prompts (strict, runtime and minimal) are reproduced
verbatim below. Each is a system prompt plus a user template. The placeholders
are filled with the query and the retrieved documents at generation time.

The three prompts give progressively less instruction. The retrieved set and the
expected target are identical across them (Table~\ref{tab:promptladder}):

\begin{table}[htbp]\centering{\tabstyle\setlength{\tabcolsep}{3pt}
\begin{tabular*}{\linewidth}{@{\extracolsep{\fill}}l@{}}\toprule
\textbf{Condition, length and what the prompt still states}\\\midrule
\textbf{strict} --- $1{,}965$ words ($49\times$)\\
\hspace{1em}full taxonomy, verdict heuristics, threshold and\\
\hspace{1em}edge-case rules, citation policy, output format\\[3pt]
\textbf{runtime} --- $195$ words ($5\times$)\\
\hspace{1em}three-stage schema, allowed labels, abstention\\
\hspace{1em}and composition rules, citations, end marker\\[3pt]
\textbf{minimal} --- $40$ words ($1\times$)\\
\hspace{1em}evidence-only answering and when to abstain.\\
\hspace{1em}Nothing else\\
\bottomrule\end{tabular*}}
\caption{The three end-to-end prompts, ordered by system-prompt length.}
\label{tab:promptladder}
\end{table}

The minimal system prompt is four sentences long and \emph{never mentions} the
think trace, the three stages, the conflict taxonomy, the citation format or
the end-of-answer marker. A trace-tuned model nonetheless emits all of them, which
is what \S\ref{sec:internalization} measures. Under this prompt, baseline grounding
falls to $0.031$--$0.108$ while trace-tuned grounding is $0.792$--$0.857$, with
both given the same forty words.

The oracle conditions of \S\ref{sec:setup} add a block with the gold conflict
label, the gold per-document notes, or both to each prompt. Those templates are
released with the artifacts and differ from the prompts below only by that block.

\PFNeed{6\baselineskip}
\subsection{Strict}
\PromptFile{Strict system prompt (\texttt{system\_e2e.txt})}{inference_prompts/system_e2e.txt}
\PromptFileShort{Strict user template (\texttt{user\_e2e.txt})}{inference_prompts/user_e2e.txt}

\PFNeed{6\baselineskip}
\subsection{Runtime}
\PromptFile{Runtime system prompt (\texttt{system\_e2e\_runtime.txt})}{inference_prompts/system_e2e_runtime.txt}
\PromptFileShort{Runtime user template (\texttt{user\_e2e\_runtime.txt})}{inference_prompts/user_e2e_runtime.txt}

\PFNeed{6\baselineskip}
\subsection{Minimal}
\PromptFileShort{Minimal system prompt (\texttt{system\_e2e\_minimal.txt})}{inference_prompts/system_e2e_minimal.txt}
\PromptFileShort{Minimal user template (\texttt{user\_e2e\_minimal.txt})}{inference_prompts/user_e2e_minimal.txt}

\FloatBarrier

\section{CATS Judge Prompts}
\label{app:catsprompts}

The complete judge bundle is reproduced verbatim below: one shared system
message, the three task prompts, and the five-type behaviour rubric that the
Behaviour Adherence prompt selects from. Placeholders in braces are filled at
evaluation time. Grounded Refusal needs no prompt.

\paragraph{Behaviour is isolated from the other components.} The Behaviour
Adherence prompt instructs the judge to ignore factual correctness and states
that answerability correctness, factual entailment, citation validity,
unsupported-claim detection and single-truth recall must not be used as hidden
criteria. It supplies exactly one rubric line, for the sample's own conflict type. This
keeps Behaviour Adherence separate from the other three components, and
empirically it is the CATS component least correlated with citation behaviour
(Appendix~\ref{app:citation}).

\paragraph{Factual Grounding is permissive, so it is an upper bound on strict
grounding.} The prompt enumerates seven distinct routes to support, including
paraphrase, entailment, more-specific instances, more-general statements,
containing ranges and partial coverage of a claim's substance. It also instructs
the judge to ``bias toward inclusion when in doubt''.

\paragraph{Provenance of the per-document verdicts in the grounding prompt.} The
grounding prompt calls the per-document verdicts it shows the judge ``ground truth
annotations made by human experts''. The annotation committee, a panel of LLM
annotators (Appendix~\ref{app:annotation}), produced these verdicts. Human review
covered the conflict-type label, the benchmark preselection fields and a 100-pair
sample of the verdicts (Appendix~\ref{app:human-review}). The phrase tells the
judge to take the verdicts as given and check only the claim, so we expect little
effect on the reported grounding scores.

\subsection{Shared system message}
\PromptFileShort{Committee system message (\texttt{committee\_json\_system\_prompt.txt})}{cats_prompts/committee_json_system_prompt.txt}

\PFNeed{6\baselineskip}
\subsection{Behaviour Adherence}
\PromptFile{Behaviour Adherence prompt (\texttt{behavior\_adherence\_prompt.\allowbreak template.txt})}{cats_prompts/behavior_adherence_prompt.template.txt}
\PromptFile{Five-type behaviour rubric (\texttt{behavior\_rubric.md})}{cats_prompts/behavior_rubric.md}

\PFNeed{6\baselineskip}
\subsection{Factual Grounding}
\PromptFile{Factual Grounding prompt (\texttt{factual\_grounding\_prompt.\allowbreak template.txt})}{cats_prompts/factual_grounding_prompt.template.txt}

\PFNeed{6\baselineskip}
\subsection{Single-Truth Recall}
\PromptFile{Single-Truth Recall prompt (\texttt{single\_truth\_recall\_prompt.\allowbreak template.txt})}{cats_prompts/single_truth_recall_prompt.template.txt}

\FloatBarrier

\section{Complete Agreement Statistics}
\label{app:allagreement}

The tables in this section report every agreement statistic for the resource
reviews and the judge validation, with denominators and confusion counts. The
Stage-1 verdict study is in Appendix~\ref{app:stage1review}. Raw agreement and $\kappa$ are
reported together, because $\kappa$ is near zero when almost all items share one
label. A dash in a $\kappa$ column marks a field with a single value across its
subset, where $\kappa$ is undefined and the rows report exact agreement. No missing
judgment is imputed. Agreement among the three LLM judges is reported separately
from human agreement.

Table~\ref{tab:catsbinfull} gives binary agreement for four comparisons (human vs
human, human vs committee, human consensus vs committee, human vs each judge), by
subset where the comparison defines one. Human behaviour agreement ranges from
$0.819$ to $1.000$ across reviewer pairs (the \textsc{r4} pairs cover only $10$--$21$
responses).

Tables~\ref{tab:catsbinfull} and~\ref{tab:catscontfull} use the following notation: $P_{pos}$ and $P_{neg}$ are positive- and
negative-specific agreement, $n_{11}$--$n_{00}$ are the confusion counts, and
$\alpha$ is nominal Krippendorff's alpha where computed. Human--human rows are
response-level over the 300-response frame, and human--committee rows are
judgment-level and pool all accepted reviews.

Table~\ref{tab:catscontfull} compares the grounding ratio, the fraction of an
answer's claims judged grounded, as a continuous value. Two reviewers can agree
on most claims and still differ on the ratio when their extracted claim sets
differ in size.

Table~\ref{tab:cifull} reports committee-internal agreement on the
300-response frame. Agreement is highest for grounding ($\alpha \approx 0.87$),
intermediate for recall ($\approx 0.68$) and lowest for behaviour ($\approx
0.46$). Behaviour agreement differs between judge pairs, which matches the
per-judge comparison with human reviewers in Table~\ref{tab:judges}.

\onecolumn
\begin{multicols}{2}
Table~\ref{tab:iaafull} covers both resource-review workflows. The retain-or-change
and confidence fields of the training review have low $\kappa$ ($0.1341$ and
$0.0159$) because most reviewers kept the committee label and reported
high confidence, so we treat the five-way conflict-type field ($\kappa=0.7694$) as
primary. \textsc{r1}--\textsc{r4} are the four reviewers of the human study
(Appendix~\ref{app:human}). Field names follow the released review files:
\texttt{label\_action} and \texttt{changed\_label} record whether the reviewer kept
the annotation-committee label, and \texttt{human\_preselect\_decision} is the
retention decision.

\end{multicols}

{\tabstyle\setlength{\tabcolsep}{3pt}\def\LTcaptype{}
\begin{longtable}{@{}>{\raggedright\arraybackslash}p{80pt}>{\raggedright\arraybackslash}p{134pt}>{\raggedleft\arraybackslash}p{36pt}>{\raggedleft\arraybackslash}p{44pt}>{\raggedleft\arraybackslash}p{40pt}>{\raggedleft\arraybackslash}p{42pt}>{\raggedleft\arraybackslash}p{41.5pt}@{}}
\toprule \textbf{Workflow / scope} & \textbf{Field} & $n$ & \textbf{agree} & \textbf{dis.} & \textbf{raw} & $\kappa$\\\midrule\endfirsthead
\toprule \textbf{Workflow / scope} & \textbf{Field} & $n$ & \textbf{agree} & \textbf{dis.} & \textbf{raw} & $\kappa$\\\midrule\endhead
\bottomrule\endfoot
\multicolumn{7}{@{}l}{\textit{Training conflict-type review (943 records, two reviewers each from a pool of three)}}\\
& reviewed\_conflict\_type & 943 & 787 & 156 & 0.835 & 0.7694\\
& label\_action & 943 & 791 & 152 & 0.839 & 0.1341\\
& changed\_label & 943 & 791 & 152 & 0.839 & 0.1341\\
& review\_confidence & 943 & 607 & 336 & 0.644 & 0.0159\\
\addlinespace\multicolumn{7}{@{}l}{\textit{Benchmark preselection (1,454 records, 7 reviewers)}}\\
& human\_preselect\_decision & 1454 & 1428 & 26 & 0.982 & 0.9228\\
& preliminary\_conflict\_type & 1454 & 1378 & 76 & 0.948 & 0.9217\\
& preselection\_confidence & 1454 & 1445 & 9 & 0.994 & 0.9767\\
& retrieval\_quality & 1454 & 1436 & 18 & 0.988 & 0.9550\\
& evidence\_sufficiency & 1454 & 1436 & 18 & 0.988 & 0.9600\\
& conflict\_clarity & 1454 & 1441 & 13 & 0.991 & 0.9660\\
& query\_specificity & 1454 & 1449 & 5 & 0.997 & 0.9875\\
& source\_reliability & 1454 & 1446 & 8 & 0.994 & 0.9881\\
& relevant\_doc\_count\_bin & 1454 & 1426 & 28 & 0.981 & 0.9527\\
& gold\_answer\_possible & 1454 & 1440 & 14 & 0.990 & 0.9669\\
\addlinespace\multicolumn{7}{@{}l}{\textit{Refusal-quality subset (128 released refusal examples)}}\\
& human\_preselect\_decision & 128 & 128 & 0 & 1.000 & --\\
& preliminary\_conflict\_type & 128 & 128 & 0 & 1.000 & 1.0000\\
& preselection\_confidence & 128 & 128 & 0 & 1.000 & --\\
& retrieval\_quality & 128 & 128 & 0 & 1.000 & --\\
& evidence\_sufficiency & 128 & 128 & 0 & 1.000 & --\\
& conflict\_clarity & 128 & 128 & 0 & 1.000 & --\\
& query\_specificity & 128 & 128 & 0 & 1.000 & --\\
& source\_reliability & 128 & 128 & 0 & 1.000 & --\\
& relevant\_doc\_count\_bin & 128 & 128 & 0 & 1.000 & --\\
& gold\_answer\_possible & 128 & 128 & 0 & 1.000 & --\\
& refusal\_required & 128 & 128 & 0 & 1.000 & --\\
& refusal\_ground\_truth\_valid & 128 & 128 & 0 & 1.000 & --\\
& refusal\_rationale\_quality & 128 & 128 & 0 & 1.000 & --\\
& refusal\_quality\_label & 128 & 128 & 0 & 1.000 & --\\
\end{longtable}}\par\nopagebreak[4]\vskip-9pt
\captionof{table}{\textbf{Complete resource-review agreement statistics.} Every recorded field in both review workflows. $n$ is the number of review pairs; \emph{agree}/\emph{disagree} are exact counts. A dash in the $\kappa$ column marks a field with a single value across the subset.}\label{tab:iaafull}\par\vskip 7pt

\FloatBarrier

{\tabstyle\setlength{\tabcolsep}{3pt}\def\LTcaptype{}
\begin{longtable}{@{}>{\raggedright\arraybackslash}p{66pt}>{\raggedright\arraybackslash}p{40pt}>{\raggedright\arraybackslash}p{85pt}>{\raggedleft\arraybackslash}p{19pt}>{\raggedleft\arraybackslash}p{21.5pt}>{\raggedleft\arraybackslash}p{21.5pt}>{\raggedleft\arraybackslash}p{21.5pt}>{\raggedleft\arraybackslash}p{21.5pt}>{\raggedleft\arraybackslash}p{16pt}>{\raggedleft\arraybackslash}p{16pt}>{\raggedleft\arraybackslash}p{16pt}>{\raggedleft\arraybackslash}p{16pt}>{\raggedleft\arraybackslash}p{21.5pt}@{}}
\toprule \textbf{Comparison} & \textbf{Metric} & \textbf{Subset} & $n$ & \textbf{agr} & $\kappa$ & $P_{pos}$ & $P_{neg}$ & $n_{11}$ & $n_{10}$ & $n_{01}$ & $n_{00}$ & $\alpha$\\\midrule\endfirsthead
\toprule \textbf{Comparison} & \textbf{Metric} & \textbf{Subset} & $n$ & \textbf{agr} & $\kappa$ & $P_{pos}$ & $P_{neg}$ & $n_{11}$ & $n_{10}$ & $n_{01}$ & $n_{00}$ & $\alpha$\\\midrule\endhead
\bottomrule\endfoot
Human vs human & Behaviour & overall & 300 & 0.853 & 0.669 & 0.891 & 0.776 & 180 & 33 & 11 & 76 & 0.667\\
 & Behaviour & R1 / R2 & 83 & 0.819 & 0.587 & 0.870 & 0.706 & 50 & 14 & 1 & 18 & 0.578\\
 & Behaviour & R1 / R3 & 84 & 0.821 & 0.595 & 0.870 & 0.717 & 50 & 13 & 2 & 19 & 0.589\\
 & Behaviour & R1 / R4 & 21 & 0.952 & 0.859 & 0.970 & 0.889 & 16 & 0 & 1 & 4 & 0.862\\
 & Behaviour & R2 / R3 & 83 & 0.867 & 0.719 & 0.893 & 0.825 & 46 & 5 & 6 & 26 & 0.720\\
 & Behaviour & R2 / R4 & 19 & 0.895 & 0.756 & 0.923 & 0.833 & 12 & 1 & 1 & 5 & 0.763\\
 & Behaviour & R3 / R4 & 10 & 1.000 & 1.000 & 1.000 & 1.000 & 6 & 0 & 0 & 4 & 1.000\\
 & Behaviour & type 1 No conflict & 63 & 0.937 & 0.714 & 0.964 & 0.750 & 53 & 2 & 2 & 6 & 0.716\\
 & Behaviour & type 2 Complementary & 60 & 0.783 & 0.536 & 0.835 & 0.683 & 33 & 12 & 1 & 14 & 0.522\\
 & Behaviour & type 3 Conflicting op. & 58 & 0.862 & 0.725 & 0.862 & 0.862 & 25 & 6 & 2 & 25 & 0.727\\
 & Behaviour & type 4 Outdated & 60 & 0.817 & 0.505 & 0.879 & 0.621 & 40 & 8 & 3 & 9 & 0.504\\
 & Behaviour & type 5 Misinformation & 59 & 0.864 & 0.725 & 0.879 & 0.846 & 29 & 5 & 3 & 22 & 0.727\\
 & STR & overall & 215 & 0.902 & 0.773 & 0.929 & 0.842 & 138 & 19 & 2 & 56 & 0.772\\
 & STR & R1 / R2 & 61 & 0.902 & 0.783 & 0.925 & 0.857 & 37 & 5 & 1 & 18 & 0.784\\
 & STR & R1 / R3 & 57 & 0.860 & 0.672 & 0.900 & 0.765 & 36 & 8 & 0 & 13 & 0.668\\
 & STR & R1 / R4 & 16 & 0.938 & 0.818 & 0.960 & 0.857 & 12 & 1 & 0 & 3 & 0.823\\
 & STR & R2 / R3 & 61 & 0.902 & 0.765 & 0.930 & 0.833 & 40 & 5 & 1 & 15 & 0.766\\
 & STR & R2 / R4 & 12 & 1.000 & 1.000 & 1.000 & 1.000 & 8 & 0 & 0 & 4 & 1.000\\
 & STR & R3 / R4 & 8 & 1.000 & 1.000 & 1.000 & 1.000 & 5 & 0 & 0 & 3 & 1.000\\
 & STR & type 1 No conflict & 56 & 0.911 & 0.711 & 0.945 & 0.762 & 43 & 5 & 0 & 8 & 0.710\\
 & STR & type 2 Complementary & 43 & 0.814 & 0.549 & 0.871 & 0.667 & 27 & 7 & 1 & 8 & 0.543\\
 & STR & type 4 Outdated & 57 & 0.930 & 0.812 & 0.953 & 0.857 & 41 & 4 & 0 & 12 & 0.812\\
 & STR & type 5 Misinformation & 59 & 0.932 & 0.865 & 0.931 & 0.933 & 27 & 3 & 1 & 28 & 0.866\\
 & FG claim & overall & 477 & 0.897 & 0.780 & 0.918 & 0.861 & 276 & 25 & 24 & 152 & 0.780\\
 & FG claim & R1 / R2 & 137 & 0.905 & 0.786 & 0.930 & 0.854 & 86 & 0 & 13 & 38 & 0.784\\
 & FG claim & R1 / R3 & 127 & 0.945 & 0.888 & 0.952 & 0.936 & 69 & 4 & 3 & 51 & 0.888\\
 & FG claim & R1 / R4 & 35 & 0.914 & 0.821 & 0.930 & 0.889 & 20 & 0 & 3 & 12 & 0.822\\
 & FG claim & R2 / R3 & 131 & 0.840 & 0.663 & 0.874 & 0.779 & 73 & 21 & 0 & 37 & 0.655\\
 & FG claim & R2 / R4 & 27 & 1.000 & 1.000 & 1.000 & 1.000 & 19 & 0 & 0 & 8 & 1.000\\
 & FG claim & R3 / R4 & 20 & 0.750 & 0.519 & 0.783 & 0.706 & 9 & 0 & 5 & 6 & 0.501\\
 & FG claim & type 1 No conflict & 77 & 0.987 & 0.972 & 0.990 & 0.982 & 49 & 0 & 1 & 27 & 0.972\\
 & FG claim & type 2 Complementary & 109 & 0.908 & 0.803 & 0.928 & 0.875 & 64 & 6 & 4 & 35 & 0.803\\
 & FG claim & type 3 Conflicting op. & 115 & 0.887 & 0.749 & 0.914 & 0.835 & 69 & 6 & 7 & 33 & 0.750\\
 & FG claim & type 4 Outdated & 85 & 0.800 & 0.566 & 0.844 & 0.721 & 46 & 10 & 7 & 22 & 0.568\\
 & FG claim & type 5 Misinformation & 91 & 0.912 & 0.821 & 0.923 & 0.897 & 48 & 3 & 5 & 35 & 0.821\\
\addlinespace
Human vs committee & Behaviour & overall & 650 & 0.745 & 0.407 & 0.814 & 0.591 & 364 & 68 & 98 & 120 & --\\
 & Behaviour & type 1 No conflict & 133 & 0.865 & 0.325 & 0.924 & 0.400 & 109 & 7 & 11 & 6 & --\\
 & Behaviour & type 2 Complementary & 130 & 0.708 & 0.279 & 0.804 & 0.424 & 78 & 4 & 34 & 14 & --\\
 & Behaviour & type 3 Conflicting op. & 128 & 0.820 & 0.641 & 0.816 & 0.824 & 51 & 13 & 10 & 54 & --\\
 & Behaviour & type 4 Outdated & 130 & 0.746 & 0.338 & 0.829 & 0.507 & 80 & 19 & 14 & 17 & --\\
 & Behaviour & type 5 Misinformation & 129 & 0.581 & 0.149 & 0.630 & 0.518 & 46 & 25 & 29 & 29 & --\\
 & Behaviour & Llama-3.1-8B & 328 & 0.738 & 0.355 & 0.817 & 0.538 & 192 & 39 & 47 & 50 & --\\
 & Behaviour & Qwen2.5-7B & 322 & 0.752 & 0.451 & 0.811 & 0.636 & 172 & 29 & 51 & 70 & --\\
 & STR strict & overall & 465 & 0.908 & 0.790 & 0.931 & 0.858 & 292 & 24 & 19 & 130 & --\\
 & STR strict & type 1 No conflict & 118 & 0.924 & 0.769 & 0.952 & 0.816 & 89 & 8 & 1 & 20 & --\\
 & STR strict & type 2 Complementary & 94 & 0.851 & 0.657 & 0.891 & 0.767 & 57 & 8 & 6 & 23 & --\\
 & STR strict & type 4 Outdated & 124 & 0.935 & 0.841 & 0.955 & 0.886 & 85 & 6 & 2 & 31 & --\\
 & STR strict & type 5 Misinformation & 129 & 0.907 & 0.814 & 0.910 & 0.903 & 61 & 2 & 10 & 56 & --\\
 & STR strict & Llama-3.1-8B & 233 & 0.914 & 0.801 & 0.938 & 0.863 & 150 & 11 & 9 & 63 & --\\
 & STR strict & Qwen2.5-7B & 232 & 0.901 & 0.779 & 0.925 & 0.854 & 142 & 13 & 10 & 67 & --\\
 & STR soft & overall & 465 & 0.912 & 0.798 & 0.935 & 0.863 & 295 & 21 & 20 & 129 & --\\
 & STR soft & type 1 No conflict & 118 & 0.924 & 0.762 & 0.952 & 0.809 & 90 & 7 & 2 & 19 & --\\
 & STR soft & type 2 Complementary & 94 & 0.872 & 0.701 & 0.908 & 0.793 & 59 & 6 & 6 & 23 & --\\
 & STR soft & type 4 Outdated & 124 & 0.935 & 0.841 & 0.955 & 0.886 & 85 & 6 & 2 & 31 & --\\
 & STR soft & type 5 Misinformation & 129 & 0.907 & 0.814 & 0.910 & 0.903 & 61 & 2 & 10 & 56 & --\\
 & STR soft & Llama-3.1-8B & 233 & 0.923 & 0.818 & 0.944 & 0.873 & 153 & 8 & 10 & 62 & --\\
 & STR soft & Qwen2.5-7B & 232 & 0.901 & 0.779 & 0.925 & 0.854 & 142 & 13 & 10 & 67 & --\\
 & FG claim & overall & 1029 & 0.899 & 0.783 & 0.920 & 0.862 & 599 & 38 & 66 & 326 & --\\
 & FG claim & type 1 No conflict & 165 & 0.945 & 0.880 & 0.958 & 0.922 & 103 & 4 & 5 & 53 & --\\
 & FG claim & type 2 Complementary & 233 & 0.910 & 0.805 & 0.929 & 0.876 & 138 & 8 & 13 & 74 & --\\
 & FG claim & type 3 Conflicting op. & 250 & 0.888 & 0.759 & 0.911 & 0.848 & 144 & 12 & 16 & 78 & --\\
 & FG claim & type 4 Outdated & 183 & 0.825 & 0.606 & 0.870 & 0.733 & 107 & 9 & 23 & 44 & --\\
 & FG claim & type 5 Misinformation & 198 & 0.929 & 0.855 & 0.939 & 0.917 & 107 & 5 & 9 & 77 & --\\
 & FG claim & Llama-3.1-8B & 518 & 0.880 & 0.711 & 0.916 & 0.795 & 336 & 21 & 41 & 120 & --\\
 & FG claim & Qwen2.5-7B & 511 & 0.918 & 0.834 & 0.926 & 0.907 & 263 & 17 & 25 & 206 & --\\
\addlinespace
Human consensus vs committee & Behaviour & overall & 256 & 0.789 & 0.483 & 0.852 & 0.630 & 156 & 24 & 30 & 46 & --\\
 & Behaviour & type 1 No conflict & 59 & 0.881 & 0.299 & 0.935 & 0.364 & 50 & 3 & 4 & 2 & --\\
 & Behaviour & type 2 Complementary & 47 & 0.787 & 0.391 & 0.865 & 0.500 & 32 & 1 & 9 & 5 & --\\
 & Behaviour & type 3 Conflicting op. & 50 & 0.880 & 0.760 & 0.885 & 0.875 & 23 & 2 & 4 & 21 & --\\
 & Behaviour & type 4 Outdated & 49 & 0.796 & 0.419 & 0.868 & 0.545 & 33 & 7 & 3 & 6 & --\\
 & Behaviour & type 5 Misinformation & 51 & 0.588 & 0.165 & 0.632 & 0.533 & 18 & 11 & 10 & 12 & --\\
 & STR strict & overall & 194 & 0.948 & 0.876 & 0.964 & 0.912 & 132 & 6 & 4 & 52 & --\\
 & STR strict & type 1 No conflict & 51 & 0.961 & 0.865 & 0.976 & 0.889 & 41 & 2 & 0 & 8 & --\\
 & STR strict & type 2 Complementary & 35 & 0.914 & 0.767 & 0.943 & 0.824 & 25 & 2 & 1 & 7 & --\\
 & STR strict & type 4 Outdated & 53 & 0.962 & 0.898 & 0.975 & 0.923 & 39 & 2 & 0 & 12 & --\\
 & STR strict & type 5 Misinformation & 55 & 0.945 & 0.891 & 0.947 & 0.943 & 27 & 0 & 3 & 25 & --\\
 & STR soft & overall & 194 & 0.954 & 0.888 & 0.967 & 0.920 & 133 & 5 & 4 & 52 & --\\
 & STR soft & type 1 No conflict & 51 & 0.961 & 0.865 & 0.976 & 0.889 & 41 & 2 & 0 & 8 & --\\
 & STR soft & type 2 Complementary & 35 & 0.943 & 0.838 & 0.963 & 0.875 & 26 & 1 & 1 & 7 & --\\
 & STR soft & type 4 Outdated & 53 & 0.962 & 0.898 & 0.975 & 0.923 & 39 & 2 & 0 & 12 & --\\
 & STR soft & type 5 Misinformation & 55 & 0.945 & 0.891 & 0.947 & 0.943 & 27 & 0 & 3 & 25 & --\\
 & FG claim & overall & 428 & 0.949 & 0.888 & 0.960 & 0.928 & 264 & 12 & 10 & 142 & --\\
 & FG claim & type 1 No conflict & 76 & 0.947 & 0.885 & 0.959 & 0.926 & 47 & 2 & 2 & 25 & --\\
 & FG claim & type 2 Complementary & 99 & 0.949 & 0.890 & 0.961 & 0.930 & 61 & 3 & 2 & 33 & --\\
 & FG claim & type 3 Conflicting op. & 102 & 0.961 & 0.912 & 0.971 & 0.941 & 66 & 3 & 1 & 32 & --\\
 & FG claim & type 4 Outdated & 68 & 0.897 & 0.762 & 0.925 & 0.837 & 43 & 3 & 4 & 18 & --\\
 & FG claim & type 5 Misinformation & 83 & 0.976 & 0.951 & 0.979 & 0.971 & 47 & 1 & 1 & 34 & --\\
\addlinespace
Human vs individual judge & Behaviour & local/deepseek-\newline r1-distill-32b & 650 & 0.735 & 0.339 & 0.820 & 0.503 & 391 & 41 & 131 & 87 & --\\
 & Behaviour & local/mistral-small-4 & 650 & 0.643 & 0.229 & 0.720 & 0.506 & 299 & 133 & 99 & 119 & --\\
 & Behaviour & local/qwen3.5-397b-a17b & 649 & 0.743 & 0.405 & 0.812 & 0.592 & 361 & 70 & 97 & 121 & --\\
\addlinespace
\end{longtable}}\par\nopagebreak[4]\vskip-9pt
\captionof{table}{\textbf{Complete CATS human-evaluation binary agreement.} All subsets for all four comparisons.  Conflicting opinions (type 3) is excluded from Single-Truth Recall, because that metric needs a single correct conclusion, and so has no STR rows.}\label{tab:catsbinfull}\par\vskip 7pt

\FloatBarrier

{\tabstyle\setlength{\tabcolsep}{3pt}\def\LTcaptype{}
\begin{longtable}{@{}>{\raggedright\arraybackslash}p{76pt}>{\raggedright\arraybackslash}p{88pt}>{\raggedleft\arraybackslash}p{20pt}>{\raggedleft\arraybackslash}p{24pt}>{\raggedleft\arraybackslash}p{24pt}>{\raggedleft\arraybackslash}p{28pt}>{\raggedleft\arraybackslash}p{34pt}>{\raggedleft\arraybackslash}p{41pt}>{\raggedleft\arraybackslash}p{32pt}>{\raggedleft\arraybackslash}p{32pt}@{}}
\toprule \textbf{Comparison} & \textbf{Subset} & $n$ & \textbf{exact} & \textbf{MAE} & \textbf{RMSE} & \textbf{Pearson} & \textbf{Spearman} & \textbf{mean A} & \textbf{mean B}\\\midrule\endfirsthead
\toprule \textbf{Comparison} & \textbf{Subset} & $n$ & \textbf{exact} & \textbf{MAE} & \textbf{RMSE} & \textbf{Pearson} & \textbf{Spearman} & \textbf{mean A} & \textbf{mean B}\\\midrule\endhead
\bottomrule\endfoot
Human vs human & overall & 300 & 0.860 & 0.096 & 0.278 & 0.807 & 0.796 & 0.647 & 0.635\\
 & R1 / R2 & 83 & 0.855 & 0.074 & 0.224 & 0.887 & 0.871 & 0.601 & 0.676\\
 & R1 / R3 & 84 & 0.929 & 0.062 & 0.236 & 0.865 & 0.869 & 0.617 & 0.627\\
 & R1 / R4 & 21 & 0.905 & 0.048 & 0.163 & 0.937 & 0.923 & 0.647 & 0.694\\
 & R2 / R3 & 83 & 0.783 & 0.169 & 0.383 & 0.693 & 0.686 & 0.738 & 0.569\\
 & R2 / R4 & 19 & 1.000 & 0.000 & 0.000 & 1.000 & 1.000 & 0.697 & 0.697\\
 & R3 / R4 & 10 & 0.600 & 0.233 & 0.408 & 0.641 & 0.638 & 0.433 & 0.667\\
 & type 1 No conflict & 63 & 0.984 & 0.016 & 0.126 & 0.965 & 0.965 & 0.640 & 0.656\\
 & type 2 Complementary & 60 & 0.850 & 0.092 & 0.263 & 0.826 & 0.798 & 0.672 & 0.625\\
 & type 3 Conflicting op. & 58 & 0.793 & 0.135 & 0.329 & 0.696 & 0.680 & 0.645 & 0.654\\
 & type 4 Outdated & 60 & 0.783 & 0.161 & 0.368 & 0.661 & 0.642 & 0.694 & 0.633\\
 & type 5 Misinformation & 59 & 0.881 & 0.079 & 0.251 & 0.855 & 0.862 & 0.583 & 0.606\\
\addlinespace
Human vs committee & overall & 650 & 0.865 & 0.095 & 0.281 & 0.805 & 0.798 & 0.627 & 0.655\\
 & type 1 No conflict & 133 & 0.932 & 0.050 & 0.207 & 0.901 & 0.885 & 0.657 & 0.662\\
 & type 2 Complementary & 130 & 0.854 & 0.109 & 0.305 & 0.751 & 0.754 & 0.633 & 0.660\\
 & type 3 Conflicting op. & 128 & 0.797 & 0.118 & 0.294 & 0.765 & 0.752 & 0.604 & 0.654\\
 & type 4 Outdated & 130 & 0.815 & 0.146 & 0.356 & 0.687 & 0.677 & 0.651 & 0.685\\
 & type 5 Misinformation & 129 & 0.922 & 0.054 & 0.213 & 0.897 & 0.894 & 0.588 & 0.614\\
 & Llama-3.1-8B & 328 & 0.845 & 0.109 & 0.299 & 0.766 & 0.755 & 0.675 & 0.713\\
 & Qwen2.5-7B & 322 & 0.885 & 0.082 & 0.261 & 0.836 & 0.832 & 0.578 & 0.596\\
\addlinespace
Consensus vs committee & overall & 258 & 0.938 & 0.045 & 0.195 & 0.909 & 0.906 & 0.648 & 0.636\\
 & type 1 No conflict & 62 & 0.935 & 0.046 & 0.195 & 0.914 & 0.897 & 0.651 & 0.648\\
 & type 2 Complementary & 51 & 0.922 & 0.069 & 0.252 & 0.836 & 0.844 & 0.650 & 0.641\\
 & type 3 Conflicting op. & 46 & 0.935 & 0.025 & 0.101 & 0.972 & 0.974 & 0.654 & 0.643\\
 & type 4 Outdated & 47 & 0.894 & 0.089 & 0.280 & 0.816 & 0.805 & 0.681 & 0.642\\
 & type 5 Misinformation & 52 & 1.000 & 0.000 & 0.000 & 1.000 & 1.000 & 0.606 & 0.606\\
\addlinespace
\end{longtable}}\par\nopagebreak[4]\vskip-9pt
\captionof{table}{\textbf{Complete continuous grounding-ratio agreement.} Per-sample Factual-Grounding ratios compared as continuous values. \emph{exact} uses a configured tolerance for equality. Correlation measures association, so it is read together with MAE/RMSE and the two means.}\label{tab:catscontfull}\par\vskip 7pt

\FloatBarrier

{\tabstyle\setlength{\tabcolsep}{3pt}\def\LTcaptype{}
\begin{longtable}{@{}>{\raggedright\arraybackslash}p{80pt}>{\raggedright\arraybackslash}p{46pt}>{\raggedright\arraybackslash}p{115.5pt}>{\raggedleft\arraybackslash}p{22pt}>{\raggedleft\arraybackslash}p{26pt}>{\raggedleft\arraybackslash}p{66pt}>{\raggedleft\arraybackslash}p{28pt}>{\raggedleft\arraybackslash}p{28pt}@{}}
\toprule \textbf{Subset} & \textbf{Metric} & \textbf{Judge pair} & $n$ & \textbf{agr} & $\kappa$ & $P_{pos}$ & $P_{neg}$\\\midrule\endfirsthead
\toprule \textbf{Subset} & \textbf{Metric} & \textbf{Judge pair} & $n$ & \textbf{agr} & $\kappa$ & $P_{pos}$ & $P_{neg}$\\\midrule\endhead
\bottomrule\endfoot
300-response frame & Behaviour & \textit{alpha / mean pairwise} & 300 & & 0.4568 / 0.4699 & & \\
 & & DeepSeek-32B vs Mistral-4 & 300 & 0.730 & 0.368 & 0.810 & 0.532\\
 & & DeepSeek-32B vs Qwen-397B & 300 & 0.853 & 0.611 & 0.903 & 0.703\\
 & & Mistral-4 vs Qwen-397B & 300 & 0.743 & 0.431 & 0.806 & 0.621\\
 & STR strict & \textit{alpha / mean pairwise} & 215 & & 0.6780 / 0.6874 & & \\
 & & DeepSeek-32B vs Mistral-4 & 215 & 0.809 & 0.613 & 0.839 & 0.766\\
 & & DeepSeek-32B vs Qwen-397B & 215 & 0.944 & 0.874 & 0.958 & 0.915\\
 & & Mistral-4 vs Qwen-397B & 215 & 0.791 & 0.575 & 0.825 & 0.740\\
 & FG claim & \textit{alpha / mean pairwise} & 477 & & 0.8706 / 0.8705 & & \\
 & & DeepSeek-32B vs Mistral-4 & 477 & 0.950 & 0.886 & 0.963 & 0.924\\
 & & DeepSeek-32B vs Qwen-397B & 477 & 0.941 & 0.869 & 0.956 & 0.913\\
 & & Mistral-4 vs Qwen-397B & 477 & 0.937 & 0.857 & 0.953 & 0.903\\
\addlinespace
\end{longtable}}\par\nopagebreak[4]\vskip-9pt
\captionof{table}{\textbf{Complete committee-internal agreement} among the three judge models on the 300-response frame.}\label{tab:cifull}\par\vskip 7pt

\FloatBarrier

\begin{multicols}{2}
\section{Complete Results Matrix}
\label{app:full}

Table~\ref{tab:full} lists decision accuracy (the share of examples on which the
model correctly answers or refuses), Refusal F1 and the CATS components for all
twelve model$\times$prompt configurations. Refusal recall (not tabulated; the
share of the 128 refusal examples that the model refuses) is $0.000$--$0.633$ for
baseline models and $0.898$--$1.000$ for trace-tuned models, and several
trace-tuned models refuse all 128.
\end{multicols}

\vskip 6pt
{\nolinenumbers{\tabstyle\setlength{\tabcolsep}{7.65pt}\def\LTcaptype{}\renewcommand{\arraystretch}{0.9}
\begin{longtable}{@{}lll rrrrr rrr@{}}
\toprule \textbf{Model} & \textbf{Prompt} & \textbf{Run} & \textbf{Dec-Acc} & \textbf{Ref-F1} & \textbf{BA} & \textbf{FG} & \textbf{STR} & \textbf{AQ} & \textbf{CATS-P} & \textbf{CATS-B}\\\midrule\endfirsthead
\toprule \textbf{Model} & \textbf{Prompt} & \textbf{Run} & \textbf{Dec-Acc} & \textbf{Ref-F1} & \textbf{BA} & \textbf{FG} & \textbf{STR} & \textbf{AQ} & \textbf{CATS-P} & \textbf{CATS-B}\\\midrule\endhead
\bottomrule\endfoot
Qwen2.5-7B & strict & base & 0.707 & 0.429 & 0.811 & 0.617 & 0.561 & 0.524 & 0.450 & 0.459\\
Qwen2.5-7B & strict & SFT & \textbf{0.967} & \textbf{0.911} & 0.780 & \textbf{0.838} & \textbf{0.687} & \textbf{0.677} & \textbf{0.630} & \textbf{0.684}\\
Qwen2.5-7B & runtime & base & 0.784 & 0.361 & 0.768 & 0.651 & 0.648 & 0.541 & 0.453 & 0.375\\
Qwen2.5-7B & runtime & SFT & \textbf{0.966} & \textbf{0.906} & \textbf{0.774} & \textbf{0.829} & \textbf{0.693} & \textbf{0.690} & \textbf{0.629} & \textbf{0.679}\\
Qwen2.5-7B & minimal & base & 0.827 & 0.031 & 0.737 & 0.031 & 0.757 & 0.033 & 0.022 & 0.017\\
Qwen2.5-7B & minimal & SFT & \textbf{0.958} & \textbf{0.881} & \textbf{0.768} & \textbf{0.837} & 0.686 & \textbf{0.680} & \textbf{0.617} & \textbf{0.659}\\
Qwen2.5-32B & strict & base & 0.855 & 0.367 & 0.789 & 0.851 & 0.733 & 0.724 & 0.574 & 0.432\\
Qwen2.5-32B & strict & SFT & \textbf{0.963} & \textbf{0.905} & \textbf{0.821} & 0.812 & \textbf{0.749} & 0.686 & \textbf{0.661} & \textbf{0.707}\\
Qwen2.5-32B & runtime & base & 0.855 & 0.389 & 0.799 & 0.864 & 0.780 & 0.760 & 0.600 & 0.455\\
Qwen2.5-32B & runtime & SFT & \textbf{0.962} & \textbf{0.901} & \textbf{0.804} & 0.814 & 0.758 & 0.701 & \textbf{0.658} & \textbf{0.716}\\
Qwen2.5-32B & minimal & base & 0.826 & 0.015 & 0.747 & 0.108 & 0.773 & 0.105 & 0.086 & 0.057\\
Qwen2.5-32B & minimal & SFT & \textbf{0.966} & \textbf{0.911} & \textbf{0.809} & \textbf{0.822} & 0.738 & \textbf{0.687} & \textbf{0.652} & \textbf{0.716}\\
Llama-3.1-8B & strict & base & 0.762 & 0.255 & 0.721 & 0.705 & 0.634 & 0.577 & 0.426 & 0.363\\
Llama-3.1-8B & strict & SFT & \textbf{0.973} & \textbf{0.926} & \textbf{0.805} & \textbf{0.852} & \textbf{0.727} & \textbf{0.706} & \textbf{0.680} & \textbf{0.744}\\
Llama-3.1-8B & runtime & base & 0.826 & 0.030 & 0.751 & 0.897 & 0.720 & 0.746 & 0.510 & 0.335\\
Llama-3.1-8B & runtime & SFT & \textbf{0.974} & \textbf{0.930} & \textbf{0.806} & 0.849 & 0.705 & 0.690 & \textbf{0.673} & \textbf{0.748}\\
Llama-3.1-8B & minimal & base & 0.825 & 0.015 & 0.743 & 0.074 & 0.710 & 0.055 & 0.029 & 0.018\\
Llama-3.1-8B & minimal & SFT & \textbf{0.976} & \textbf{0.934} & \textbf{0.795} & \textbf{0.857} & \textbf{0.714} & \textbf{0.694} & \textbf{0.672} & \textbf{0.732}\\
Mistral-7B & strict & base & 0.845 & 0.352 & 0.660 & 0.624 & 0.563 & 0.489 & 0.378 & 0.303\\
Mistral-7B & strict & SFT & \textbf{0.927} & \textbf{0.826} & \textbf{0.793} & \textbf{0.796} & \textbf{0.644} & \textbf{0.632} & \textbf{0.620} & \textbf{0.677}\\
Mistral-7B & runtime & base & 0.826 & 0.200 & 0.676 & 0.739 & 0.590 & 0.566 & 0.398 & 0.299\\
Mistral-7B & runtime & SFT & \textbf{0.929} & \textbf{0.831} & \textbf{0.760} & \textbf{0.762} & \textbf{0.613} & \textbf{0.593} & \textbf{0.587} & \textbf{0.657}\\
Mistral-7B & minimal & base & 0.823 & 0.000 & 0.727 & 0.079 & 0.700 & 0.073 & 0.048 & 0.025\\
Mistral-7B & minimal & SFT & \textbf{0.933} & \textbf{0.839} & \textbf{0.786} & \textbf{0.792} & 0.627 & \textbf{0.620} & \textbf{0.608} & \textbf{0.663}\\
\end{longtable}}\par\nopagebreak[4]\vskip-9pt
\captionof{table}{\textbf{End-to-end results, all twelve model$\times$prompt configurations} (24 rows). Run: base is the untuned model and SFT the trace-tuned model; bold marks an SFT value that exceeds its base value. Dec-Acc is decision accuracy and Ref-F1 the F1 of the refuse decision. BA, FG and STR are Behaviour Adherence, Factual Grounding and Single-Truth Recall; AQ, CATS-P and CATS-B are defined in Appendix~\ref{app:aggregate}. $n=736$ per row.}\label{tab:full}
\vskip 8pt
}

\end{document}